\documentclass[iicol,sn-mathphys-num, pdflatex]{sn-jnl}
\usepackage{graphicx}
\usepackage{adjustbox} 
\usepackage{lipsum}
\usepackage{multirow}
\usepackage{amsmath,amssymb,amsfonts}
\usepackage{amsthm}
\usepackage{mathrsfs}
\usepackage[title]{appendix}
\usepackage{xcolor}
\usepackage{textcomp}
\usepackage{manyfoot}
\usepackage{booktabs}
\usepackage{algorithm}
\usepackage{algorithmicx}
\usepackage{algpseudocode}
\usepackage{listings}
\usepackage{tabularx}
\usepackage{xurl}   
\usepackage{hyperref} 
\usepackage{float}
\usepackage{longtable}
\usepackage{enumitem} 
\usepackage{array}   
\usepackage{caption}  
\usepackage{ragged2e} 
\usepackage{ltablex} 
\usepackage{xltabular}
\usepackage{etoolbox}
\usepackage{diagbox}
\usepackage{color}

\usepackage[table]{xcolor}
\usepackage[T1]{fontenc}
\theoremstyle{thmstyleone}

\theoremstyle{thmstyletwo}

\theoremstyle{thmstylethree}

\begin{document}

\title[Article Title]{A Versatile Foundation Model for AI-enabled Mammogram Interpretation}

\author[1]{\fnm{Fuxiang} \sur{Huang}}
\author[2]{\fnm{Jiayi} \sur{Zhu}}
\author[3,4,5]{\fnm{Yunfang} \sur{Yu}}
\author[6]{\fnm{Yu} \sur{Xie}}
\author[7]{\fnm{Yuan} \sur{Guo}}
\author[8]{\fnm{Qingcong} \sur{Kong}}
\author[9]{\fnm{Mingxiang} \sur{Wu}}
\author[1]{\fnm{Xinrui} \sur{Jiang}}
\author[1]{\fnm{Shu} \sur{Yang}}
\author[1]{\fnm{Jiabo} \sur{Ma}}
\author[1]{\fnm{Ziyi} \sur{Liu}}
\author[1]{\fnm{Zhe} \sur{Xu}}
\author[1]{\fnm{Zhixuan} \sur{Chen}}
\author[3]{\fnm{Yujie} \sur{Tan}}
\author[3]{\fnm{Zifan} \sur{He}}
\author[3]{\fnm{Luhui} \sur{Mao}}
\author[1]{\fnm{Yequan} \sur{Bie}}
\author[1]{\fnm{Junlin} \sur{Hou}}
\author[10,11]{\fnm{Lei} \sur{Zhang}}
\author*[1]{\fnm{Xi} \sur{Wang}}\email{vancywangxi@ust.hk}
\author*[1,2]{\fnm{Qiong} \sur{Luo}}\email{luo@cse.ust.hk}
\author*[6]{\fnm{Zhenhui} \sur{Li}}\email{lizhenhui@kmmu.edu.cn}
\author*[3]{\fnm{Herui} \sur{Yao}}\email{yaoherui@mail.sysu.edu.cn}
\author*[1,12,13,14,15]{\fnm{Hao} \sur{Chen}}\email{jhc@ust.hk}

\affil[1]{\orgdiv{Department of Computer Science and Engineering}, \orgname{The Hong Kong University of Science and Technology}, \state{Hong Kong}, \country{China}}
\affil[2]{\orgdiv{Data Science and Analytics Thrust}, \orgname{The Hong Kong University of Science and Technology (Guangzhou)}, \city{Guangzhou}, \state{Guangdong}, \country{China}}
\affil[3]{\orgdiv{Guangdong Provincial Key Laboratory of Malignant Tumor Epigenetics and Gene Regulation, Guangdong-Hong Kong Joint Laboratory for RNA Medicine, Department of Medical Oncology, Breast Tumor Centre, Phase I Clinical Trial Centre}, \orgname{Sun Yat-sen Memorial Hospital, Sun Yat-sen University}, \city{Guangzhou}, \country{China}}
\affil[4]{\orgdiv{Guangdong Provincial Key Laboratory of Cancer Pathogenesis and Precision Diagnosis and Treatment, AI Big Data Laboratory, Department of Medical Oncology}, \orgname{Shenshan Medical Center, Memorial Hospital of Sun Yat-sen University}, \city{Shanwei}, \state{Guangdong}, \country{China}}
\affil[5]{\orgdiv{Institute for AI in Medicine and Faculty of Medicine}, \orgname{Macau University of Science and Technology}, \city{Taipa}, \state{Macao}, \country{China}}
\affil[6]{\orgdiv{Department of Radiology}, \orgname{The Third Affiliated Hospital of Kunming Medical University, Yunnan Cancer Hospital, Peking University Cancer Hospital Yunnan}, \city{Kunming}, \state{Yunnan}, \country{China}}
\affil[7]{\orgdiv{Department of Radiology}, \orgname{Guangzhou First People's Hospital, South China University of Technology}, \city{Guangzhou}, \state{Guangdong}, \country{China}}
\affil[8]{\orgdiv{Department of Radiology}, \orgname{The Third Affiliated Hospital, Sun Yat-Sen University}, \city{Guangzhou}, \state{Guangdong}, \country{China}}
\affil[9]{\orgdiv{Department of Radiology}, \orgname{Shenzhen People’s Hospital}, \city{Shenzhen}, \state{Guangdong}, \country{China}}
\affil[10]{\orgdiv{Chongqing Key Laboratory of Bio-perception and Multimodal Intelligent Information Processing}, \orgname{Chongqing University}, \state{Chongqing}, \country{China}}
\affil[11]{\orgdiv{School of Microelectronics and Communication Engineering}, \orgname{Chongqing University}, \state{Chongqing}, \country{China}}
\affil[12]{\orgdiv{Department of Chemical and Biological Engineering}, \orgname{The Hong Kong University of Science and Technology}, \state{Hong Kong}, \country{China}}
\affil[13]{\orgdiv{Division of Life Science}, \orgname{The Hong Kong University of Science and Technology}, \state{Hong Kong}, \country{China}}
\affil[14]{\orgname{HKUST Shenzhen-Hong Kong Collaborative Innovation Research Institute}, \city{Shenzhen}, \state{Guangdong}, \country{China}}
\affil[15]{\orgdiv{State Key Laboratory of Nervous System Disorders}, \orgname{The Hong Kong University of Science and Technology}, \state{Hong Kong}, \country{China}}

\abstract{Breast cancer is the most commonly diagnosed cancer and the leading cause of cancer-related mortality among women worldwide. Mammography plays a crucial role in the early detection and diagnosis of breast lesions. Despite advances in foundation models (FMs) for mammogram analysis, clinical translation is hindered by limitations such as insufficient diversity in training data and restricted model generalizability. To address these issues, we introduce VersaMammo, a versatile foundation model developed using the largest multi-institutional mammogram dataset to date, comprising 679,911 images. The proposed two-stage pre-training strategy enhances generalization through self-supervised learning and knowledge distillation. We established a comprehensive benchmark of 92 experiments across five clinical task types, where VersaMammo achieved state-of-the-art performance, ranking first in 50 of 68 internal and 20 of 24 external validation experiments. Additionally, in a controlled reader study, VersaMammo demonstrated performance on par with senior radiologists, underscoring its potential for reliable breast cancer screening and diagnosis.}

\keywords{Artificial Intelligence, Breast Cancer, Mammogram, Foundation Model, Knowledge Distillation, Generalizability.}

\maketitle

\section{Introduction}\label{intro}
Breast cancer is the most common malignancy worldwide and the leading cause of cancer-related mortality among women \cite{bray2024global, siegel2024cancer}. Early screening and accurate diagnosis are crucial for improving survival and treatment outcomes. Mammography remains a key screening and diagnostic tool due to its effectiveness in detecting early lesions \cite{marmot2013benefits, katalinic2020breast}. However, interpreting mammograms is challenging because they encompass diverse tissue types and lesion features, accompanied by substantial inter- and intra-patient variation in breast density, tissue patterns, breast size, compression, and imaging conditions. These factors demand high levels of radiologist expertise and experience. Dense breast tissue (e.g., densities C and D) can obscure subtle findings such as masses, architectural distortion, and microcalcifications, thereby increasing the risk of false negatives \cite{sorkhei2021csaw}. This variability also complicates automated analysis, as the same lesion may present with markedly different contrast and texture across patients and imaging conditions. Consequently, even experienced radiologists may face diagnostic uncertainty when interpreting such complex images. Studies have shown considerable inter-radiologist variability in mammogram interpretation, with \(\kappa=0.50\) for BI-RADS breast density \cite{ahmadzade2025ai} and \(\kappa=0.46\) for BI-RADS assessment when cancer is present \cite{kerlikowske1998variability}. 

With rapid advances in artificial intelligence (AI), particularly in medical image analysis \cite{rao2025multimodal, ma2025generalizable, zhao2025foundation, dong2026generative}, opportunities have emerged to address mammogram interpretation \cite{li2021dual, liu2021act, eisemann2025nationwide}. Most current AI approaches focus on specific tasks (e.g., classification, segmentation, or detection) by designing task-specific models. While effective in controlled settings, these methods often struggle in real clinical practice due to data heterogeneity and domain shifts across populations, imaging devices, and protocols. Foundation models (FMs) pretrained on large-scale data have shown promise for broad generalization and multi-task capabilities \cite{germani2025bias, he2024foundation}, but several studies reveal systematic biases and limited cross-institution generalization for mammography, with performance declines when deployed on data from different demographics, imaging protocols, or lesion prevalence \cite{mckinney2020international, ricci2022addressing}.

\begin{figure*}
\centering
\includegraphics[width=0.95\textwidth]{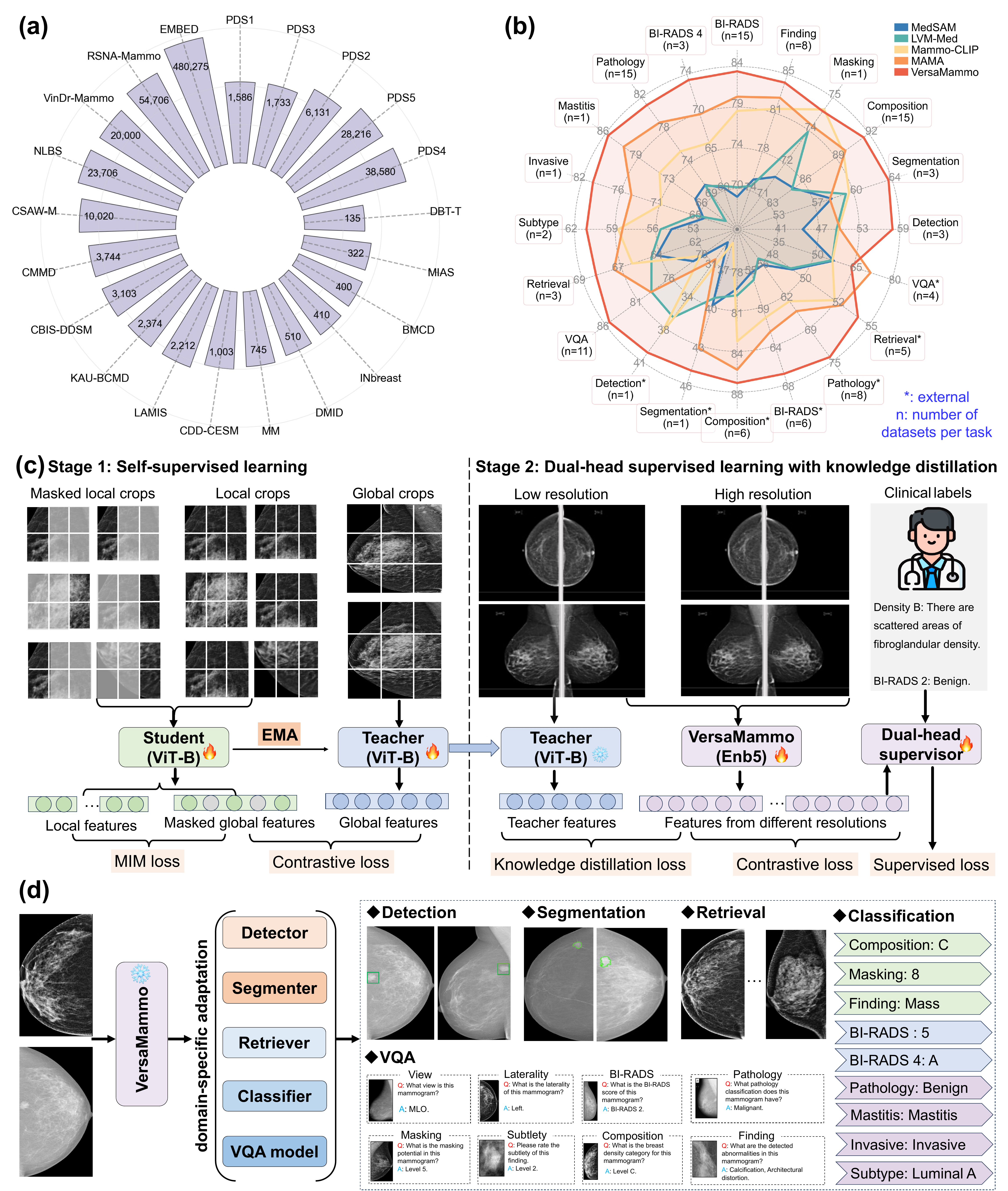}
\caption{\textbf{Overview of the VersaMammo}. (a) The VersaMammo dataset comprises a large-scale collection of 679,911 mammograms spanning 21 datasets, enabling comprehensive model training and evaluation.  Data splits, data distribution, and usage details are provided in Supplementary Tables 35-49. (b) Average performance of FMs across a diverse set of experiments.  If there are different backbones of the same FM, only the best model is presented here. (c) Overview of the two-stage hybrid pre-training strategy for VersaMammo. In the first stage, a teacher model is trained using self-supervised learning, involving Masked Image Modeling (MIM) loss and contrastive loss. The parameters are updated through Exponential Moving Average (EMA). 
The second stage utilizes a combination of losses, including knowledge distillation loss, dual-head supervised loss, and contrastive loss, integrating both teacher model knowledge and clinical supervision (guideline-grounded labels). 
(d) Downstream tasks used to evaluate the FMs.} 
\label{fig:versamammo}
\end{figure*}

FMs have demonstrated remarkable generalization and multi-task abilities across domains, including natural language processing \cite{cai2024internlm2, dubey2024llama, yang2024qwen2}, computer vision \cite{caron2021emerging, oquab2023dinov2, he2022masked, kirillov2023segment}, and medical image analysis \cite{lvmmed, medsam, hao2024large, bluethgen2024vision, wang2024pathology}. Pre-training strategies generally fall into three paradigms: unsupervised/self-supervised \cite{oquab2023dinov2, he2022masked, lvmmed}, weakly supervised \cite{kim2021vilt, bluethgen2024vision, zhao2025clip}, and supervised \cite{he2016identity, dosovitskiy2020image}. Self-supervised methods learn generic representations from unlabeled data via pretext tasks such as Masked Image Modeling (MIM) or contrastive learning. They leverage abundant unlabeled data and incur no annotation cost, but may capture features not perfectly aligned with downstream tasks. Weakly supervised approaches use imperfect supervisory signals (e.g., image-text pairs or noisy labels) to enable cross-modal understanding and robustness to real-world noise, yet label noise can limit precision. Supervised pre-training relies on large-scale, accurately labeled data to optimize task-specific performance. It can yield strong results when test data closely match training distributions, but it is vulnerable to domain shift, and obtaining high-quality annotations is costly. These pre-training strategies offer complementary avenues for building foundation models. Unsupervised and weakly supervised methods are effective at learning generalizable visual features from vast data with minimal annotation, providing a robust starting point for diverse applications. Supervised pre-training can inject precise, task-aligned semantic knowledge when high-quality labeled data are available. Their promise has inspired medical-imaging work such as Mammo-CLIP \cite{mammoclip} and MAMA \cite{du2024multi}, demonstrating the potential of FMs in mammogram analysis.

However, many existing mammogram FMs rely on high-quality paired image–report data, which are scarce. While several mammography datasets exist, most are designed for a single downstream task and do not provide robust image–report pairs, hindering multimodal representation learning and multi-task development. For example, MammoCLIP \cite{mammoclip} was trained on 13,829 patient–report pairs (25,355 screening mammograms) restricted to BI-RADS 0–2, excluding higher-suspicion and malignant cases, limiting dataset scale and clinical coverage. MAMA attempted to mitigate data scarcity by generating structured reports from publicly available embeddings, but this synthetic approach still fails to capture the full spectrum of clinical presentations and pathologies.
Consequently, limitations in data scale and diversity lead to poor generalization to unseen imaging data, restricting real-world clinical applicability. Recently, benchmarks for medical FMs have begun to emerge \cite{jin2024fairmedfm, zhu2025benchmark}. However, a comprehensive evaluation of mammogram FMs across a wide range of clinically relevant tasks remains insufficient. Most studies focus on common tasks such as classification or lesion detection, often neglecting segmentation, image retrieval, and mammogram visual question answering (VQA). As a result, current evaluations may not fully reflect their utility in real-world mammography workflows.

To address these gaps, we present VersaMammo, as illustrated in Fig. \ref{fig:versamammo}, a comprehensive foundation model trained on the largest multi-institutional mammogram dataset to date. VersaMammo adopts a two-stage pre-training strategy that blends self-supervised feature extraction with supervised learning and knowledge distillation from clinical guidance. We evaluate VersaMammo with an extensive benchmark spanning five clinical task types and 92 experiments, enabling a holistic assessment of its diagnostic performance and generalization. Our contributions are summarized as follows:

\begin{itemize}
\item 
We curated the largest and most diverse mammogram dataset to date, totaling 679,911 mammograms from 21 datasets (16 public, 5 private). We used 389,100 mammograms from the training splits of the three largest datasets for pre-training, followed by internal and external validation on a wide range of downstream tasks.

\item We introduce VersaMammo, an FM for mammograms, featuring a two-stage pre-training strategy that integrates self-supervised representation learning with supervised learning and knowledge distillation.

\item We established a comprehensive benchmark covering five clinical task types: lesion detection, segmentation, mammogram retrieval, mammogram VQA, and clinical classification. For clinical classification, we adopt nine distinct configurations (e.g., breast composition assessment, BI-RADS assessment, etc.). Across all task types and configurations, we evaluated a total of 92 experiments (68 internal validation and 24 external validation). VersaMammo is compared against four state-of-the-art FMs across seven variants.

\item Extensive results demonstrate that VersaMammo outperforms baselines in 50 of 68 internal experiments (average rank 1.5) and 20 of 24 external experiments (average rank 1.2), demonstrating strong generalization. BI-RADS assessment performance surpasses that of junior and mid-level radiologists and approaches that of senior radiologists. Moreover, human–AI collaboration leveraging VersaMammo outputs improves diagnostic accuracy, efficiency, and consistency.
\end{itemize}

\begin{figure*}
\centering
\includegraphics[width=0.95\textwidth]{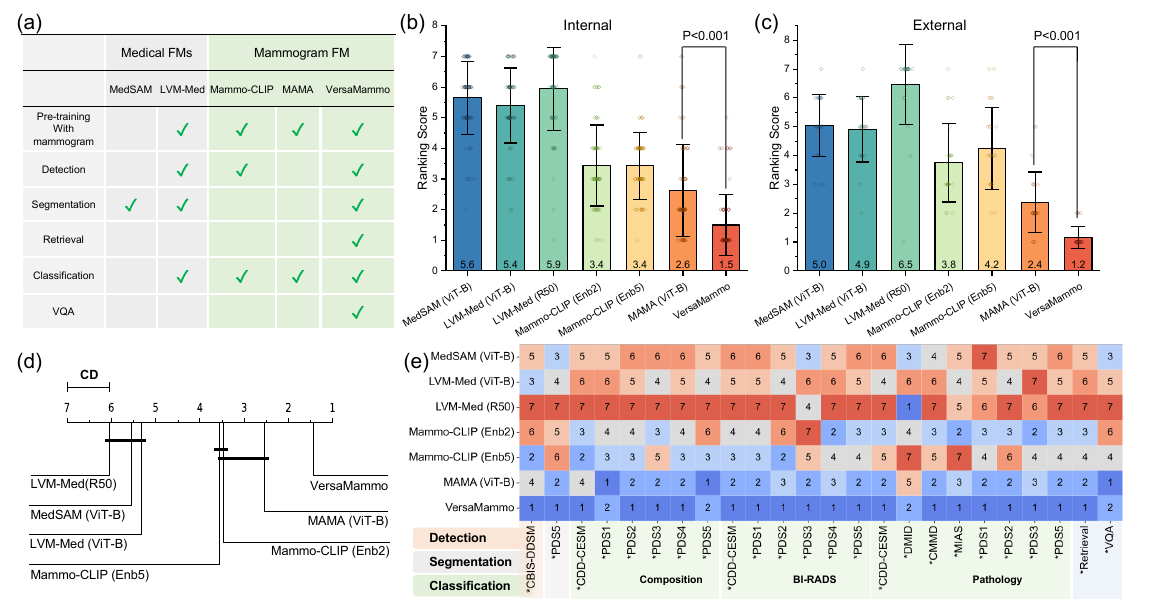}
\caption{\textbf{Comprehensive Comparison of FMs across 92 {experiments}.} (a) Comparison of different FMs. (b-c) Average rank of FMs across 68 internal downstream tasks and 24 external validation {experiments}. The error bars indicate the 95\% CI, and the box limits represent the standard error. The Wilcoxon signed-rank two-side test is employed for data analysis (1000 bootstrap replicates). (d) Critical differences (CD) diagram of an average ranking score with the Nemenyi test. In the CD figure, there are no significant differences between the models covered by the black line. (e) Ranks of FMs across 24 external validation {experiments}. $*$ means external dataset. If a model achieves the best performance, its rank value is set to 1. 
Ranks of 68 internal validation {experiments} is provided in  Extended Data Fig. 1. }
\label{fig:all_results}
\end{figure*}

\begin{figure*}
\centering
\includegraphics[width=0.95\textwidth]{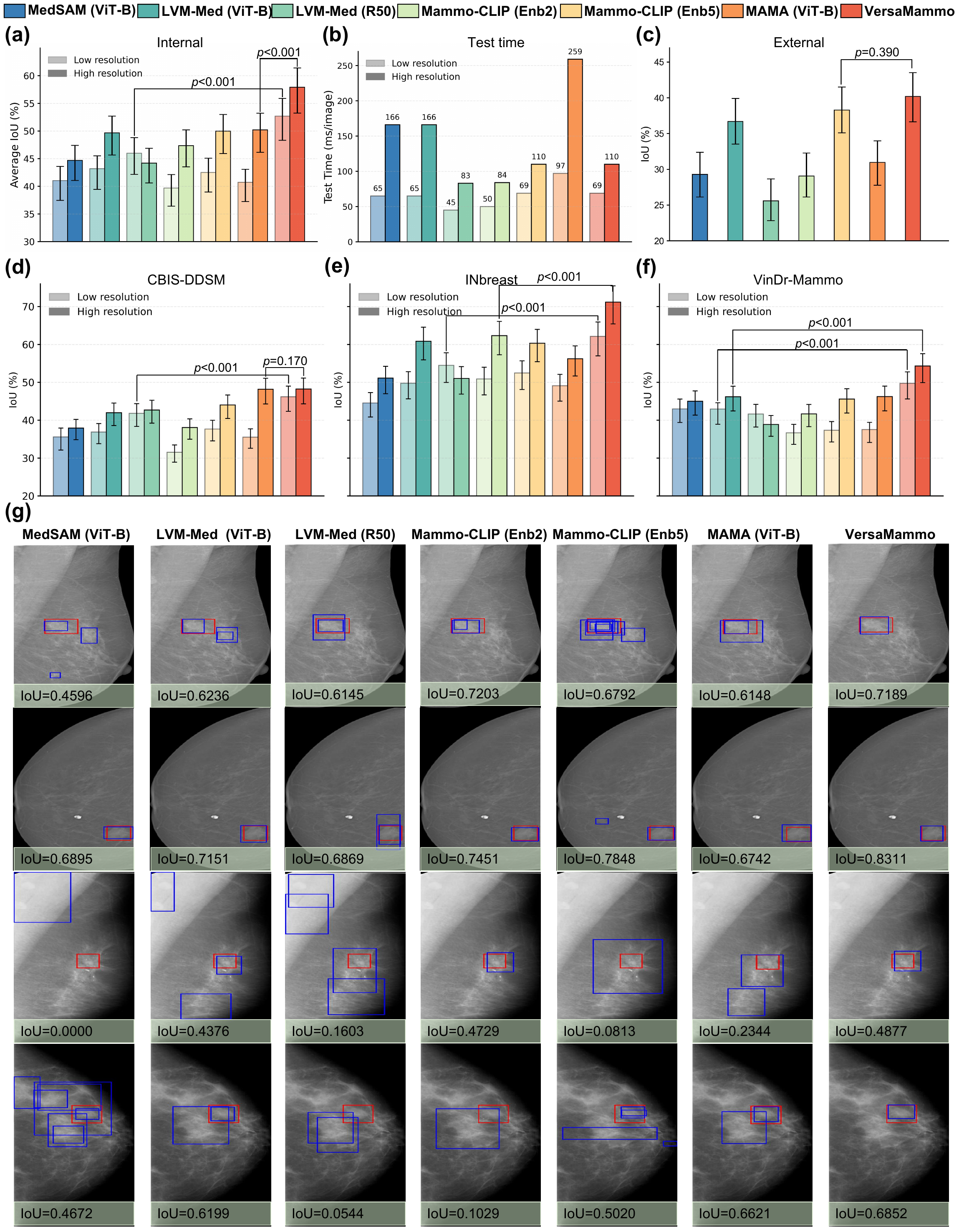}
\caption{\textbf{Performance of FMs on lesion detection tasks.} (a) Average IoU of different resolutions across all internal lesion detection experiments. (b) Average test time of different resolutions. (c) IoU of the external lesion detection experiment. (d-f) IoU of different resolutions on different datasets for internal lesion detection experiments. The Wilcoxon signed-rank two-side test is employed for data analysis (1000 bootstrap replicates). Error bars represent 95\% CI. The centre indicates mean. For the internal validation, n = 344 (CBIS-DDSM), 19 (INbreast), and 198 (VinDr-Mammo) independent cases. For the external validation, n = 344 independent cases. (g) Visualization results of lesion detection. Red bounding boxes represent ground truth annotations, while blue bounding boxes indicate model predictions. {The first two rows are selected from the top 50\% worst-performing and best-performing cases in the INbreast test set, respectively. The last two rows are selected from the top 50\% worst-performing and best-performing cases in the CBIS-DDSM test set, respectively.}
}
\label{fig:det}
\end{figure*}

\begin{figure*}
\centering
\includegraphics[width=0.95\textwidth]{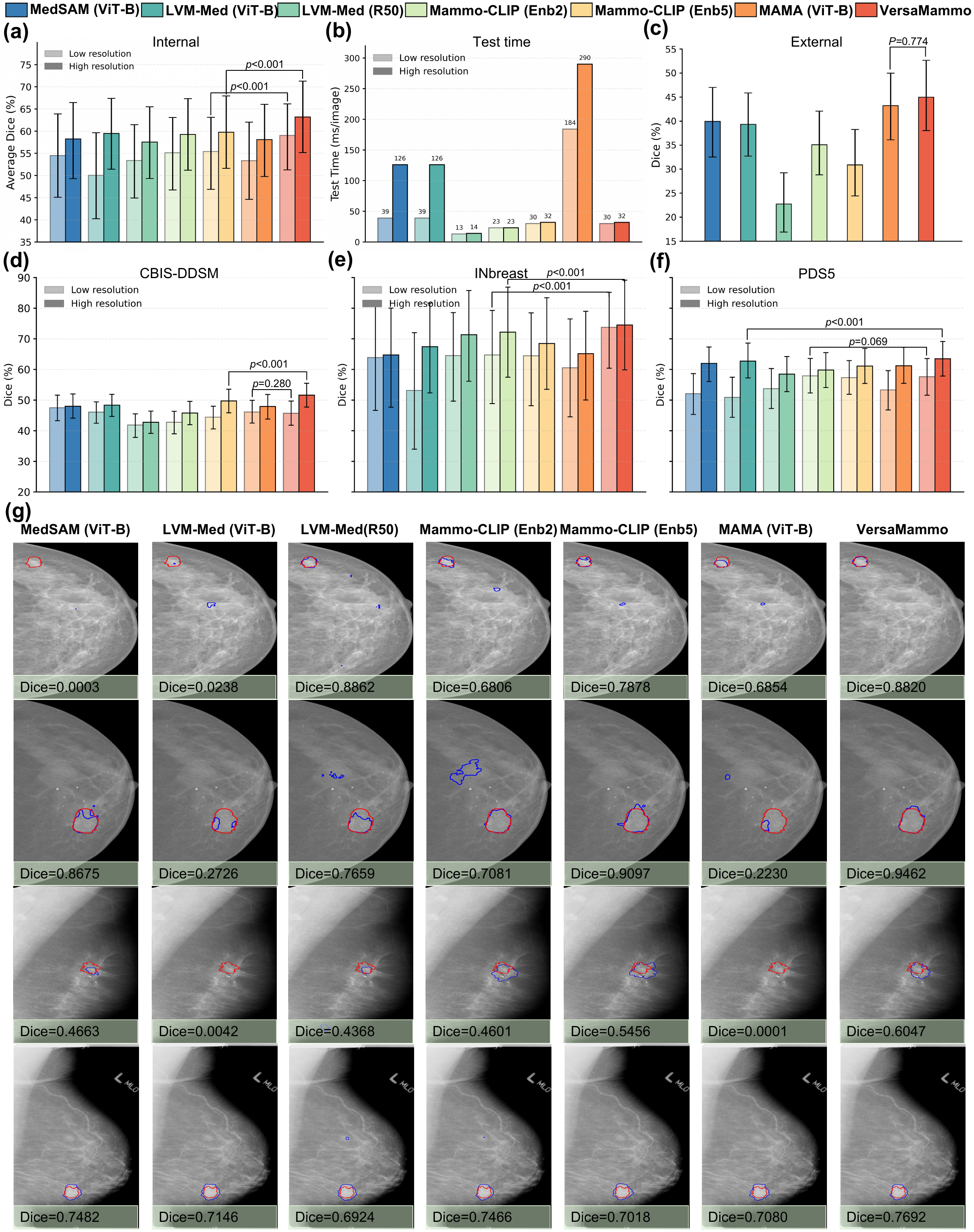}
\caption{\textbf{Performance of FMs on lesion segmentation tasks.} (a) Average Dice of different resolutions across all internal lesion detection experiments. (b) Average test time of different resolutions. (c) Dice of the external lesion detection experiment. (d-f) Dice of different resolutions on different datasets for internal lesion detection experiments. The Wilcoxon signed-rank two-side test is employed for data analysis (1000 bootstrap replicates). Error bars represent 95\% CI. The centre indicates mean. For the internal validation, n = 344 (CBIS-DDSM), 19 (INbreast), and 115 (PDS5) independent cases. For the external validation, n = 115 independent cases. (g) Representative visualization of lesion segmentation. Red regions denote ground truth masks, while blue regions represent predictions generated by the models. {The first two rows are selected from the top 50\% worst-performing and best-performing cases in the INbreast test set, respectively. The last two rows are selected from the top 50\% worst-performing and best-performing cases in the CBIS-DDSM test set, respectively.} }
\label{fig:seg}
\end{figure*}

\section{Results}
\label{sec:results}

\subsection{Comprehensive Benchmark Performance}
We evaluated VersaMammo and other FMs on 21 datasets across 5 downstream task types, totaling 92 experiments: lesion detection \cite{ren2016faster, lin2017feature}, lesion segmentation \cite{chen2024transunet, ronneberger2015u}, clinical classification, mammogram retrieval, and mammogram VQA \cite{zhu2025benchmark}. Evaluations were conducted under internal validation (fine-tuning only task-specific heads) and external validation (held out from both pre-training and fine-tuning). All procedures followed the evaluation protocol described in the Supplementary (Section 1) to ensure fair comparisons and robust assessment of generalization.

Statistical comparisons used nonparametric bootstrapping with 1,000 resamples. Performance metrics included: balanced accuracy (BACC) \cite{brodersen2010balanced}, Area Under the receiver operating characteristic Curve (AUC) \cite{bradley1997use}, and macro-F1 score \cite{hand2021f} for classification and VQA; Top-$k$ accuracy (ACC@k) \cite{han2015efficient} for retrieval; Dice coefficient (Dice) \cite{jha2019neutrosophic} for lesion segmentation and Intersection over Union (IoU) \cite{rezatofighi2019generalized} for detection tasks. For each metric, 95\% confidence intervals were derived from the 2.5th and 97.5th percentiles of the bootstrap distribution. Two-sided p-values were obtained from bootstrapped performance differences. Benjamini–Hochberg \cite{benjamini1995controlling} false discovery rate ($FDR = 0.05$) corrections were applied separately for internal and external validations, stratified by task type and metric. 

VersaMammo ranked first in 50 of 68 internal validation experiments (average rank 1.5, in Fig. \ref{fig:all_results}(b) and Extended Data Fig. 1) and 20 of 24 external validation experiments (average rank 1.2, in Fig. \ref{fig:all_results}(c) and (e)). The second-best model (MAMA) ranked first in 13 internal and 3 external experiments. A Wilcoxon signed-rank test comparing VersaMammo to MAMA yielded $p < 0.001$ in both internal and external settings, confirming the superiority of VersaMammo. The Nemenyi \cite{demvsar2006statistical} test likewise showed a statistically significant difference between VersaMammo and other FMs. To compare models across heterogeneous metrics, we used an average-rank approach with a critical-difference (CD) diagram \cite{ma2025generalizable} (Fig. \ref{fig:all_results}(d)). More detailed results are presented in Supplementary Table 1. Collectively, these results demonstrate state-of-the-art performance and superior generalization across diverse mammogram interpretation tasks.

\subsubsection{Lesion Detection}
\label{subsec:lesion_detection}

We conducted three internal lesion detection experiments: all FMs were fine-tuned and evaluated on CBIS-DDSM \cite{lee2017curated}, INbreast \cite{moreira2012inbreast}, and VinDr-Mammo \cite{nguyen2023vindr} at high resolution ($518\times518$ for MAMA (ViT-B) per original implementation; $512\times512$ for others) and low resolution ($224\times224$), reporting IoU and inference time (NVIDIA L20 GPU, 48GB). VersaMammo achieved the highest average IoU across all experiments, significantly outperforming the second-best models LVM-Med (R50) and MAMA (ViT-B) ($p<0.001$) (Fig. \ref{fig:det} (a-b,d-f)). At high resolution, VersaMammo averaged 57.92\% IoU, surpassing MAMA (ViT-B) by 7.71\% ($p<0.001$), while maintaining superior efficiency: it achieves a 1.51$\times$  speedup over MedSAM (ViT-B) and LVM-Med (ViT-B) at 166 ms per image, and a 2.4$\times$ speedup over MAMA (ViT-B) at 259 ms, with a runtime of only 110 ms per image. At low resolution, VersaMammo achieved a mean IoU of 52.69\% (46.18\% on CBIS-DDSM, 62.15\% on INbreast, 49.74\% on VinDr-Mammo) with 69 ms per image, outperforming LVM-Med (R50) by 6.7\% ($p<0.001$) while remaining competitive in speed. Consistent localization efficacy is visually corroborated in Fig. \ref{fig:det} (g).

For external validation at high resolution, which more closely reflects clinical reading conditions, all FMs were jointly fine-tuned on the training splits of INbreast and VinDr-Mammo and evaluated on the test split of CBIS-DDSM as an independent cohort. VersaMammo delivered the best performance with 40.17\% IoU (Fig. \ref{fig:det} (c)), demonstrating robust generalization to unseen clinical data. Detailed results are in Supplementary Tables 2-4.

\subsubsection{Lesion Segmentation}
\label{subsec:lesion_segmentation}

We conducted three internal lesion segmentation experiments, where all FMs were fine-tuned and evaluated on CBIS-DDSM, INbreast, and PDS5 at both high ($512\times512$) and low ($224\times224$) resolutions, reporting Dice and inference time (NVIDIA L20 GPU, 48GB memory). At high resolution, VersaMammo achieved an average Dice of 63.21\% (51.61\% on CBIS-DDSM, 74.53\% on INbreast, 63.50\% on PDS5), surpassing the second-best model Mammo-CLIP (Enb5) by 3.44\% ($p<0.001$) (Fig. \ref{fig:seg} (a-b,d-f)). It required only 32 ms per image, achieving a 3.94$\times$ speedup over MedSAM (ViT-B) and LVM-Med (ViT-B), both of which require 126 ms. At low resolution, VersaMammo obtained an average Dice of 59.04\% (45.71\% on CBIS-DDSM, 73.79\% on INbreast, 57.63\% on PDS5) with 30 ms per image, notably outperforming Mammo-CLIP (Enb5) by 9.34\% on INbreast ($p<0.001$). Qualitative visualizations confirm accurate boundary delineation.

For external validation at high resolution, all FMs were jointly fine-tuned on the training splits of INbreast and CBIS-DDSM and evaluated on the test split of PDS5 as an independent cohort. VersaMammo achieved a Dice of 44.95\% (Fig. \ref{fig:seg} (c)), outperforming the second-best model, MAMA (ViT-B), by 1.75\% ($p<0.001$). Detailed results are in Supplementary Tables 5-7.
\begin{figure*}
\centering
\includegraphics[width=0.95\textwidth]{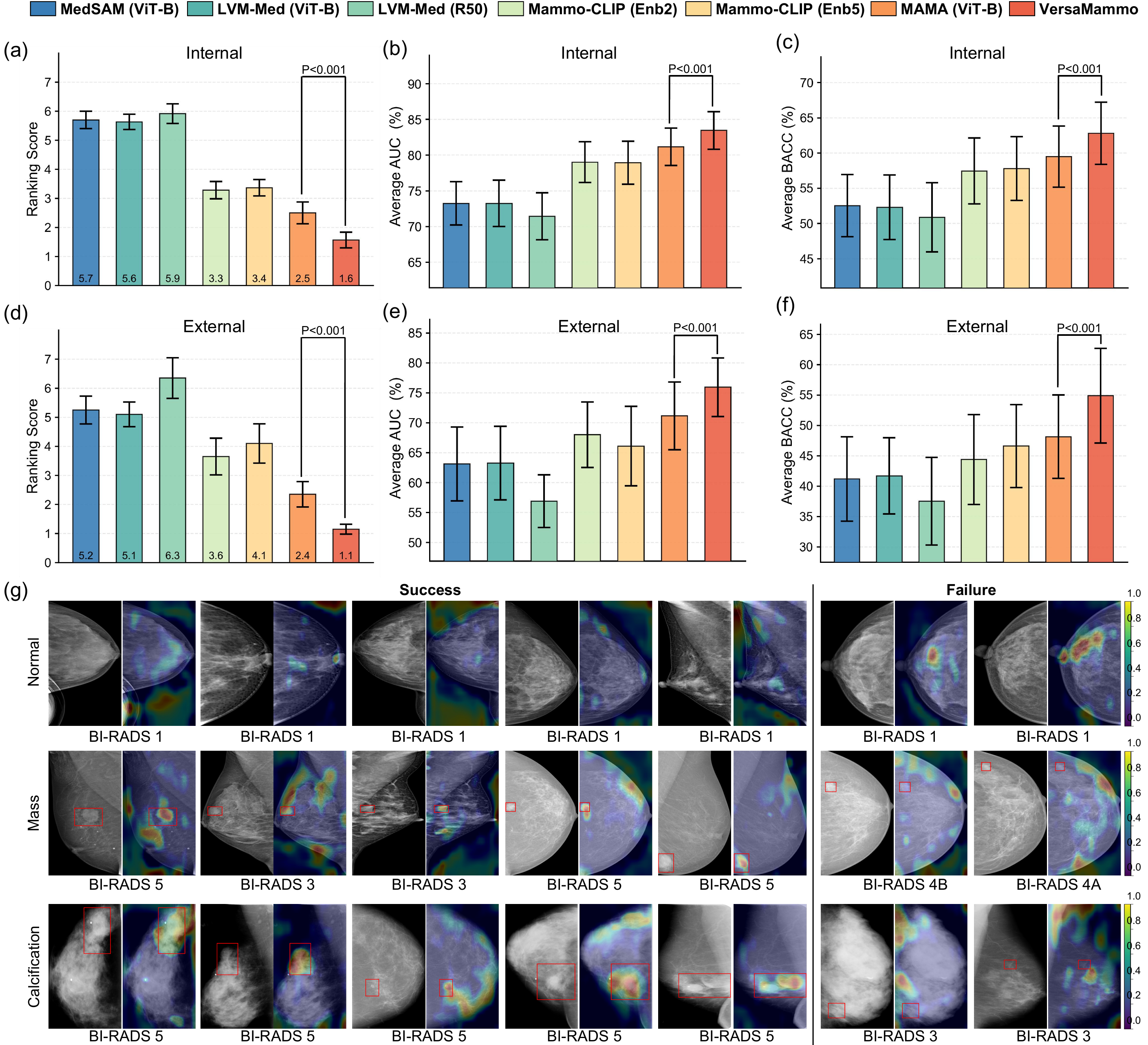}
\caption{\textbf{Performance evaluation of FMs across classification tasks}. (a,d) Average model rank based on AUC across 60 internal and 20 external classification validation {experiments}, respectively. (b,c,e,f) Average AUC and BACC for internal and external classification validation {experiments}, respectively. The Wilcoxon signed-rank two-side test is employed for data analysis (1000 bootstrap replicates). Error bars represent 95\% CI. The centre indicates mean. For the internal validation, n = 143,504 independent cases.
For the external validation, n = 25,640 independent cases.
(g) Attention visualization for three clinical categories: normal tissue, mass, and calcification, including both successful and failure cases. The red bounding boxes represent the areas of concern marked by the radiologists. 
}
\label{fig:cls}
\end{figure*}

\subsubsection{Mammogram Classification for Diverse Clinical Tasks}
\label{subsec:clinical_classification}
We conducted 60 internal evaluation experiments at high ($512\times512$) resolution across nine clinically critical configurations: breast composition assessment, masking potential risk evaluation, finding abnormality identification, BI-RADS assessment, BI-RADS 4 subclassification, pathological diagnosis prediction, mastitis differentiation, invasive status determination, and molecular subtype prediction. Label details and distributions across datasets are provided in \textbf{Extended Data Tables 1-2} and Supplementary Tables 39-49. FMs were fine-tuned and evaluated on respective internal datasets using AUC, BACC, and macro-F1. VersaMammo achieved an average rank of 1.6 (Fig. \ref{fig:cls} (a)), significantly outperforming the second-best model MAMA (ViT-B) with an average rank of 2.5. It demonstrated consistently high performance across all experiments (Fig. \ref{fig:cls} (b-c)), with an average AUC of 83.44\% and BACC of 62.80\% ($p<0.001$).

For external validation, we conducted 20 experiments: all FMs were jointly fine-tuned on the training splits of the three pre-training datasets. For each experiment, datasets without corresponding labels were excluded from training; evaluation was performed on test splits of independent external cohorts. VersaMammo maintained robust performance with a superior average rank of 1.2 (Fig. \ref{fig:cls} (d)), achieving an average AUC of 75.91\% (Fig. \ref{fig:cls} (e)) and BACC of 54.88\% (Fig. \ref{fig:cls} (f)) ($p<0.001$). Detailed results are presented in Supplementary Tables 8-19.

\textbf{Interpretability Analysis.} For transparent decision-making, we analyze VersaMammo's attention patterns using Grad-CAM \cite{selvaraju2017grad}. The results showed attention patterns that were generally consistent with clinically relevant regions (Fig. \ref{fig:cls} (g)). For normal tissue, successful cases show no specific focus, while failures exhibit scattered attention over dense parenchyma. In mass detection, attention concentrates on masses with clear margins, but low-contrast masses in dense tissue cause attention dispersion or missed detection. For calcifications, malignant cases receive more precise, clustered attention, while benign calcifications trigger weaker or dispersed focus. Model attention generally aligns with clinically relevant features; discrepancies occur mainly with subtle findings in dense tissue. These visualizations confirm that VersaMammo learns clinically relevant features while revealing limitations in challenging cases.

\subsubsection{Mammogram Retrieval}
\label{subsec:retrieval}
\begin{figure*}
\centering
\includegraphics[width=0.95\textwidth]{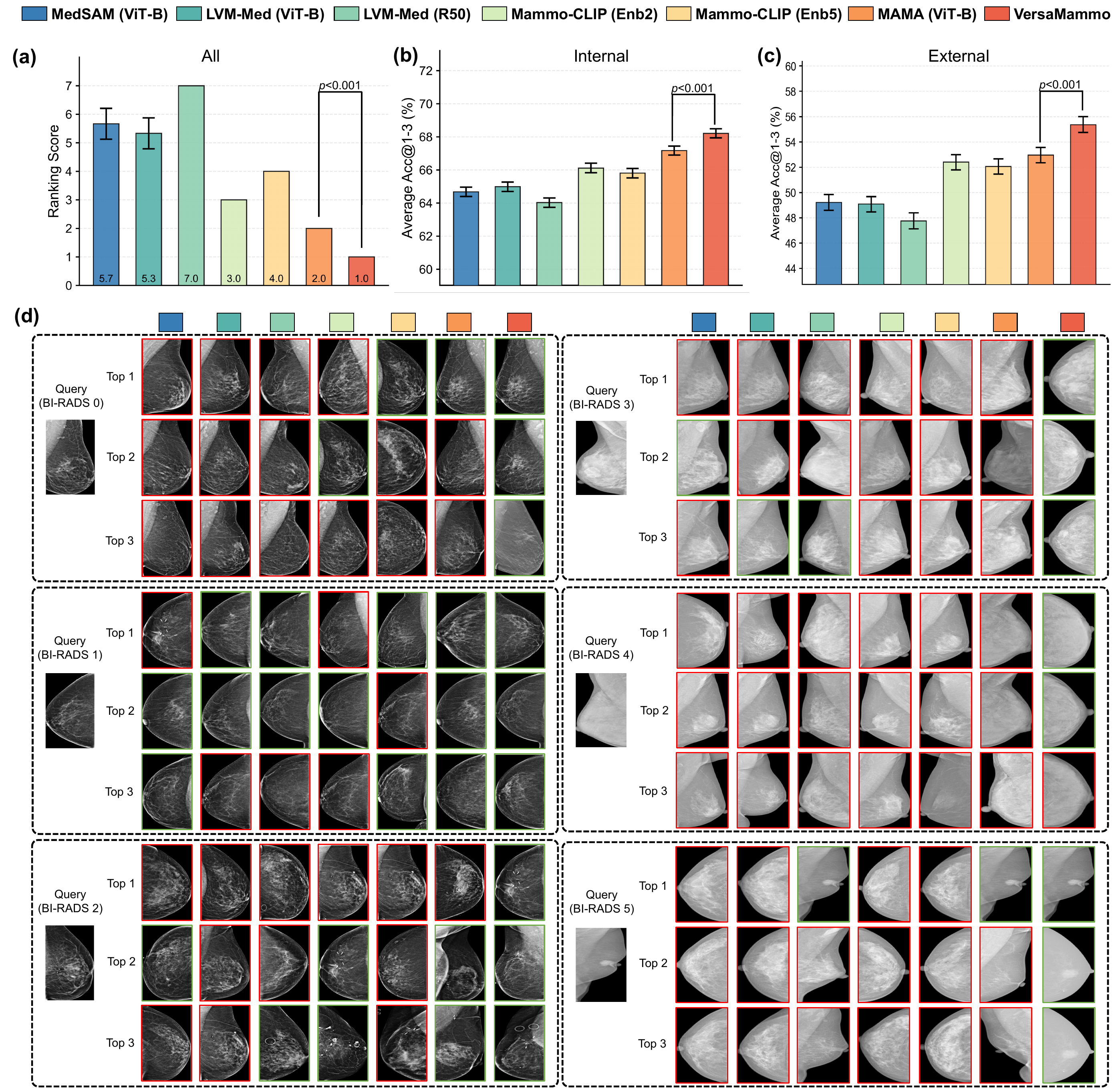}
\caption{\textbf{Performance of FMs on mammogram retrieval tasks.} (a) Average rank of FMs on image retrieval tasks based on top-1 to top-3 accuracy across internal and external datasets. Without a confidence interval, the rank appears to be fixed at a certain position. (b-c) Average accuracy for internal and external {experiments}. The Wilcoxon signed-rank two-side test is employed for data analysis (1000 bootstrap replicates). Error bars represent 95\% CI. The centre indicates mean. For the internal validation, n = 109,270 independent cases.
For the external validation, n = 24,523 independent cases. (d) Qualitative results for mammogram retrieval. The red box and the green box indicate incorrect and correct retrieval results, respectively.}
\label{fig:retrieval}
\end{figure*}

We evaluated mammogram retrieval at high resolution ($512\times512$) without task-specific training, using each dataset's training split as the search database and test split images as queries. Internal validation was performed on the test splits of three pre-training datasets (EMBED, RSNA, VinDr-Mammo), and external validation on the test splits of five private clinical datasets (PDS1–PDS5) unseen during pre-training, following patient-wise splits detailed in Supplementary Tables 35-36. VersaMammo achieved the highest rank (Fig. \ref{fig:retrieval} (a)), with leading average accuracies of 68.21\% (internal) and 55.35\% (external) (Fig. \ref{fig:retrieval} (b-c)). Detailed results are in Supplementary Table 20.

\subsubsection{Mammogram Visual Question Answering}
\label{subsec:vqa}

\begin{figure*}
\centering
\includegraphics[width=0.95\textwidth]{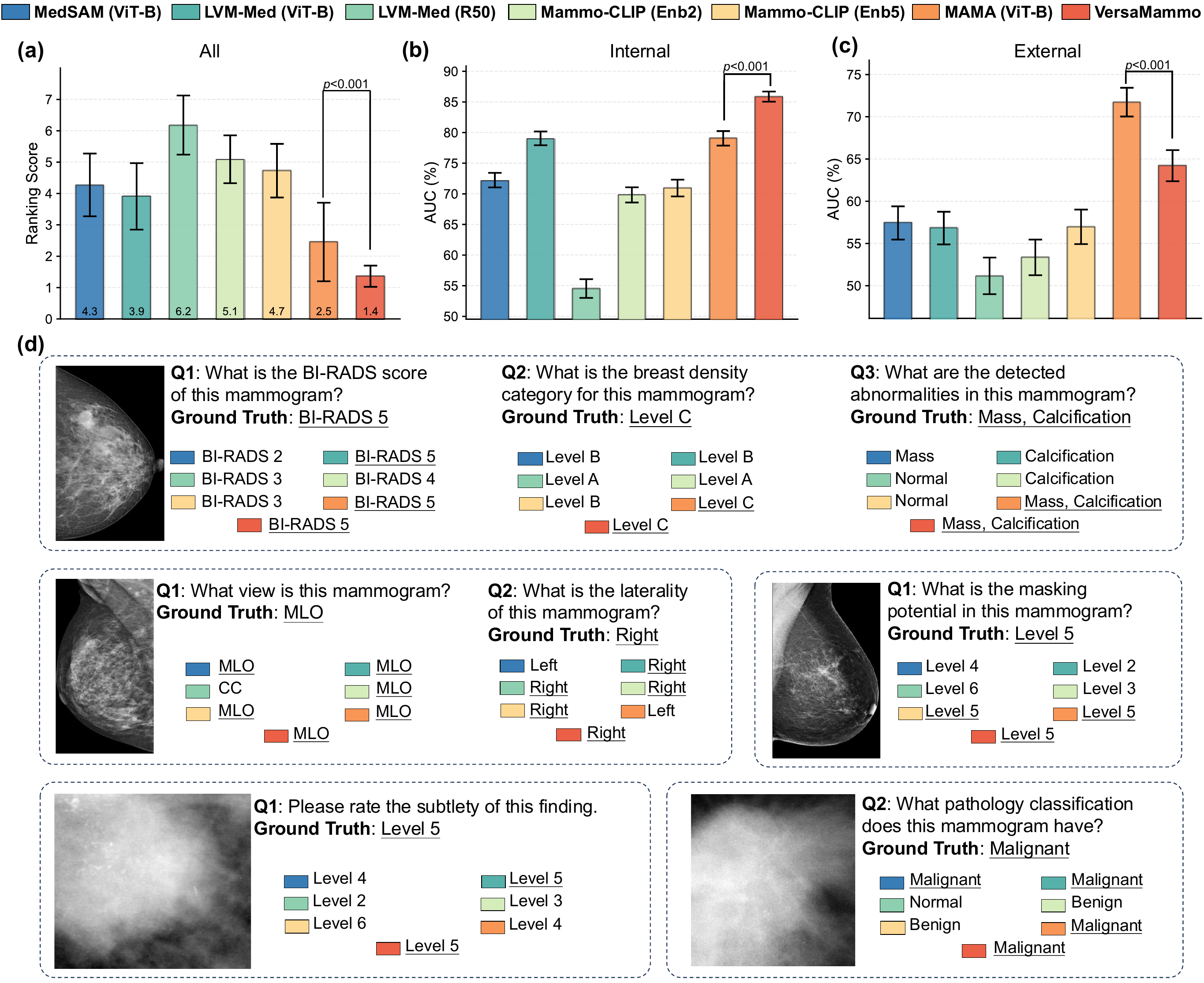}
\caption{\textbf{Performance of FMs on VQA tasks.} (a) Average rank of FMs on VQA tasks based on AUC across internal and external validation {experiments}. (b-c) AUC for internal and external validation {experiments}. The Wilcoxon signed-rank two-side test is employed for data analysis (1000 bootstrap replicates). Error bars represent 95\% CI. The centre indicates mean. For the internal validation , n = 21,033 independent cases.
For the external validation, n = 26,933 independent cases. (d) Qualitative results for mammogram VQA.}
\label{fig:vqa}
\end{figure*}

We conducted internal and external VQA evaluations at high resolution ($512\times512$) following the MammoVQA benchmark \cite{zhu2025benchmark}, which comprises 15 public datasets (11 internal, 4 external). Performance was measured using AUC, BACC, and macro-F1 across diverse question types. For internal evaluation, models were fine-tuned on the training split of the internal benchmark and evaluated on the test split. VersaMammo achieved the highest overall performance with an average rank of 1.4 (Fig.~\ref{fig:vqa}(a)), demonstrating strong and balanced capability across question types.

For external validation, all models were evaluated on the four external datasets unseen during both pre-training and fine-tuning. In this setting, MAMA (ViT-B) achieved the best performance. This gap, where VersaMammo substantially underperformed MAMA (ViT-B), highlights the impact of pre-training strategies: VersaMammo excels in unimodal tasks (e.g., classification, retrieval) but lacks large-scale image-text pre-training, which limits its aligned vision-language representations needed for complex VQA reasoning, a capability likely strengthened by MAMA's vision-language pre-training. Detailed external results are in Supplementary Tables 21-22.

\begin{figure*}
\centering
\includegraphics[width=0.95\textwidth]{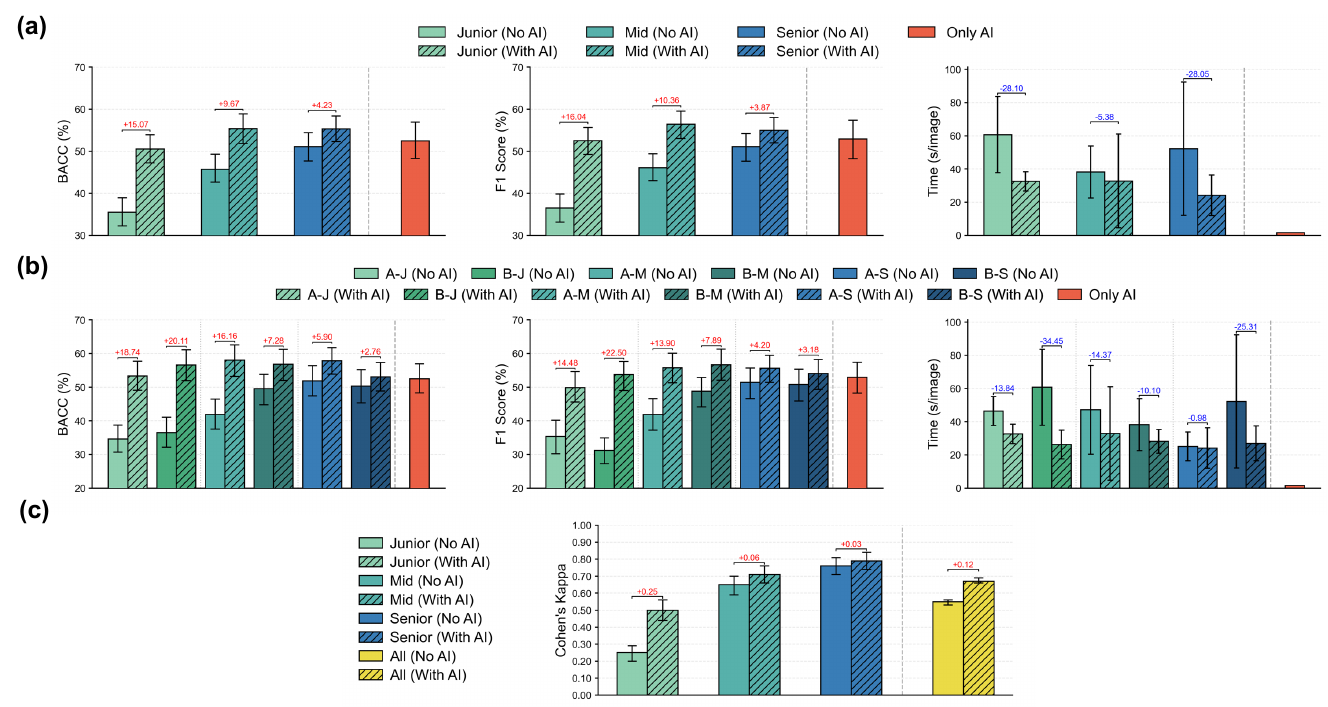}
    \caption{\textbf{Analysis of the reader study}. (a) represents the BACC, macro-F1 score, and average image reading time for each group of radiologists with different experience levels. (b) represents the BACC, macro-F1 score, and average reading time for each individual radiologist. $A$ and $B$ denote two balanced experimental groups of radiologists, while $J$, $M$, and $S$ represent Junior, Mid-level, and Senior, respectively. (c) displays the mean pairwise Cohen's Kappa coefficients for each group of radiologists with different experience levels and all individual radiologists combined. The Wilcoxon signed-rank two-side test is employed for data analysis (1000 bootstrap replicates). Error bars in (a-c) represent the 95\% confidence intervals for performance metrics and the standard deviation for reading time. The centre indicates mean. n = 500 independent cases.
    }
\label{fig:human_ai_comprehensive}
\end{figure*}
\subsection{Subgroup Analysis}
We compared VersaMammo with other FMs for BI-RADS assessment on EMBED across four dimensions: breast composition, age, clinical scenario, and ethnicity (Extended Data Fig. 3). VersaMammo achieved superior overall performance, leading AUC of 74.46\% (2.44\% above second-best) and ranked first in 17 of 18 subgroups, with particular strength in high-density cases and complex diagnostic scenarios. Performance gaps persist in certain racial and younger age groups, suggesting future fairness improvements via increased representation. 
\textbf{Breast composition:} Performance declines with increasing density, consistent with clinical observation; VersaMammo maintains the highest relative performance across all density levels (Extended Data Fig. 3 (a) and Supplementary Table 23).
\textbf{Age:} VersaMammo performs highest in the 70+ group, declining progressively in younger cohorts, with the largest drop in under‑40 due to limited data and atypical presentation (Extended Data Fig. 3 (b) and Supplementary Table 24).
\textbf{Clinical scenario:} Performance is slightly higher in screening than diagnostic cases, but VersaMammo's advantage is most pronounced in diagnostic cases (more complex lesions), indicating stronger utility in difficult scenarios (Extended Data Fig. 3 (c) and Supplementary Table 25).
\textbf{Ethnicity:} Performance varies across racial groups, with declines for Native Hawaiian/Pacific Islander and Asian participants. VersaMammo ranks first in nearly all subgroups, showing relative robustness (Extended Data Fig. 3 (d) and Supplementary Table 26).

{To assess whether its strong overall performance might mask underlying disparities, we conducted a dedicated fairness analysis using standard metrics (Supplementary Table 27): true positive rate gap (TPR Gap), false positive rate gap (FPR Gap), equalized odds difference (EOD), and average odds difference (AOD). VersaMammo maintained relatively low subgroup disparities on all metrics while achieving superior diagnostic accuracy. By contrast, LVM-Med (R50) showed lower disparity metrics but substantially weaker classification performance, whereas Mammo-CLIP (Enb5) and MedSAM exhibited larger subgroup disparities.}

In summary, VersaMammo demonstrates leading robust performance, particularly in high‑density and complex diagnostic cases, supporting strong clinical deployment potential.

\subsection{Ablation Study}
\label{subsec:ablation_study}
To comprehensively evaluate the contributions of key components in VersaMammo, we conducted systematic ablations across 10 experiments: classification tasks on RSNA and PDS1, detection on INbreast, and segmentation on INbreast. Key components were sequentially removed or modified:
\begin{itemize}
    \item \textit{Task-specific} represents that the model is trained from scratch for a specific task.
    \item\textit{w/o Mammogram} represents the FM pre-trained on the ImageNet dataset \cite{deng2009imagenet}, without any subsequent pre-training on mammogram data.
    \item\textit{w/o Stage 2} represents the omission of the second stage of pre-training VersaMammo.
    \item\textit{w/o Distillation} represents the removal of the knowledge distillation component in the second stage of pre-training VersaMammo.
    \item\textit{w/o Supervision} represents the elimination of the supervision component in the second stage of pre-training VersaMammo.
    \item \textit{w/o CNN} represents replacing the Enb5 backbone of VersaMammo with ViT-B in stage 2, using ViT backbone in both pre-training stages to validate the necessity of CNN in local texture feature modeling.
\end{itemize}

Results are in Extended Data Fig. 4 and Supplementary Tables 28-29. \textbf{1) Mammogram pre-training is fundamental for improving performance in mammogram analysis.} Omitting it causes substantial performance drops: classification AUC drops from 85.59\% to 54.20\% (‑31.39\%) on RSNA and 84.22\% to 49.76\% (‑34.46\%) on PDS1; segmentation Dice from 74.53\% to 68.97\%; detection IoU from 71.20\% to 50.10\%. Here, the larger drop of classification on held-out PDS1 suggests genuine domain knowledge transfer rather than dataset memorization. \textbf{2) Stage 2 multi-task pre-training improves generalization.} Removing it reduces segmentation Dice from 74.53\% to 60.96\% and detection IoU from 71.20\% to 40.96\%, as stage 1 lacks local structure sensitivity and stage 2 adds CNN-based multi-task learning.
\textbf{3) Distillation and supervision offer complementary task-dependent benefits.} Using only one component in stage 2 degrades classification performance (e.g., RSNA AUC drops 9.75\% and 7.98\%, respectively).
\textbf{4) CNN backbone (EfficientNet-B5) outperforms ViT-B for mammogram analysis.} Replacing Enb5 with ViT-B reduces classification AUC to 77.91\% (RSNA) and 69.56\% (PDS1), and segmentation IoU to 64.38\%, indicating convolutional inductive biases benefit local texture extraction.
\textbf{5) Unified framework surpasses task-specific alternatives.} VersaMammo raises classification AUC from 69.12\% to 85.59\% (RSNA) and 56.94\% to 84.22\% (PDS1), and detection IoU from 62.86\% to 71.20\%, validating multi-stage multi-task pre-training as a robust foundation with exceptional data efficiency on small datasets like PDS1.

\textbf{Parameter Sensitivity Analysis.} On VinDr-Mammo validation (Extended Data Fig. 4 (m-n)), BI-RADS classification is sensitive to loss weights \(\lambda_1,\lambda_2\) (AUC 74.51\%–76.65\% over \{0.01,0.1,1,10,100\}), while breast composition is robust (91.97\%–92.62\%). Increasing pre-training sample size steadily improves AUC, supporting scaling benefits (Extended Data Fig. 4 (o-p) and Supplementary Table 30).

\subsection{Reader Study}
\label{subsec:results_human_machine}
We evaluated six radiologists (junior: 1–5 years; mid-level: 6–10 years; senior: >10 years) on BI‑RADS assessment (standardized decision-critical framework) with and without AI assistance (VersaMammo), plus standalone AI (see section \ref{meth}). Results are presented in Fig.~\ref{fig:human_ai_comprehensive} and Supplementary Tables 31-33.
\textbf{1) Standalone AI matches senior radiologists.} Its macro‑F1 of 52.94\% exceeds senior radiologists without AI (51.10\%) by 1.84\%, and outperforms mid‑level (46.07\%) and junior radiologists (36.47\%) (Fig.~\ref{fig:human_ai_comprehensive} (a,b)).
\textbf{2) AI assistance consistently improves accuracy.} All groups improve with AI. Juniors show the largest gain: macro‑F1 from 36.47\% to 52.51\%, balanced accuracy from 35.48\% to 50.55\%, reaching senior-like levels. The Macro‑F1 of mid‑level and senior readers improved from 46.07\% to 56.43\% and from 51.10\% to 54.97\%, respectively, with both surpassing unassisted seniors and standalone AI, confirming the superiority of human‑AI collaboration. 
\textbf{3) AI assistance enhances efficiency.} Reading average time drops substantially with AI: juniors from 60.70s to 32.60s; mid‑level from 38.20s to 32.82s; seniors from 52.18s to 24.13s. Standard deviations also decreased, indicating more consistent reading efficiency with AI, especially for junior readers (from 22.95s to 5.79s).
\textbf{4) AI assistance improves diagnostic consistency.} Within same experience level, consistency ($\kappa$) increases: junior readers from 0.25 to 0.50; mid‑level readers from 0.65 to 0.71; senior readers from 0.76 to 0.79. Cross‑level consistency rises from 0.55 to 0.67 (Fig.~\ref{fig:human_ai_comprehensive} (c))

\textbf{Clinical implications.} VersaMammo provides robust standalone performance and effectively augments radiologists across experience levels, enabling junior and mid‑level readers to achieve senior‑level accuracy while improving efficiency and consistency, supporting its role as both an assistive tool and quality benchmark.

\section{Discussion}
\label{sec:discussion}

\subsection{Advantages and Strengths}
VersaMammo is trained on the largest multi‑institutional mammography dataset to date. Its two‑stage pre‑training combines self‑supervised learning with supervised knowledge distillation, transferring knowledge from a ViT teacher to a lightweight model while integrating clinical labels. We validated the model on 68 internal and 24 external experiments across five task types, covering multinational public datasets and private datasets. VersaMammo ranked first in 50 internal and 20 external experiments, demonstrating strong generalization.
In the reader study, VersaMammo alone matched senior radiologists in diagnostic performance. It improved accuracy, efficiency, and inter‑reader consistency across all experience levels, highlighting its value as an assistive tool that reduces human variance in BI‑RADS assessment.

\subsection{Limitations and Future Directions}
Several limitations inform future work. While external validation included data from multiple countries, the core PDS benchmarks remain predominantly Chinese, leaving global performance unevaluated. Future research should prioritize large-scale multi-regional collaborations to assess performance and ensure equitable applicability across healthcare systems and demographics. Domain shifts from scanner vendors, acquisition parameters, and patient populations remain challenging; test-time adaptation, domain-invariant learning, and federated fine-tuning offer promising avenues.

The technical scope is currently unimodal (mammogram only), whereas clinical decisions often integrate multiple sources. Future iterations should incorporate patient history, radiology reports, EHR data, and complementary imaging (ultrasound, MRI); vision-language models exemplify this potential. The ablation study shows mammogram-specific pre-training is the primary performance driver, but disentangling its contribution from dataset scale and distillation architecture remains future work. Grad-CAM visualizations may not fully align with CNN subtleties in detecting micro-calcifications. In the reader study, the one-month washout period follows standard practice, though a minor recall bias risk exists for highly distinctive cases.

VersaMammo is a promising step toward clinical translation, but further efforts in regulatory alignment, system integration, and responsible governance are needed. With regulatory review for safety and effectiveness, standardized PACS integration, radiologist-in-the-loop design, explainability mechanisms, and post-deployment monitoring, VersaMammo could advance breast cancer screening while upholding safety, equity, and accountability.

\subsection{Clinical Integration Pathways}
To enhance translational relevance, we outline clinical integration pathways. As a triage tool, VersaMammo could prioritize urgent cases, reducing diagnosis time for high-risk patients. As a concurrent second reader, it could provide real-time decision support by highlighting regions of interest and malignancy scores alongside radiologist review, requiring low-latency inference and an intuitive interface, though it risks automation bias. As an independent reader for quality assurance, it could retrospectively analyze studies, flagging discrepancies for secondary review, demanding robust audit workflows. For automated preliminary reporting, it could generate draft reports for standardized screening cases, improving throughput, requiring high specificity and reliability, along with adapted legal frameworks. Each pathway requires careful consideration of implementation requirements, expected efficiency gains, and potential failure modes to ensure effective clinical integration.

\section{Methods}\label{meth}
\subsection{Data Collection and Processing}\label{dataset}

\textbf{Data collection and annotation.} 
To support the development of VersaMammo and evaluate existing FMs, we curate the largest and most diverse mammogram dataset to date, consisting of 679,911 mammograms from 21 datasets, as shown in Fig. \ref{fig:versamammo} (a). 
Specifically, we collected a total of 603,665 mammograms from 16 public datasets, the details and acquisition paths of which are provided in Extended Data Table 3. The private data were collected from five institutes across five years, namely private dataset 1 (PDS1, Shenzhen People's Hospital), private dataset 2 (PDS2, Guangzhou First People's Hospital), private dataset 3 (PDS3, The Third Affiliated Hospital, Sun Yat-sen University), private dataset 4 (PDS4, Sun Yat-sen Memorial Hospital, Sun Yat-sen University), and private dataset 5 (PDS5, Yunnan Cancer Hospital), with a total of 76,246 mammograms.
For public data, we adopted the labels released by the original authors. For private data, all labels were first assigned by two mid-level radiologists, and any disagreements were resolved by a senior radiologist.
More data details are presented in Supplementary Tables 34-49.

\textbf{Pre-processing.} Following Mammo-CLIP \cite{mammoclip}, we applied a rule-based ROI detection: set values below 40 to 0 and removed constant rows/columns (background). Images were resized to $224\times224$, $512\times512$, or $518\times518$ to evaluate different FMs across resolutions. All models shared identical preprocessing and augmentation for consistent evaluation.

\textbf{Data Split.} To prevent leakage and ensure reliability, we maintained consistent splits per dataset (typically 7:1:2 at patient level, or image level when patient info unavailable). Pre-training used the training splits of EMBED, RSNA, and VinDr-Mammo; their test splits were used for downstream evaluation. All other datasets were strictly reserved for downstream tasks and unseen during pre-training (Supplementary Table 35).

\subsection{Model Design and Pre-training for VersaMammo}
VersaMammo is pre-trained using a two-stage strategy that synergistically combines self-supervised learning on unlabeled data with supervised distillation of clinical {annotations}, thereby enhancing both global contextual understanding and local feature perception for diverse downstream tasks (Fig. \ref{fig:versamammo} (c)).

\textbf{Stage 1: Unsupervised Pre-training with Self-Supervised Learning.}
To build a robust foundation model, we first pre-trained a Vision Transformer (ViT) on 389,100 unlabeled mammograms from a combined training split of EMBED, RSNA, and VinDr-Mammo. This stage uses a DINOv2-style \cite{oquab2023dinov2} self-supervised approach, i.e., self-distillation with multi-crop augmentation, to learn generalizable visual representations without relying on specific annotation protocols. This approach, which integrates contrastive learning with masked image modeling, is designed to encourage the model to focus on invariant features and to be robust to common mammography variations, such as differences in contrast, intensity, and scanner or protocol styles.

\textbf{Stage 2: Dual-Head Supervised Learning with Knowledge Distillation.}
While the ViT teacher model excels at capturing global context, its computational demands can be prohibitive for real-world deployment. Conversely, Convolutional Neural Networks (CNNs) offer a favorable balance of strong local feature extraction and operational efficiency.  For mammography, this global-local trade-off is essential: breast-wide structural patterns and composition provide context for interpretation, whereas key malignancy cues can be extremely small and high-frequency (e.g., micro-calcifications), which benefit from CNN locality bias. In clinical practice, physicians need to analyze high-resolution mammograms, integrating global contextual information and local lesion details to formulate diagnostic conclusions and treatment recommendations. Therefore, we propose a dual-head supervised learning with a knowledge distillation method to transfer the acquired general knowledge from the ViT into a more practical architecture that is better suited to full-resolution mammogram deployment while simultaneously infusing it with expert clinical supervision (guideline-grounded labels). We achieve this through a supervised knowledge distillation framework \cite{xu2023bilateral}, using an EfficientNet-b5 (Enb5) as the student model (VersaMammo). As shown in Fig. \ref{fig:versamammo} (c), VersaMammo is trained not only on ground-truth labels, including BI-RADS classifications and breast composition, to absorb clinical supervision but is also guided by the feature representations of the ViT teacher. The more detailed model architecture is provided in Extended Data Fig. 2.

To enhance VersaMammo's robustness to varying image resolutions, we introduce a contrastive learning loss that encourages consistent feature representations across different resolution scales. This design targets a mammography-specific practical issue: images are acquired or stored at different pixel spacings, and readers routinely zoom across scales, so models should remain stable under resolution changes. 
The overall training objective integrates multiple optimization goals through a composite loss function that combines contrastive learning, knowledge distillation, and dual-head clinical supervision, using the training splits of EMBED, RSNA, and VinDr-Mammo (365,428 images with labels). 
\begin{equation}
\mathcal{L}_{{total}} = \mathcal{L}_{{cont}} + \lambda_1 \mathcal{L}_{{dist}} + \lambda_2 \mathcal{L}_{{dhsup}},
\label{eq:total_loss}
\end{equation}
where the weighting coefficients $\lambda_1$ and $\lambda_2$ balance the contributions of the contrastive learning, distillation loss, and the dual‑head supervised loss. $\mathcal{L}_{{cont}}$ is a contrastive loss implemented following the InfoNCE framework \cite{oord2018representation}. 
\begin{equation}
\mathcal{L}_{{cont}} = -\frac{1}{N} \sum_{i=1}^{N} \log \frac{exp((k(z_i^S,z_j^{S+})/\tau) }{\sum_{j=1}^{N}exp((k(z_i^S,z_j^{S-})/\tau)},
\label{eq:contrastive_loss}
\end{equation}
where $z_j^{S+}$ and $z_j^{S-}$ represent the positive and negative feature samples extracted from low-resolution images, respectively. $\tau$ is the temperature factor. $N$ denotes the number of samples within a mini-batch used for contrastive learning.
The knowledge distillation loss $\mathcal{L}_{{dist}}$ transfers rich feature representations from the pre-trained ViT teacher to the Enb5 student through Kullback-Leibler divergence:

\begin{equation}
\begin{split}
\mathcal{L}_{{dist}} &= D_{{KL}}(P_T \| P_S) \\
   & = \frac{1}{N} \sum_{i=1}^{N} \sum_{j=1}^{C} P_T^{(j)}(\mathbf{z}_i^T) \log\left( \frac{P_T^{(j)}(\mathbf{z}_i^T)}{P_S^{(j)}(\mathbf{z}_i^S)} \right),
\end{split}
\label{eq:kl_loss}
\end{equation}
where $p_T(\mathbf{z}_i^T) = {softmax}(\mathbf{z}_i^T)$ and $p_S(\mathbf{z}_i^S) = {softmax}(\mathbf{z}_i^S)$ are feature probability distributions, 
with $\mathbf{z}_i^T$ and $\mathbf{z}_i^S$ denoting feature vectors from teacher and student models, respectively. $C$ denotes the number of classes.
The dual-head supervised loss $\mathcal{L}_{{dhsup}}$ combines two cross-entropy losses:
\begin{equation}
\mathcal{L}_{{dhsup}} = \mathcal{L}_{ce}^{conf},  \quad conf \in \{\text{BI-RADS, composition}\},
\end{equation}
where $\mathcal{L}_{ce}^{conf}$ is a standard cross-entropy loss for BI-RADS classifications and breast composition, respectively:
\begin{equation}
\begin{split}
\mathcal{L}_{ce}^{conf} = -\frac{1}{N} \sum_{i=1}^{N} \sum_{c=1}^{C_{conf}} y_{i,c}^{conf} \log(\hat{y}_{i,c}^{conf}), \\
conf \in \{\text{BI-RADS, composition}\}.
\end{split}
\label{eq:ce_loss}
\end{equation}
Here, $y_{i,c}^{conf}$ is the ground-truth label (one-hot encoded) and $\hat{y}_{i,c}^{conf}$ is the predicted probability for class $c$ of the $i^{th}$ sample, with $C_{conf}$ denoting the number of classes for each configuration. We use AdamW \cite{loshchilov2017decoupled} as the optimizer. 

\textbf{Hyperparameter Selection.}
The weighting coefficients $\lambda_1$ and $\lambda_2$ were optimized via a grid search over the set $\{0.01, 0.1, 1, 10, 100\}$ for both parameters. The combination $\lambda_1 = 0.1$ and $\lambda_2 = 10$ achieved the best average performance on VinDr-Mammo, striking an effective balance between knowledge distillation loss, dual-head supervised loss, and contrastive loss. See Extended Data Fig. 4 (m–n) for details. The temperature factor $\tau$ is empirically set to 0.07 following the practice in \cite{huang2024dynamic}.

This two-stage pre-training paradigm equips VersaMammo with a unique combination of broad visual knowledge and targeted clinical understanding. As a result, the model demonstrates remarkable versatility and state-of-the-art performance across a wide spectrum of tasks, paving the way for powerful and adaptable AI tools in breast cancer screening.

\subsection{Reader Study}
\label{subsec:human_machine}

To evaluate the clinical utility of the AI model, we conducted a controlled, retrospective reader study. The study cohort consisted of six board-certified radiologists, who were stratified by clinical experience into three categories: junior (1–5 years), mid-level (6–10 years), and senior ($>$10 years). Participants were allocated into two balanced groups (Group A and Group B) to minimize the order effects, each containing one radiologist from each experience level. 

\textbf{Study Design and Blinding:}
A randomized crossover design was implemented. The study comprised two reading phases, separated by a mandatory one-month washout period to mitigate recall bias. In Phase 1, Group A interpreted cases without AI assistance, while Group B used AI assistance. In Phase 2, the conditions were switched. Throughout the study, all radiologists were blinded to the ground truth, final patient outcomes, and the assessments of other readers.

\textbf{Case Selection and AI Interface:}
A total of 500 mammograms were randomly selected from the held-out test splits, with a 1:1 ratio between internal and external cases. This evaluation set had no overlap with any data used during model training or development. For the AI-assisted condition, a dedicated interface presented the original mammogram alongside the model's comprehensive outputs, including lesion detection bounding boxes, segmentation masks, and predicted BI-RADS categories with confidence scores. All participants completed a standardized training session on the study protocol and interface prior to commencing the study.

\section{Data Availability}
The training and validation datasets in the public domain used in this study are available and can be downloaded via the links provided in Extended Data Table 3. For the data from PDS1 to PDS5, these datasets are not publicly available due to patient privacy obligations, institutional review board requirements, and data use agreements. However, researchers interested in accessing deidentified data may submit a reasonable request directly to the corresponding authors, subject to obtaining the necessary ethical approvals and complying with institutional policies.

\section{Code Availability}
The code and weights of the VersaMammo have been made available on GitHub (\url{https://github.com/fuxianghuang1/VersaMammo}).

\section{Acknowledgements}
We acknowledge Shenzhen People's Hospital, Guangzhou First People's Hospital, The Third Affiliated Hospital of Sun Yat-sen University, Sun Yat-sen Memorial Hospital of Sun Yat-sen University, and Yunnan Cancer Hospital for providing research samples.

\section{Funding}
This work was supported by the National Key R\&D Program of China (Project No. 2023YFE0204000), the Hong Kong Innovation and Technology Commission (Project No. MHP/002/22 and ITCPD/17-9), the National Natural Science Foundation of China (Project No. 92570110 and 82001986), the Outstanding Youth Science Foundation of Yunnan Basic Research Project (Project No. 202401AY070001-316), the Innovative Research Team of Yunnan Province (Project No. 202505AS350013), and Shenzhen Science and Technology Innovation Committee Fund (Project No. KCXFZ20230731094059008).
The funders had no role in study design, data collection and analysis, decision to publish or preparation of the manuscript.
\section{Author Contributions}
F.H. and H.C. conceived and designed the study. F.H. collected all public datasets, developed the technical framework, performed model pre-training, and evaluated downstream tasks. J.Z. assisted in dataset collection and conducted evaluations of multiple existing FMs on downstream tasks. Y.Y., Y.X., Y.G., Q.K., M.W., Y.T., Z.H., and L.M. (medical professionals) collected and annotated private clinical datasets from hospitals. J.M., Z.L., Z.X., Z.C., X.W., J.H., X.J., S.Y., Y.B., and L.Z. provided critical feedback on the manuscript and experimental design. F.H. wrote the manuscript with input from all authors. X.W., Q.L., Z.L., H.Y., and H.C. supervised the research. All authors reviewed and approved the final manuscript.

\section{Competing Interests}
The authors declare no competing interests.

\section{Ethics Declarations}
This project has been reviewed and approved by the Human and Artefacts Research Ethics Committee (HAREC). The protocol number is HREP-2025-0025.

\clearpage


\section*{Extended Data}
\captionsetup[table]{name=Extended Data Table}
\begin{figure*}
\centering
\includegraphics[width=0.95\textwidth]{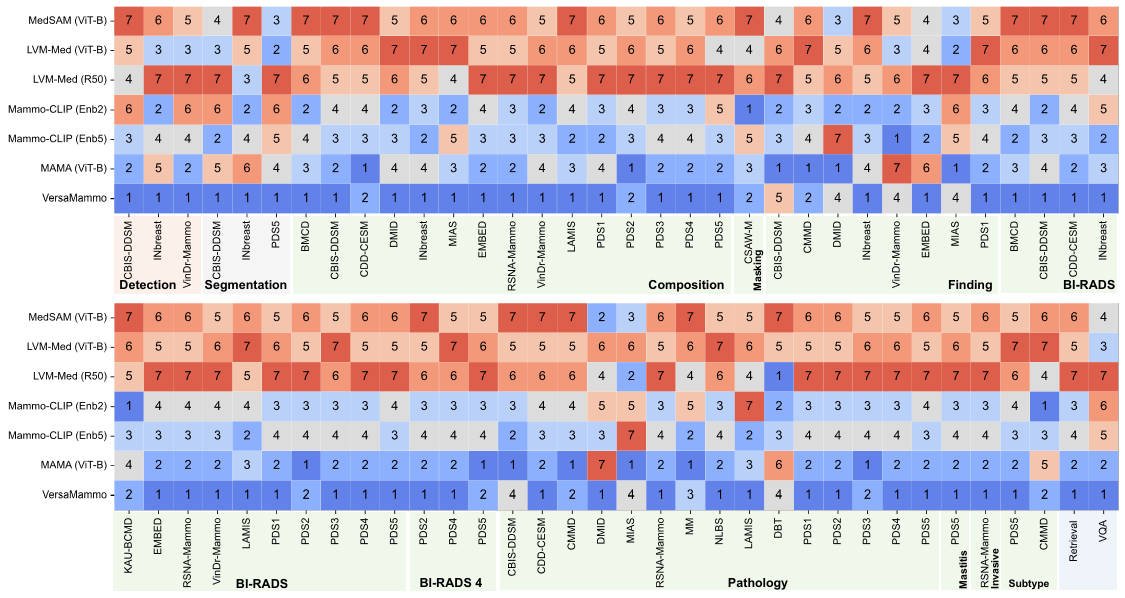}
\caption{\textbf{Ranks of 68 internal validation {{experiments}}. }}
\label{fig:extend_rank_results}
\end{figure*}

\begin{figure*}
\centering
\includegraphics[width=0.95\textwidth]{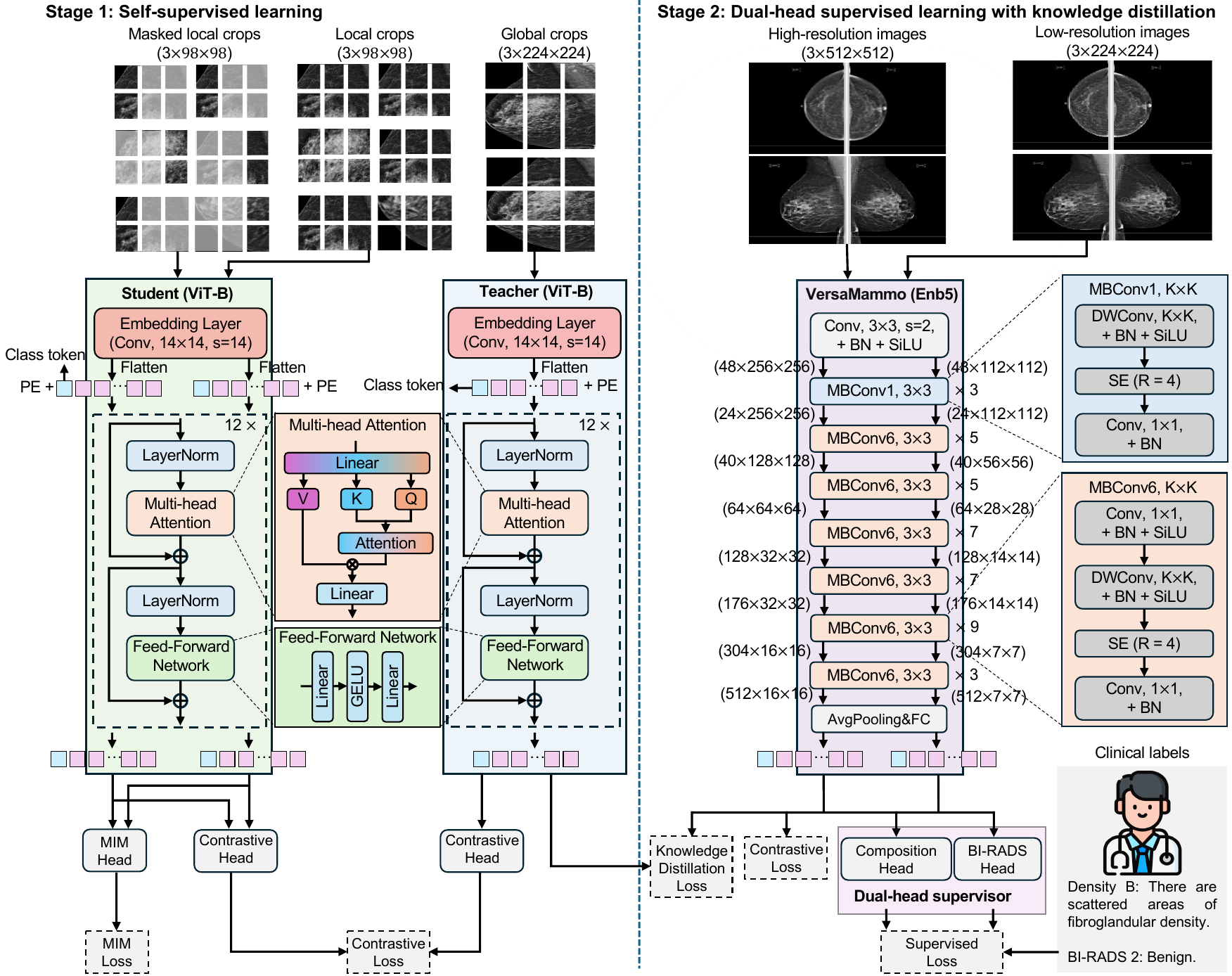}
\caption{\textbf{Detail architecture of VersaMammo}. In the first stage, a teacher model is trained using self-supervised learning, involving MIM loss and contrastive loss. 
The second stage utilizes a combination of losses, including knowledge distillation loss, dual-head supervised loss, and contrastive loss, integrating both teacher model knowledge and clinical supervision (guideline-grounded labels). 
}
\label{fig:detail_architecture}
\end{figure*}

\begin{figure*}
\centering
\includegraphics[width=0.95\textwidth]{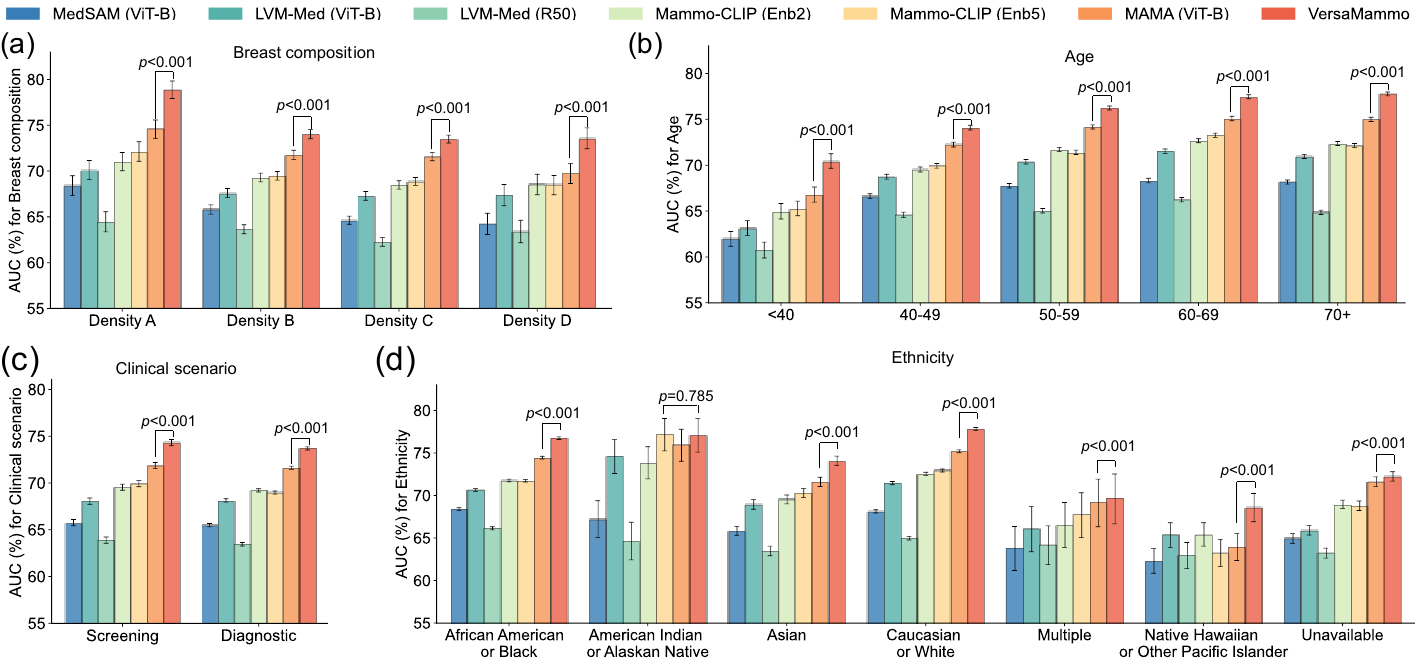}
\caption{\textbf{Performance of FMs on subgroup analysis.} (a-d) correspond to analyses of breast composition, age, clinical scenario, and {ethnicity}, respectively. The Wilcoxon signed-rank two-side test is employed for data analysis (1000 bootstrap replicates). Error bars represent 95\% CI. The centre indicates mean. n = 94,158 independent cases.
}
\label{fig:dimension_analysis}
\end{figure*}

\begin{figure*}
\centering
\includegraphics[width=0.95\textwidth]{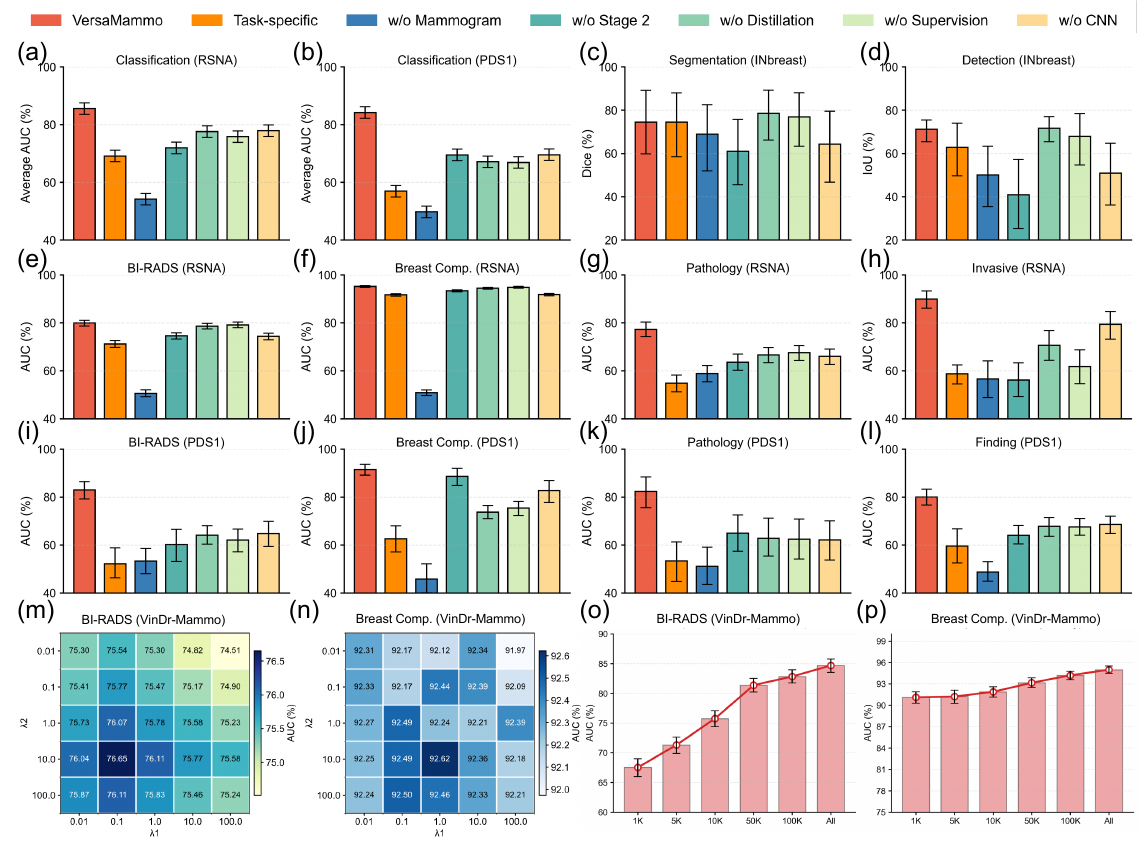}
    \caption{
    \textbf{The effectiveness of different components of the VersaMammo across 10 {{experiments}}.} (a) Average AUC of 4 classification {{experiments}} on RSNA. (b) Average AUC of 4 classification {{experiments}} on PDS1. (c) IoU of the detection task on INbreast (Supplementary Table 29). n = 19 independent cases. (d) Dice of the segmentation task on INbreast (Supplementary Table 29). n = 19 independent cases. (e-l) AUC of the 4 classification {{experiments}} on RSNA and PDS1, respectively (Supplementary Table 28). n = 10,908 (RSNA), 319 (PDS1) independent cases. (m-n) Sensitivity analysis of the loss weights $\lambda_1$ and $\lambda_2$. (o-p) Sensitivity analysis of the pre-training sample size (Supplementary Table 30). n = 4,000 independent cases. The Wilcoxon signed-rank two-side test is employed for data analysis (1000 bootstrap replicates). Error bars represent 95\% CI. The centre indicates mean.} 
    \label{fig:ablation}
\end{figure*}

\clearpage

\onecolumn{
\begin{table*}
\setcounter{table}{0}
\tiny
\centering
\caption{Specific label descriptions used for the classification task, where the ground truths are annotated by radiologists.}
\label{tab:label1}

\renewcommand{\arraystretch}{1.25}
\setlist[itemize]{leftmargin=*, nosep, topsep=0pt, partopsep=0pt, after=\vspace{-\baselineskip}}

\begin{tabularx}{\textwidth}{
  >{\raggedright\arraybackslash}p{0.15\textwidth}
  |>{\raggedright\arraybackslash}p{0.25\textwidth}
  |>{\raggedright\arraybackslash}p{0.5\textwidth}
}
\hline
\textbf{Configuration} & \textbf{Description} & \textbf{Category description} \\ \hline

Composition: breast composition assessment \cite{nguyen2023vindr, rsna2023challenge, jeong2023emory, oza2024digital, suckling1994mammographic} & There are two types of classification for breast composition as follows.
\begin{itemize}
  \item Breast density \cite{nguyen2023vindr, rsna2023challenge, jeong2023emory}: The proportion of glandular and fibrous tissue to fatty tissue in the breast.
  \item Background tissue \cite{oza2024digital, suckling1994mammographic}: The type of breast tissue visible on mammograms.
\end{itemize}
&
Breast density
\begin{itemize}
  \item Density A: The breast is almost entirely fatty.
  \item Density B: There are scattered areas of fibroglandular density.
  \item Density C: The breasts are heterogeneously dense, which may obscure small masses.
  \item Density D: The breasts are extremely dense, which lowers the sensitivity of mammography. 
  \item[] 
\end{itemize} 

Background tissue
\begin{itemize}
  \item Fatty-glandular: Indicates a mix of fatty and fibrous tissues.
  \item Fatty: Majority of the breast tissue is fat, which may make it easier to spot abnormalities.
  \item Dense-glandular: Indicates high density of glandular tissue, which can sometimes make it hard to detect tumors due to less fat.
  \item
\end{itemize}
\\ \hline

Masking: masking potential risk evaluation \cite{sorkhei2021csaw}& The possibility for cancer to be obscured in a mammogram. &
\begin{itemize}
  \item Level 1 to Level 8: Lowest potential to be obscured to the highest potential to be obscured.
\end{itemize}
\\ \hline

Finding: abnormality identification \cite{jeong2023emory}& 
Any unusual or atypical finding in the breast tissue that differs from the normal appearance. & 
\begin{itemize}
  \item Normal: No abnormalities.
  \item Calcification: Small calcium deposits.
  \item Mass: A space-occupying lesion.
  \item Architectural distortion: Distortion of normal breast structure.
  \item Asymmetry: Unequal tissue density.
  \item Miscellaneous: Other rare findings.
  \item Nipple retraction: Inward-turning nipple.
  \item Suspicious lymph node: Unusual lymph nodes.
  \item Skin thickening: Thickened breast skin.
  \item Skin retraction: Skin pulling inward.
  \item
\end{itemize}
\\ \hline

BI-RADS : BI-RADS  assessment \cite{alsolami2021king, nguyen2023vindr, rsna2023challenge, jeong2023emory, imane2024lamis}& 
A system used to categorize the findings on mammograms into levels that describe the risk of cancer. & 
\begin{itemize}
  \item BI-RADS  0: Incomplete (Need additional imaging evaluation).
  \item BI-RADS  1: Negative.
  \item BI-RADS  2: Benign finding.
  \item BI-RADS  3: Probably benign finding.
  \item BI-RADS  4: Suspicious malignancy.
  \item BI-RADS  5: Highly suggestive of malignancy.
  \item
  \end{itemize}
\\ \hline

BI-RADS  4: BI-RADS  4 subclassification \cite{alsolami2021king} &
A suspicious finding with a moderate risk of malignancy (2–95\%), requiring biopsy for definitive diagnosis. &
\begin{itemize}
  \item BI-RADS  4A: Low suspicion (2–10\% malignancy risk).
  \item BI-RADS  4B: Mid-level suspicion (10–50\% malignancy risk).
  \item BI-RADS  4C: High suspicion (50–95\% malignancy risk).
  \item
\end{itemize}
\\ \hline
\end{tabularx}
\end{table*}

\begin{table*}
\setcounter{table}{1}
\tiny
\centering
\caption{Specific label descriptions used for the classification task, where the ground truths come from additional biopsy or surgical excision.}
\label{tab:label2}
\begin{tabularx}{\textwidth}{
  >{\raggedright\arraybackslash}p{0.15\textwidth}
  |>{\raggedright\arraybackslash}p{0.25\textwidth}
  |>{\raggedright\arraybackslash}X
}
\hline
\textbf{Configuration} & \textbf{Description} & \textbf{Category description} \\ \hline

Pathology: pathological diagnosis prediction \cite{lee2017curated, oza2024digital, suckling1994mammographic, khaled2022categorized}&
The definitive tissue diagnosis from biopsy or surgical excision. &
\begin{itemize}
  \item Malignant: Cancerous through biopsy confirmation.
  \item Benign: Non-cancerous through biopsy confirmation.
\end{itemize}
\\ \hline

Mastitis: mastitis differentiation \cite{chang2019risk} &
Differentiation between inflammatory breast conditions and cancerous lesions. &
\begin{itemize}
  \item Mastitis: Benign inflammatory condition.
  \item Malignancy: Cancerous through biopsy confirmation.
\end{itemize}
\\ \hline

Invasive: invasive status determination \cite{rsna2023challenge} &
Classification based on whether cancer cells have spread beyond the ducts or lobules into the surrounding breast tissue. &
\begin{itemize}
  \item Non-Invasive: Cancer cells are confined within the ducts or lobules and have not invaded surrounding tissues (e.g., Ductal Carcinoma In Situ (DCIS), Lobular Carcinoma In Situ (LCIS)). 
  \item Invasive: Cancer cells have penetrated the surrounding breast tissue, indicating a more aggressive form of cancer (e.g., Invasive Ductal Carcinoma (IDC), Invasive Lobular Carcinoma (ILC)).
\end{itemize}
\\ \hline

Subtype: molecular subtype prediction \cite{cancer2012comprehensive, tsang2020molecular, fragomeni2018molecular} &
Categorized by receptor status and gene expression. The ground truths come from additional surgical excision. &
\begin{itemize}
  \item Luminal A: Characterized by positive estrogen (ER+) and progesterone (PR+) receptors, HER2 negative, and low proliferation index, typically associated with a good prognosis and responsiveness to hormonal therapy.
  \item Luminal B: Defined by positive ER, variable PR status, and either HER2 positive or negative, with a high proliferation index, usually associated with a poorer prognosis and may require chemotherapy in addition to hormonal therapy.
  \item HER2-enriched: Characterized by negative ER and PR status, but with overexpression of the HER2 protein, often aggressive and requiring targeted HER2 therapies.
  \item Triple-Negative Breast Cancer (TNBC): Defined by the negative for ER, PR, and HER2, this subtype is often more aggressive with limited treatment options and a poorer prognosis.
\end{itemize}
\\ \hline
\end{tabularx}
\end{table*}

\begin{table*}
\setcounter{table}{2}
\tiny
\centering
\caption{The number of images of the public datasets used in this study. Some datasets require permission for access. The discrepancy in the number of images for certain datasets compared to the original reports is due to the removal of samples with data quality issues.}
\label{tab:dataset}
\renewcommand{\arraystretch}{1.1} 
\begin{tabularx}{\textwidth}{>{\raggedright\arraybackslash}p{0.1\textwidth}
              >{\raggedleft\arraybackslash}p{0.1\textwidth}
              >{\raggedright\arraybackslash}X}
\toprule
\textbf{Dataset} & \textbf{Images} & \textbf{Download link} \\ \hline
BMCD \cite{loizidou2021digital} & 400 & \url{https://zenodo.org/records/5036062} \\ \addlinespace
CBIS-DDSM \cite{lee2017curated} & 3,103 & \url{https://www.kaggle.com/datasets/awsaf49/cbis-ddsm-breast-cancer-image-dataset} \\ \addlinespace
CDD-CESM \cite{khaled2022categorized} & 1,003 & \url{https://www.kaggle.com/datasets/krinalkasodiya/new-cdd-cesm-classification-data} \\ \addlinespace
DMID \cite{oza2024digital} & 510 & \url{https://figshare.com/articles/dataset/_b_Digital_mammography_Dataset_for_Breast_Cancer_Diagnosis_Research_DMID_b_DMID_rar/24522883} \\ \addlinespace
INbreast \cite{moreira2012inbreast} & 410 & \url{https://www.kaggle.com/datasets/tommyngx/inbreast2012} \\ \addlinespace
MIAS \cite{suckling1994mammographic} & 322 & \url{https://www.kaggle.com/datasets/kmader/mias-mammography} \\ \addlinespace
CSAW-M \cite{sorkhei2021csaw} & 10,020 & \url{https://figshare.scilifelab.se/articles/dataset/CSAW-M_An_Ordinal_Classification_Dataset_for_Benchmarking_Mammographic_Masking_of_Cancer/14687271} \\ \addlinespace
CMMD \cite{cai2023online} & 3,744 & \url{https://www.kaggle.com/datasets/tommyngx/cmmd2022} \\ \addlinespace
KAU-BCMD \cite{alsolami2021king} & 2,374 & \url{https://www.kaggle.com/datasets/asmaasaad/king-abdulaziz-university-mammogram-dataset} \\ \addlinespace
VinDr-Mammo \cite{nguyen2023vindr} & 20,000 & \url{https://www.physionet.org/content/vindr-mammo/1.0.0/} \\ \addlinespace
RSNA-Mammo \cite{rsna2023challenge} & 54,706 & \url{https://www.kaggle.com/competitions/rsna-breast-cancer-detection/data} \\ \addlinespace
EMBED \cite{jeong2023emory} & 480,275 & \url{https://registry.opendata.aws/emory-breast-imaging-dataset-embed/} \\ \addlinespace
DBT-T \cite{buda2021data} & 135 & \url{https://www.cancerimagingarchive.net/collection/breast-cancer-screening-dbt/}; \url{https://arxiv.org/pdf/2011.07995}\\ \addlinespace
LAMIS \cite{imane2024lamis} & 2,212 & \url{https://github.com/LAMISDMDB/LAMISDMDB_Sample} \\ \addlinespace
MM \cite{aqdar2024mammogram} & 745 & \url{https://data.mendeley.com/datasets/fvjhtskg93/1} \\ \addlinespace
NLBS \cite{kendall2024full} & 23,706 & \url{https://www.frdr-dfdr.ca/repo/dataset/cb5ddb98-ccdf-455c-886c-c9750a8c34c2} \\ \hline
\textbf{Total} & \textbf{603,665} & - \\
\bottomrule
\end{tabularx}
\end{table*}
}
\clearpage
\captionsetup[table]{name=
Supplementary Table}
\setcounter{table}{0}
\section*{Supplementary}
\subsection*{S1. Downstream Task Evaluation Details for FMs}
\subsubsection*{S1.1 Comparison Models}
\label{sec:baselines}

To rigorously evaluate VersaMammo's performance, we conducted comprehensive comparisons with both medical foundation models (FMs) and mammogram FMs. Our baseline selection includes the following models.

\textbf{Medical FMs} include MedSAM (ViT-B) [27], a SAM-based model pre-trained on 1.1 million medical images across 24 modalities, and LVM-Med (ViT-B\&R50) [26], vision transformers pre-trained on 1.5 million multimodal medical images, including mammograms. These models leverage large-scale and diverse medical datasets to learn general features and knowledge, enabling them to perform well across various medical image analysis tasks.

{\textbf{Mammogram FMs}}, such as Mammo-CLIP (Enb2\&Enb5) [35], vision-language models pre-trained on 25,355 mammogram-report pairs, and MAMA (ViT-B) [36], a multi-view alignment model trained on the EMBED dataset, focus on mammogram analysis. These specialist models are designed to excel in tasks related to breast cancer detection and diagnosis by utilizing mammogram data and specific training strategies.

All models were evaluated using consistent input resolutions and pre-processing pipelines. Feature extraction was performed using each model's recommended protocol, with task-specific heads trained from scratch for fair comparison. 

\subsubsection*{S1.2 Evaluation Protocol} \label{evaluation_protocol}
To comprehensively evaluate the performance of different FMs across various downstream tasks, we designed a rigorous testing framework and selected appropriate evaluation metrics for each task. 
For fair comparison, during the fine-tuning stage of downstream tasks, the pre-trained weights of all FMs were kept fixed, and only task-specific heads were trained.

More specifically, we designed two evaluation scenarios to assess model performance under different generalization conditions. \textbf{In internal validation}, we used the training split of a specific dataset to train a task-specific head and employed the corresponding validation set for model selection. The final performance was then quantified on the held-out test split from the same dataset. This setup measures how well a model adapts to a known data distribution and task.
\textbf{In external validation}, models were evaluated on the test splits of completely independent datasets that were not used during pre-training and fine-tuning. This rigorous protocol assesses the model's ability to generalize to unseen clinical settings and populations, reflecting real-world applicability.

\textbf{Performance Metrics and Statistical Comparison.}
We evaluate performance using established metrics: balanced accuracy (BACC), Area Under the receiver operating characteristic Curve (AUC) and macro-F1 for classification and visual question answering (VQA) tasks; average accuracy for retrieval; and Dice and IoU for segmentation and detection tasks. {For consistency across tasks, we adopt macro-averaged metrics whenever applicable. Binary classification is treated as a two-class case when computing AUC and F1, ensuring that both classes are equally weighted. In this setting, BACC corresponds to the average recall across classes. For multiclass classification, we compute AUC using a one-vs-rest (OvR) strategy. Specifically, each class is treated as the positive class against all remaining classes, and a binary AUC is computed for each class. The overall performance is reported as macro-AUC, defined as the unweighted mean of per-class AUC values. For multilabel classification, each label is treated as an independent binary classification task. We compute AUC for each label separately and report macro-AUC as the unweighted average across all labels.}
All statistical comparisons were performed using non-parametric bootstrapping with 1,000 resamples.
For each comparison, we computed the 95\% confidence interval for each metric from the 2.5th and 97.5th percentiles of its bootstrap distribution. Two-sided $p$-values were obtained from the proportion of bootstrap samples where the absolute performance difference was equal to or greater than the observed difference. 
Multiple comparison corrections were conducted separately for internal and external validation. Within each scheme, {experiments} were stratified by task type, and the Benjamini-Hochberg [49] procedure (false discovery rate, FDR $= 0.05$) was applied independently to each stratum and to each metric within multi-metric tasks. 

\textbf{Fairness in Model Comparison.} All comparative analyses between VersaMammo and baseline FMs were conducted using \textit{identical} training, validation, and test splits, as documented in  Supplementary Tables~\ref{tab:dataset_splits}-\ref{tab:external_setup}. This strict protocol ensures that performance differences are attributable to model architecture and training strategy rather than data sampling variations, providing a fair and consistent evaluation baseline. The downstream tasks evaluated across these datasets, along with their respective label information, are summarized in  Supplementary Table~\ref{tab:dataset_annotation_info}. 

\subsubsection*{S1.3 Lesion Detection}
Lesion detection involves identifying the presence and providing an
approximate bounding box localization of suspicious abnormalities (e.g., masses, calcifications) within a mammogram.
To evaluate the detection performance of different FMs, we adopted a standardized Faster R-CNN [39] framework while maintaining consistent testing protocols across all models. Given the inherent architectural differences among FMs, the output feature maps varied in size even for identical input resolutions. To ensure fair comparison, we resized the four feature maps from models with input resolution $224\times224$ to dimensions of $56\times56$, $28\times28$, $14\times14$, and $7\times7$, while those from $512\times512$, $518\times518$ inputs were resized to $128\times128$, $64\times64$, $32\times32$ and $16\times16$. These features were then processed through a feature pyramid network (FPN [40]) to construct multi-scale representations, followed by a region proposal network (RPN) to generate candidate bounding boxes. The Region of Interest (RoI) Align module extracted fixed-size features from these proposals, which were fed into the detection head for final box predictions. Non-maximum suppression (NMS) with a threshold of 0.5 was applied to eliminate overlapping boxes and filter low-confidence detections. During training, we employed random horizontal and vertical flips for data augmentation, and optimized the models using AdamW with a learning rate of $1e^{-3}$ and a batch size of 8, validating every 500 iterations and implementing early stopping after 20 consecutive validation rounds without performance improvement. 

\subsubsection*{S1.4 Lesion Segmentation}
Lesion segmentation requires generating a pixel-level mask that delineates
the precise contours and spatial extent of a lesion.
To evaluate the segmentation performance of different FMs with different model architectures, we implemented a standardized evaluation framework tailored to each type of model. For ViT [34]-based models, such as MedSAM (ViT-B) [27], LVM-Med (ViT-B) [26], and MAMA (ViT-B) [36], we employed the TransUNet [41] architecture, in which input images were first processed through a ResNetV2 [33] backbone for feature extraction, followed by feature projection into the transformer network (with the patch embedding layer removed) and subsequent decoding through a convolutional decoder. In contrast, for CNN-based models, such as LVM-Med (R50) [26], Mammo-CLIP (Enb2) [35], Mammo-CLIP (Enb5) [35], and VersaMammo, we adopted a conventional UNet [42] framework, using FMs as encoder components paired with a symmetric convolutional decoder. This approach enabled us to fairly compare the segmentation capabilities of different FMs with different model architectures. The training protocol incorporated a composite loss function combining binary cross-entropy and Dice loss, with data augmentation limited to random horizontal and vertical flips to preserve anatomical plausibility. Optimization was performed using AdamW [58] with a learning rate of $1e^{-3}$ and a batch size of 8, with validation conducted every 500 iterations and early termination triggered after 20 consecutive validation rounds without performance improvement. 

\subsubsection*{S1.5 Mammogram Classification}
Mammogram classification encompasses {configurations} such as categorizing breast tissue density (A–D), estimating masking risk levels (1-8), identifying abnormalities including calcifications, masses, distortions, and asymmetries, predicting BI-RADS assessment categories (0–5) and subclassifying category 4 lesions (4A–4C), determining benign versus malignant status, differentiating mastitis from cancer, assessing invasive versus in situ disease, and classifying molecular subtypes such as Luminal A, Luminal B, HER2-enriched, and triple-negative breast cancer based on mammographic features.
To evaluate the classification performance of different FMs, we utilized a standard classification model architecture and followed a consistent testing methodology across all models. Specifically, we extracted features from the input images using the respective FMs and then fed these features into a linear classifier for the final classification decision. This approach allowed us to assess the models' ability to generalize across different classification tasks while maintaining a fair comparison. Optimization was performed using AdamW  [58] with a learning rate of $1e^{-3}$ and a batch size of 64, with validation conducted every epoch and early termination triggered after 20 consecutive validation rounds without performance improvement.

\subsubsection*{S1.6 Mammogram Retrieval}
\label{retrieval_method}
Mammogram retrieval refers to the task of retrieving the most visually and/or semantically similar cases from a large-scale database, given a query mammogram or a specific lesion within it.
We performed embedding-based retrieval for each query image and assessed performance based on label agreement between the query and the retrieved images, facilitating efficient image retrieval across different medical centers. To perform the retrieval, we first embedded all images using the respective FMs. Each image in the test split was treated as a query and compared against the images in the training split. To ensure that all features have a comparable impact on the computation of similarity, we independently normalized each feature component to the range [0, 1]. This normalization process involved calculating the mean and variance of the training split features, which were then used to normalize both the training and testing features. This approach allowed us to assess the models' ability to accurately retrieve relevant mammograms based on the query image. To evaluate the retrieval performance, we employed the top-k (k $=1,2,3$) accuracy as the evaluation metric. The top-k accuracy evaluates the proportion of times the correct item is found among the top-k predictions. We used the $L_2$ distance metric to evaluate the similarity between the query image and candidate images, with lower distance values indicating higher similarity. The retrieved images were subsequently ranked based on their similarity scores, and the corresponding class labels were utilized to evaluate the success of the retrieval process. This metric provides a direct measure of the model's ability to accurately retrieve relevant mammograms based on the query image.

\subsubsection*{S1.7 Mammogram Visual Question Answering}
The visual question answering (VQA) refers to the task of answering natural language questions about the content of a mammogram (e.g., “What is the breast density?”, “Is there a mass in the left breast?”).
To evaluate the VQA performance of different FMs, we adopted a standardized ViLT [31] framework while maintaining consistent testing protocols across all models. Specifically, we extracted visual features from input images using the respective FMs and combined them with text embeddings in the ViLT architecture. The output representations from ViLT were then fed into a linear classifier for final answer prediction. This approach enabled us to assess the models' ability to generalize across diverse question topics while ensuring fair comparison conditions. During training, we employed the AdamW optimizer with a learning rate of $1e^{-3}$ and a batch size of 16, conducting validation after each epoch and implementing early stopping if no performance improvement was observed for one consecutive validation round. For performance evaluation, we utilized the same comprehensive metrics as in classification tasks, including AUC, BACC, and macro-F1 score, to thoroughly assess the models' VQA capabilities.

\subsection*{S2. Computational Requirements}
The development of VersaMammo involved two sequential pre-training stages to build the foundational model. The first stage utilized 2 × NVIDIA H800 GPUs for 15 days (totaling 720 GPU hours). Subsequently, the second pre-training stage was conducted on 4 × NVIDIA RTX 3090 GPUs for 27 days (totaling 2,592 GPU hours). During pre-training, the peak GPU memory usage reached approximately 160 GB across devices. Once this foundation model was established, adapting it to each downstream task via training task-specific heads became highly efficient. Each task-specific adaptation required only a single NVIDIA RTX 3090 GPU or NVIDIA L20 GPU, with fine-tuning times varying from several minutes for simpler tasks (e.g., classification) to a maximum of 2 days for the most complex ones (e.g., segmentation and detection), demonstrating the parameter-efficient adaptability of our foundation model.

For inference, which is paramount for clinical deployment, our single-model architecture offers significant practical advantages. Benchmarking on a single RTX 3090 GPU or NVIDIA L20 GPU at an input resolution of $512\times512$ revealed that the inference latency for processing a single mammogram across all five task categories is consistently under one second. At this resolution, the peak GPU memory consumption during inference was only approximately 0.21 GB, highlighting the lightweight deployment characteristics of the proposed model.

\subsection*{S3. Supplementary Tables}
\onecolumn
\captionsetup[longtable]{width=10in, margin={-0.cm,-0cm}}
\newcommand{\mysmall}{\fontsize{7pt}{8pt}\selectfont}
\newcommand{\mytiny}{\fontsize{5pt}{6pt}\selectfont}

\begin{table*}[htbp]
\centering
\mysmall
\caption{Average performance of FMs for diverse tasks across different datasets. It is evaluated using IoU for detection tasks, Dice coefficient for lesion segmentation tasks, average Area Under the receiver operating characteristic Curve (AUC) for visual question answering (VQA) tasks, average accuracy for image retrieval tasks, and AUC for classification tasks. If there are different backbones of the same FM, only the best model is presented here. 
Asterisk (*) indicates the results on external validation datasets. The 95\% CI is included in parentheses. \textbf{Bold} values represent the highest performance, while \underline{underlined} values represent the second-highest performance in each task.}
\setlength{\tabcolsep}{3pt}
\renewcommand{\arraystretch}{1.2}
\begin{tabular}{l|ccccccc}
\hline
Task & Class & Dataset & MedSAM & LVM-Med & Mammo-CLIP & MAMA & VersaMammo \\ 
\hline
\multirow{2}{*}{Detection} & \multirow{2}{*}{2} & \multirow{2}{*}{3} & 44.68\% & 49.67\% & 49.97\% & \underline{50.21\%} & \textbf{57.92\%} \\ 
 & &&(41.07\%, 47.40\%) & (45.65\%, 52.69\%) & (45.93\%, 52.99\%) & (46.15\%, 53.22\%) & (53.23\%, 61.39\%) \\  
\multirow{2}{*}{Segmentation} & \multirow{2}{*}{2} & \multirow{2}{*}{3} & 58.24\% & 59.49\% & \underline{59.77\%} & 58.09\% & \textbf{63.21\%} \\ 
 &&& (49.28\%, 66.46\%) & (51.42\%, 67.39\%) & (51.60\%, 67.94\%) & (49.76\%, 66.03\%) & (55.15\%, 71.28\%) \\ 
\multirow{2}{*}{Composition} & \multirow{2}{*}{3-4} & \multirow{2}{*}{15} & 85.15\% & 84.99\% & 89.68\% & \underline{89.87\%} & \textbf{91.53\%} \\ 
& &  & (81.38\%, 88.40\%) & (81.40\%, 88.24\%) &  (86.78\%, 92.07\%) & (87.03\%, 92.23\%) & (89.09\%, 93.55\%) \\ 
\multirow{2}{*}{Masking} & \multirow{2}{*}{8} & \multirow{2}{*}{1} & 71.99\% & 73.70\% & \textbf{74.55\%} & {73.93\%} & \underline{74.48\%} \\ 
& & & (70.86\%, 73.06\%) & (72.62\%, 74.74\%) & (73.51\%, 75.49\%) & (72.91\%, 74.89\%) & (73.46\%, 75.48\%) \\ 
\multirow{2}{*}{Finding} & \multirow{2}{*}{2-10} & \multirow{2}{*}{8} & 74.82\% & 74.20\% & 81.67\% & \underline{82.75\%} & \textbf{84.21\%} \\ 
& & & (69.77\%, 80.06\%) & (68.86\%, 80.10\%) & (76.43\%, 86.40\%) & (78.12\%, 87.76\%) & (79.47\%, 89.15\%) \\
\multirow{2}{*}{BI-RADS} & \multirow{2}{*}{3-6} & \multirow{2}{*}{14} & 68.70\% & 69.78\% & 78.46\% & \underline{80.07\%} & \textbf{82.95\%} \\ 
& & & (65.04\%, 72.48\%) & (66.05\%, 73.52\%) & (75.34\%, 81.22\%) & (77.01\%, 82.80\%) & (80.20\%, 85.36\%) \\
\multirow{2}{*}{BI-RADS 4} & \multirow{2}{*}{3} & \multirow{2}{*}{3} & 60.25\% & 60.46\% & 65.88\% & \underline{69.49\%} & \textbf{73.81\%} \\ 

& & & (55.57\%, 64.85\%) & (55.76\%, 64.91\%) & (61.02\%, 70.21\%) & (65.10\%, 73.46\%) & (69.78\%, 77.53\%) \\
\multirow{2}{*}{Pathology} & \multirow{2}{*}{2} & \multirow{2}{*}{15} & 70.72\% & 70.59\% & 75.07\% & \underline{78.79\%} & \textbf{81.04\%} \\ 

& & & (62.74\%, 78.01\%) & (62.52\%, 78.04\%) & (62.91\%, 77.47\%) & (71.34\%, 85.56\%) & (74.51\%, 87.35\%) \\
\multirow{2}{*}{Mastitis} & \multirow{2}{*}{2} & \multirow{2}{*}{1} & 68.10\% & 62.22\% & 75.07\% & \underline{82.00\%} & \textbf{85.00\%} \\ 

& & & (63.74\%, 72.57\%) & (57.22\%, 66.96\%) & (70.87\%, 79.04\%) & (78.41\%, 85.43\%) & (81.87\%, 87.69\%) \\  
\multirow{2}{*}{Invasive} & \multirow{2}{*}{2} & \multirow{2}{*}{1} & 64.65\% & 65.90\% & 70.57\% & \underline{74.71\%} & \textbf{80.05\%} \\ 

& & & (60.86\%, 68.58\%) & (62.00\%, 69.71\%) & (66.71\%, 74.25\%) & (71.14\%, 78.04\%) & (76.91\%, 83.30\%) \\
\multirow{2}{*}{Subtype} & \multirow{2}{*}{4} & \multirow{2}{*}{2} & 54.85\% & 55.65\% & \underline{58.78\%} & 58.47\% & \textbf{61.12\%} \\ 

& & & (50.67\%, 58.82\%) & (51.81\%, 59.52\%) & (54.73\%, 62.36\%) & (54.83\%, 62.19\%) & (57.12\%, 65.01\%) \\
\multirow{2}{*}{Retrieval} & \multirow{2}{*}{6} & \multirow{2}{*}{3} & 64.68\% & 64.99\% & 66.11\% & \underline{67.17\%} & \textbf{68.21\%} \\ 

& & &(64.40\%, 64.96\%)
&(64.70\%, 65.27\%)
&(65.83\%, 66.41\%)
&(66.90\%, 67.44\%)
&(67.94\%, 68.49\%)\\ 
\multirow{2}{*}{VQA} & \multirow{2}{*}{2-10} & \multirow{2}{*}{11} & 72.17\% & 79.02\% & 70.98\% & \underline{79.11\%} & \textbf{85.88\%} \\ 

& & & (71.05\%, 73.43\%)&(77.93\%, 80.18\%)&(69.58\%, 72.31\%)&(77.87\%, 80.24\%)&(85.04\%, 86.70\%)\\ 
\hline
\multirow{2}{*}{Detection*} & \multirow{2}{*}{2} & \multirow{2}{*}{1} 
& 29.28\% & 36.69\%  & \underline{38.27\%} & 30.96\% & \textbf{40.17\%}\\
& & & (26.10\%, 32.37\%) & (33.52\%, 39.91\%)  & (35.08\%, 41.49\%) & (27.75\%, 33.98\%) & (36.63\%, 43.50\%) \\ 
\multirow{2}{*}{Segmentation*} & \multirow{2}{*}{2} & \multirow{2}{*}{1} & 39.90\% & 39.31\% & 35.05\% & \underline{43.20\%} & \textbf{44.95\%} \\ 

& & & (32.49\%, 46.99\%) & (32.70\%, 45.84\%) & (28.80\%, 42.05\%) & (36.10\%, 49.96\%) & (38.01\%, 52.63\%) \\ 
\multirow{2}{*}{Composition*} & \multirow{2}{*}{4} & \multirow{2}{*}{6} & 79.56\% & 80.07\% & 83.60\% & \underline{85.78\%} & \textbf{86.80\%} \\ 

& & & (78.81\%, 80.11\%) & (79.44\%, 80.59\%) & (83.00\%, 84.03\%) & (85.24\%, 86.15\%) & (86.30\%, 87.17\%) \\ 
\multirow{2}{*}{BI-RADS*} & \multirow{2}{*}{6} & \multirow{2}{*}{6} & 55.19\% & 56.06\% & 60.53\% & \underline{62.22\%} & \textbf{67.26\%} \\ 

& & & (54.35\%, 56.04\%) & (55.23\%, 56.84\%) & (59.65\%, 61.31\%) & (61.34\%, 62.98\%) & (66.49\%, 67.93\%) \\
\multirow{2}{*}{Pathology*} & \multirow{2}{*}{2} & \multirow{2}{*}{8} & 56.70\% & 55.53\% & 63.65\% & \underline{65.52\%} & \textbf{74.24\%} \\ 

& & & (54.58\%, 58.73\%) & (53.92\%, 58.20\%) & (61.71\%, 65.59\%) & (64.85\%, 68.66\%) & (72.43\%, 75.89\%) \\  
\multirow{2}{*}{Retrieval*} & \multirow{2}{*}{6} & \multirow{2}{*}{5} & 49.20\% & 49.08\% & 52.40\% & \underline{52.94\%} & \textbf{54.18\%} \\ 

& & &(48.58\%, 49.83\%)&(48.45\%, 49.66\%)&(51.77\%, 52.99\%)&(52.35\%, 53.55\%)&(53.35\%, 54.63\%)\\ 
\multirow{2}{*}{VQA*} & \multirow{2}{*}{2-6} & \multirow{2}{*}{4} & 57.47\% & 56.89\% & 56.98\% & \textbf{71.72\%} & \underline{64.25\%} \\ 

& & & (55.46\%, 59.40\%)&(54.89\%, 58.76\%)&(54.93\%, 59.01\%)&(70.03\%, 73.44\%)&(62.37\%, 66.05\%) \\ 
\hline
\end{tabular}

\label{tab:avg_all}
\end{table*}

\begin{table*}[h!]
  \centering
  \caption{Internal comparison of detection performance (IoU) across different FMs on high resolution ($518\times 518$ for MAMA (ViT-B), $512\times 512$ for others) and low resolution ($224\times 224$) mammograms. The 95\% CI is included in parentheses. The best performing model for each metric is \textbf{bolded} and the second-best performing model is \underline{underlined}. $p$-values are computed between the first-best and second-best performing models.}
  \mysmall
  \begin{tabular}{l|ccc}
    \hline
    \textbf{\diagbox[width=10em]{Model}{Dataset}} & \textbf{CBIS-DDSM}& \textbf{INbreast}& \textbf{VinDr-Mammo} \\
    \hline
    &\multicolumn{3}{c}{High resolution} \\
    \hline
    MedSAM (ViT-B) & 37.92\% (34.85\%, 40.21\%) & 51.13\% (46.99\%, 54.23\%) & 45.00\% (41.36\%, 47.75\%) \\
    LVM-Med (ViT-B) & 41.98\% (38.58\%, 44.52\%) & 60.85\% (55.93\%, 64.57\%) & 46.18\% (42.45\%, 48.98\%) \\
    LVM-Med (R50) & 42.68\% (39.22\%, 45.27\%) & 51.03\% (46.90\%, 54.14\%) & 38.88\% (35.74\%, 41.24\%) \\
    Mammo-CLIP (Enb2) & 38.06\% (34.98\%, 40.36\%) & \underline{62.32\% (57.27\%, 66.08\%)} & 41.67\% (38.30\%, 44.17\%) \\
    Mammo-CLIP (Enb5) & 44.03\% (40.46\%, 46.66\%) & 60.32\% (55.44\%, 63.99\%) & 45.57\% (41.89\%, 48.32\%) \\
    MAMA (ViT-B) & \underline{48.18\% (44.28\%, 51.07\%)} & 56.22\% (51.67\%, 59.61\%) & \underline{46.23\% (42.49\%, 48.99\%)} \\
    VersaMammo (Ours) & \textbf{48.24\% (44.33\%, 51.14\%)}& \textbf{71.20\% (65.44\%, 75.47\%)} & \textbf{54.32\% (49.92\%, 57.57\%)} \\
    $p$-value &$p=0.17$ &$p<0.001$ &$p<0.001$  \\
    \hline
    &\multicolumn{3}{c}{Low resolution} \\
    \hline
    MedSAM (ViT-B) & 35.58\% (32.13\%, 37.94\%) & 44.56\% (40.86\%, 47.28\%) & \underline{42.97\% (39.40\%, 45.59\%)} \\
    LVM-Med (ViT-B) & 36.87\% (33.81\%, 39.12\%) & 49.77\% (45.63\%, 52.80\%) & 42.95\% (38.92\%, 44.60\%) \\
    LVM-Med (R50) & \underline{41.82\% (38.35\%, 44.37\%)} & \underline{54.49\% (49.96\%, 57.81\%)} & 41.65\% (38.19\%, 44.19\%) \\
    Mammo-CLIP (Enb2) & 31.54\% (28.92\%, 33.46\%) & 50.90\% (46.68\%, 54.00\%) & 36.66\% (33.62\%, 38.90\%) \\
    Mammo-CLIP (Enb5) & 37.67\% (34.54\%, 39.97\%) & 52.47\% (48.11\%, 55.67\%) & 37.35\% (34.25\%, 39.63\%) \\
    MAMA (ViT-B) & 35.55\% (32.60\%, 37.72\%) & 49.11\% (45.03\%, 52.10\%) & 37.53\% (34.12\%, 39.43\%) \\
    VersaMammo (Ours)& \textbf{46.18\% (42.35\%, 48.99\%)} & \textbf{62.15\% (56.99\%, 65.94\%)} & \textbf{49.74\% (45.61\%, 52.77\%)} \\
    $p$-value &$p<0.001$ &$p<0.001$ &$p<0.001$  \\
    \hline
  \end{tabular}
  \label{tab:det}
\end{table*}

\begin{table*}[h!]
  \centering
  \caption{Average test time per image across different FMs on the detection task for high resolution and low resolution mammograms. The best performing model for each metric is \textbf{bolded} and the second-best performing model is \underline{underlined}. The standard deviation is included.}
  \small
  \begin{tabular}{l|cc}
    \hline
    \textbf{Model} & \textbf{High resolution} & \textbf{Low resolution} \\
    \hline
    MedSAM (ViT-B) & 166ms$\pm$110ms & 65ms$\pm$103ms \\
    LVM-Med (ViT-B) & 166ms$\pm$110ms & 65ms$\pm$103ms \\
    LVM-Med (R50) & \textbf{83ms$\pm$107ms} & \textbf{45ms$\pm$96ms} \\
    Mammo-CLIP (Enb2) & \underline{84ms$\pm$131ms} & \underline{50ms$\pm$111ms} \\
    Mammo-CLIP (Enb5) & 110ms$\pm$132ms & 69ms$\pm$126ms \\
    MAMA (ViT-B) & 259ms$\pm$262ms & 97ms$\pm$156ms \\
    VersaMammo (Ours) & 110ms$\pm$132ms & 69ms$\pm$126ms \\
    \hline
  \end{tabular}
  \label{tab:det_time}
\end{table*}

\begin{table*}[h!]
  \centering
  \caption{External comparison of lesion detection performance (IoU) across different FMs on high resolution ($518\times 518$ for MAMA (ViT-B), $512\times 512$ for others) mammograms. The training splits of INbreast and VinDr-Mammo are used to train the detection head, and the test split of CBIS-DDSM is employed as the external validation dataset. The 95\% CI is included in parentheses. The best performing model for each metric is \textbf{bolded} and the second-best performing model is \underline{underlined}. $p$-values are computed between the first-best and second-best performing models.}
    \small
  \begin{tabular}{l|c}
    \hline
    {\textbf{\diagbox[width=10em]{Model}{Dataset}}} & \textbf{CBIS-DDSM} \\
\midrule
MedSAM (ViT-B)            & 29.28\% (26.10\%, 32.37\%) \\
LVM-Med (ViT-B)           & 36.69\% (33.52\%, 39.91\%) \\
LVM-Med (R50)             & 25.57\% (22.82\%, 28.64\%) \\
Mammo-CLIP (Enb2)         & 29.06\% (26.10\%, 32.27\%) \\
Mammo-CLIP (Enb5)         & \underline{38.27\% (35.08\%, 41.49\%)} \\
MAMA (ViT-B)              & 30.96\% (27.75\%, 33.98\%) \\

\textbf{VersaMammo (Ours)} & \textbf{40.17\% (36.63\%, 43.50\%)} \\
$p$-value &$p=0.390$\\
    \hline
  \end{tabular}
  \label{tab:det_ex}
\end{table*}

\begin{table*}[h!]
\centering
\caption{Internal comparison of lesion segmentation performance (Dice) across different FMs on high resolution ($512\times 512$) and low resolution ($224\times224$) mammograms. The 95\% CI is included in parentheses. The best performing model for each metric is \textbf{bolded} and the second-best performing model is \underline{underlined}. $p$-values are computed between the first-best and second-best performing models.}
\mysmall
\begin{tabular}{l|cccc}
\hline
\textbf{\diagbox[width=10em]{Model}{Dataset}} & \textbf{CBIS-DDSM} & \textbf{INbreast} & \textbf{PDS5} \\
\hline
& \multicolumn{3}{c}{High resolution} \\
\hline
MedSAM (ViT-B) & 47.98\% (44.11\%, 51.99\%) & 64.73\% (47.67\%, 80.03\%) & 62.00\% (56.07\%, 67.35\%) \\

LVM-Med (ViT-B) & 48.34\% (44.67\%, 51.87\%) & 67.43\% (52.35\%, 81.68\%) & \underline{62.71\% (57.25\%, 68.61\%)} \\

LVM-Med (R50) & 42.77\% (39.12\%, 46.42\%) & 71.35\% (56.12\%, 85.78\%) & 58.49\% (52.75\%, 64.23\%) \\

Mammo-CLIP (Enb2) & 45.78\% (41.97\%, 49.59\%) & \underline{72.19\% (57.49\%, 86.89\%)} & 59.81\% (54.11\%, 65.51\%) \\

Mammo-CLIP (Enb5) & \underline{49.71\% (45.85\%, 53.57\%)} & 68.46\% (53.52\%, 83.40\%) & 61.13\% (55.42\%, 66.84\%) \\

MAMA (ViT-B) & 47.91\% (43.80\%, 51.83\%) & 65.16\% (50.02\%, 79.00\%) & 61.22\% (55.47\%, 67.25\%) \\

VersaMammo (Ours) & \textbf{51.61\% (47.71\%, 55.51\%)} & \textbf{74.53\% (59.89\%, 89.17\%)} & \textbf{63.50\% (57.85\%, 69.15\%)} \\
$p$-value &$p<0.001$  &$p<0.001$ &$p<0.001$  \\
\hline
& \multicolumn{3}{c}{Low resolution} \\
\hline
MedSAM (ViT-B) & \textbf{47.52\% (43.26\%, 51.61\%)} & 63.88\% (46.65\%, 81.37\%) & 52.08\% (45.32\%, 58.66\%) \\
LVM-Med (ViT-B) & 46.06\% (42.42\%, 49.41\%)& 53.17\% (33.96\%, 72.00\%) & 50.88\% (44.38\%, 57.46\%) \\
LVM-Med (R50) & 41.87\% (37.81\%, 45.51\%) & 64.54\% (49.66\%, 78.56\%) & 53.73\% (47.23\%, 60.31\%) \\
Mammo-CLIP (Enb2) & 42.75\% (39.00\%, 46.33\%) & \underline{64.74\% (48.85\%, 79.28\%)} & \textbf{57.91\% (52.31\%, 63.60\%)} \\
Mammo-CLIP (Enb5) & 44.44\% (40.60\%, 47.97\%) & 64.46\% (48.16\%, 78.48\%) & 57.33\% (51.89\%, 62.90\%) \\
MAMA (ViT-B) & \underline{46.12\% (42.51\%, 49.93\%)} & 60.55\% (44.54\%, 76.51\%) & 53.32\% (46.76\%, 59.62\%) \\
VersaMammo (Ours) & 45.71\% (41.79\%, 49.69\%) &\textbf{73.79\% (60.42\%, 85.19\%)} & \underline{57.63\% (51.59\%, 63.55\%)} \\
$p$-value &$p=0.28$ &$p<0.001$ &$p=0.069$  \\
\hline
\end{tabular}
\label{tab:seg}
\end{table*}

\begin{table*}[h!]
  \centering
  \caption{Average test time per image across different FMs on the lesion segmentation task for high-resolution and low-resolution mammograms. The best performing model for each metric is \textbf{bolded} and the second-best performing model is \underline{underlined}. The standard deviation is included.}
  \begin{tabular}{l|cc}
    \hline
    \textbf{Model} & \textbf{High resolution} & \textbf{Low resolution} \\
    \hline
    MedSAM (ViT-B) & 126ms$\pm$73ms & 39ms$\pm$61ms \\
    LVM-Med (ViT-B) & 126ms$\pm$73ms & 39ms$\pm$61ms \\
    LVM-Med (R50) & \textbf{14ms$\pm$42ms} & \textbf{13ms$\pm$40ms} \\
    Mammo-CLIP (Enb2) & \underline{23ms$\pm$51ms} & \underline{23ms$\pm$55ms} \\
    Mammo-CLIP (Enb5) & 32ms$\pm$59ms &30ms$\pm$51ms   \\
    MAMA (ViT-B) & 290ms$\pm$613ms & 184ms$\pm$144ms \\
    VersaMammo (Ours) & 32ms$\pm$59ms & 30ms$\pm$51ms  \\
    \hline
  \end{tabular}
  \label{tab:seg_time}
\end{table*}

\begin{table*}[h!]
\centering
\caption{External comparison of lesion segmentation performance (Dice) across different FMs on high resolution ($512\times 512$) mammograms. The 95\% CI is included in parentheses. The training splits of INbreast and CBIS-DDSM are used to train the segmentation head, and the test split of PDS5 is employed as the external validation dataset. The best performing model for each metric is \textbf{bolded} and the second-best performing model is \underline{underlined}. $p$-values are computed between the first-best and second-best performing models.}
  \small
\begin{tabular}{l|c}
\hline
{\textbf{\diagbox[width=10em]{Model}{Dataset}}} & \textbf{PDS5} \\

\hline
MedSAM (ViT-B) & 39.90\% (32.49\%, 46.99\%) \\

LVM-Med (ViT-B) & 39.31\% (32.70\%, 45.84\%) \\

LVM-Med (R50) & 22.73\% (16.88\%, 29.21\%) \\

Mammo-CLIP (Enb2) & 35.05\% (28.80\%, 42.05\%) \\

Mammo-CLIP (Enb5) & 30.84\% (24.41\%, 38.24\%) \\

MAMA (ViT-B) & \underline{43.20\% (36.10\%, 49.96\%)} \\

VersaMammo (Ours) & \textbf{44.95\% (38.01\%, 52.63\%)} \\
$p$-value &$p=0.774$  \\
\hline
\end{tabular}
\label{tab:seg_ex}
\end{table*}

\clearpage
{\mysmall

\begin{longtable}{lccc}
\caption{Internal validation on breast composition assessment for different FMs. Non-parametric bootstrapping with 1,000 bootstrap replicates is used for statistical analysis. The 95\% CI is included in parentheses. The best performing model for each metric is \textbf{bolded}. $p$-values are computed between the first-best and second-best performing models.}\\
\hline
\textbf{Model} & \textbf{BACC}& \textbf{macro-F1} & \textbf{macro-AUC} \\
\hline
\endfirsthead

\multicolumn{4}{c}%
{{\bfseries \tablename\ \thetable{} - continued from previous page}} \\
\hline
\textbf{Model} & \textbf{BACC}& \textbf{macro-F1} & \textbf{macro-AUC} \\
\hline
\endhead

\hline \multicolumn{4}{r}{{Continued on next page}} \\
\endfoot

\hline
\endlastfoot

\multicolumn{4}{c}{BMCD}\\
MedSAM (ViT-B) & 71.25\% (61.25\%, 81.25\%) & 51.80\% (41.69\%, 70.49\%) & 88.51\% (83.36\%, 93.05\%) \\
LVM-Med (ViT-B) & {78.75\% (68.75\%, 87.50\%)} & {74.62\% (54.31\%, 86.64\%)} & {94.01\% (90.01\%, 97.21\%)} \\
LVM-Med (R50) & 73.75\% (63.75\%, 83.75\%) & 71.11\% (47.15\%, 83.05\%) & 90.04\% (84.18\%, 94.23\%) \\
Mammo-CLIP (Enb2) & 88.75\% (81.25\%, 95.00\%) & 87.80\% (68.71\%, 96.19\%) & 97.99\% (96.00\%, 99.51\%) \\
Mammo-CLIP (Enb5) & 85.00\% (77.47\%, 92.50\%) & 85.16\% (66.43\%, 93.40\%) & 95.81\% (92.74\%, 98.33\%) \\
MAMA (ViT-B) & 83.75\% (76.22\%, 91.25\%) & 76.52\% (59.41\%, 91.79\%) & 97.54\% (95.36\%, 99.16\%) \\
VersaMammo (Ours) & \textbf{92.50\% (86.25\%, 97.50\%)} & \textbf{94.55\% (89.59\%, 98.17\%)} & \textbf{98.92\% (97.37\%, 99.89\%)} \\
\end{longtable}
}
\clearpage
{\mysmall
\begin{table*}[h!]
\centering
\caption{External validation on breast composition assessment task for different FMs. Non-parametric bootstrapping with 1,000 bootstrap replicates is used for statistical analysis. The 95\% CI is included in parentheses. The pre-training datasets is used to train the classification head. The best performing model for each metric is \textbf{bolded}. $p$-values are computed between the first-best and second-best performing models.}
\mysmall
\begin{tabular}{lccc}
\hline
\textbf{Model} & \textbf{BACC}& \textbf{macro-F1} & \textbf{macro-AUC} \\
\hline
\multicolumn{4}{c}{CDD-CESM}\\
MedSAM (ViT-B) & 45.66\% (43.91\%, 47.28\%) & 42.77\% (41.33\%, 44.13\%) & 84.69\% (83.40\%, 85.74\%) \\
LVM-Med (ViT-B) & 50.26\% (48.49\%, 51.90\%) & 50.21\% (48.63\%, 51.41\%) & 82.69\% (81.36\%, 83.77\%) \\
LVM-Med (R50) & 56.52\% (54.82\%, 58.31\%) & 54.03\% (52.57\%, 55.72\%) & 70.73\% (69.35\%, 72.22\%) \\
Mammo-CLIP (Enb2) & 43.51\% (41.70\%, 45.22\%) & 46.56\% (44.35\%, 48.29\%) & 85.22\% (83.95\%, 86.31\%) \\
Mammo-CLIP (Enb5) & 70.08\% (68.17\%, 71.96\%) & 72.31\% (70.42\%, 74.20\%) & 86.63\% (85.37\%, 87.60\%) \\
MAMA (ViT-B) & 47.96\% (46.26\%, 49.70\%) & 51.03\% (49.09\%, 52.56\%) & 84.73\% (83.46\%, 85.63\%) \\
\hline
\end{tabular}
\label{tab:compo_ex}
\end{table*}
}

\begin{table*}[h!]
\centering
\caption{Internal validation on masking potential risk evaluation on CSAW-M for different FMs. Non-parametric bootstrapping with 1,000 bootstrap replicates is used for statistical analysis. The 95\% CI is included in parentheses. The best performing model for each metric is \textbf{bolded}. $p$-values are computed between the first-best and second-best performing models.}
\mysmall
\begin{tabular}{lccc}
\hline
\textbf{Model} & \textbf{BACC}& \textbf{macro-F1} & \textbf{macro-AUC} \\
\hline
MedSAM (ViT-B) & 24.50\% (22.70\%, 26.25\%) & 15.23\% (13.62\%, 16.51\%) & 71.99\% (70.86\%, 73.06\%) \\
LVM-Med (ViT-B) & 27.20\% (25.25\%, 29.09\%) & 19.80\% (18.22\%, 21.28\%) & 73.70\% (72.62\%, 74.74\%) \\
LVM-Med (R50) & 25.10\% (23.25\%, 26.95\%) & 17.77\% (16.25\%, 19.34\%) & 72.31\% (71.26\%, 73.42\%) \\
Mammo-CLIP (Enb2) & 27.00\% (25.10\%, 28.79\%) & \textbf{24.97\% (23.07\%, 26.74\%)} & \textbf{74.55\% (73.51\%, 75.49\%)} \\
Mammo-CLIP (Enb5) & 26.15\% (24.25\%, 27.99\%) & 23.41\% (21.48\%, 25.26\%) & 72.98\% (71.95\%, 74.09\%) \\
MAMA (ViT-B) & 24.75\% (22.85\%, 26.55\%) & 21.12\% (19.34\%, 22.88\%) & 73.93\% (72.91\%, 74.89\%) \\
VersaMammo (Ours) & \textbf{27.74\% (25.85\%, 29.84\%)} & 22.66\% (21.04\%, 24.42\%) & 74.48\% (73.46\%, 75.48\%) \\
$p$-value &$p<0.001$ &$p<0.001$ &$p=0.038$ \\
\hline
\end{tabular}
\label{tab:mask}

\end{table*}

\clearpage
{\mysmall
\begin{longtable}{lccc}
\caption{Internal validation on finding abnormality identification for different FMs. Non-parametric bootstrapping with 1,000 bootstrap replicates is used for statistical analysis. The 95\% CI is included in parentheses. The best performing model for each metric is \textbf{bolded}. $p$-values are computed between the first-best and second-best performing models.} \\
\hline
\textbf{Model} & \textbf{BACC} & \textbf{macro-F1} & \textbf{macro-AUC} \\
\hline
\endfirsthead

\hline \multicolumn{4}{r}{{Continued on next page}} \\
\endfoot
\hline
\endlastfoot

\multicolumn{4}{c}{CBIS-DDSM}\\
MedSAM (ViT-B) & 88.66\% (86.13\%, 90.90\%) & 88.64\% (86.08\%, 90.87\%) & 95.66\% (94.30\%, 96.98\%) \\
LVM-Med (ViT-B) & 86.97\% (84.59\%, 89.36\%) & 86.97\% (84.58\%, 89.36\%) & 94.67\% (92.93\%, 96.13\%) \\
LVM-Med (R50) & 86.97\% (84.45\%, 89.36\%) & 86.97\% (84.45\%, 89.35\%) & 94.52\% (92.99\%, 95.99\%) \\
Mammo-CLIP (Enb2) & \textbf{91.46\% (89.22\%, 93.56\%)} & 91.46\% (89.22\%, 93.54\%) & 96.74\% (95.50\%, 97.87\%) \\
Mammo-CLIP (Enb5) & 89.78\% (87.54\%, 92.02\%) & 89.78\% (87.52\%, 92.01\%) & 95.73\% (94.25\%, 97.04\%) \\
MAMA (ViT-B) & \textbf{91.46\% (89.36\%, 93.56\%)} & \textbf{91.46\% (89.34\%, 93.54\%)} & \textbf{96.83\% (95.58\%, 97.91\%)} \\
VersaMammo (Ours) & 87.25\% (84.73\%, 89.64\%) & 87.25\% (84.73\%, 89.60\%) & 95.08\% (93.66\%, 96.51\%) \\
\end{longtable}
}

\clearpage
{\mysmall
\begin{longtable}{lccc}
\caption{Internal validation on BI-RADS assessment task for different FMs. Non-parametric bootstrapping with 1,000 bootstrap replicates is used for statistical analysis. The 95\% CI is included in parentheses. The best performing model for each metric is \textbf{bolded}. $p$-values are computed between the first-best and second-best performing models.} \\
\hline
\textbf{Model} & \textbf{BACC}& \textbf{macro-F1} & \textbf{macro-AUC} \\
\hline
\endfirsthead

\multicolumn{4}{c}%
{{\bfseries \tablename\ \thetable{} - continued from previous page}} \\
\hline
\textbf{Model} & \textbf{BACC}& \textbf{macro-F1} & \textbf{macro-AUC} \\
\hline
\endhead

\hline \multicolumn{4}{r}{{Continued on next page}} \\
\endfoot

\hline
\endlastfoot

\multicolumn{4}{c}{BMCD}\\
MedSAM (ViT-B) & 45.00\% (35.00\%, 55.00\%) & 32.11\% (23.63\%, 42.55\%) & 68.70\% (61.88\%, 75.46\%) \\
LVM-Med (ViT-B) & 42.50\% (31.25\%, 53.75\%) & 36.16\% (22.77\%, 49.09\%) & 74.37\% (67.71\%, 80.64\%) \\
LVM-Med (R50) & 55.00\% (43.75\%, 66.25\%) & 50.11\% (32.58\%, 65.51\%) & 80.73\% (74.22\%, 86.86\%) \\

Mammo-CLIP (Enb2) & 70.00\% (60.00\%, 80.00\%) & 75.03\% (66.11\%, 83.10\%) & 88.15\% (82.35\%, 93.28\%) \\
Mammo-CLIP (Enb5) & \textbf{73.75\% (63.75\%, 82.50\%)} & \textbf{78.53\% (68.84\%, 85.82\%)} & 89.93\% (84.36\%, 94.36\%) \\
MAMA (ViT-B) & 67.50\% (56.25\%, 76.28\%) & 65.02\% (44.63\%, 76.95\%) & 88.93\% (83.14\%, 93.63\%) \\
VersaMammo (Ours) & 72.50\% (62.50\%, 81.25\%) & 73.81\% (54.07\%, 82.78\%) & \textbf{91.10\% (86.52\%, 95.00\%)} \\
\end{longtable}
}
\begin{table*}[h!]
\centering
\caption{External validation on BI-RADS assessment task for different FMs. Non-parametric bootstrapping with 1,000 bootstrap replicates is used for statistical analysis. The 95\% CI is included in parentheses. The pre-training datasets is used to train the classification head. The best performing model for each metric is \textbf{bolded}. $p$-values are computed between the first-best and second-best performing models.}
\mysmall
\begin{tabular}{lccc}
\hline
\textbf{Model} & \textbf{BACC}& \textbf{macro-F1} & \textbf{macro-AUC} \\
\hline
\multicolumn{4}{c}{CDD-CESM}\\
MedSAM (ViT-B) & 23.04\% (22.40\%, 23.79\%) & 15.80\% (15.13\%, 16.35\%) & 56.87\% (55.77\%, 57.98\%) \\
LVM-Med (ViT-B) & 27.72\% (26.59\%, 28.92\%) & 23.25\% (21.85\%, 24.72\%) & 58.60\% (57.48\%, 59.74\%) \\
LVM-Med (R50) & 21.91\% (21.09\%, 22.74\%) & 16.43\% (15.25\%, 17.49\%) & 53.63\% (52.51\%, 54.73\%) \\
Mammo-CLIP (Enb2) & 27.70\% (26.27\%, 29.07\%) & 25.88\% (24.46\%, 27.26\%) & 60.00\% (58.77\%, 61.12\%) \\
Mammo-CLIP (Enb5) & 32.93\% (31.55\%, 34.39\%) & 31.87\% (30.51\%, 33.23\%) & 63.42\% (62.22\%, 64.62\%) \\
MAMA (ViT-B) & 38.16\% (36.50\%, 39.71\%) & 37.59\% (36.11\%, 39.05\%) & 67.94\% (66.86\%, 68.93\%) \\
\hline
\end{tabular}
\label{tab:birads_ex}
\end{table*}

\begin{table*}[h!]
\centering
\caption{Internal validation on BI-RADS  4 subclassification for different FMs. Non-parametric bootstrapping with 1,000 bootstrap replicates is used for statistical analysis. The 95\% CI is included in parentheses. The best performing model for each metric is \textbf{bolded}. $p$-values are computed between the first-best and second-best performing models.}
\mysmall
\begin{tabular}{lccc}
\hline
\textbf{Model} & \textbf{BACC}& \textbf{macro-F1} & \textbf{macro-AUC} \\
\hline

\multicolumn{4}{c}{PDS2}\\
MedSAM (ViT-B) & 60.61\% (54.71\%, 66.58\%) & 61.24\% (54.85\%, 67.69\%) & 59.99\% (52.14\%, 68.00\%) \\
LVM-Med (ViT-B) & 57.96\% (52.51\%, 63.18\%) & 58.57\% (52.18\%, 64.39\%) & 62.38\% (54.35\%, 69.54\%) \\
LVM-Med (R50) & 55.75\% (51.72\%, 60.31\%) & 55.11\% (49.10\%, 61.47\%) & 60.35\% (52.61\%, 68.52\%) \\
Mammo-CLIP (Enb2) & 68.07\% (61.20\%, 74.05\%) & 66.37\% (60.23\%, 71.83\%) & 67.96\% (59.67\%, 75.07\%) \\
Mammo-CLIP (Enb5) & 66.08\% (59.68\%, 72.25\%) & 66.76\% (60.08\%, 73.02\%) & 67.85\% (59.65\%, 75.36\%) \\
MAMA (ViT-B) & 69.65\% (63.50\%, 75.14\%) & 70.03\% (63.98\%, 75.59\%) & 75.77\% (68.65\%, 82.12\%) \\
\hline
\end{tabular}
\label{tab:birads4}
\end{table*}

\clearpage
{\mysmall
\begin{longtable}{lccc}
\caption{Internal validation on pathological diagnosis prediction for different FMs. Non-parametric bootstrapping with 1,000 bootstrap replicates is used for statistical analysis. The 95\% CI is included in parentheses. The best performing model for each metric is \textbf{bolded}. $p$-values are computed between the first-best and second-best performing models.}\\
\hline
\textbf{Model} & \textbf{BACC} & \textbf{macro-F1} & \textbf{macro-AUC} \\
\hline
\endfirsthead

\multicolumn{4}{c}%
{{\bfseries \tablename\ \thetable{} - continued from previous page}} \\
\hline
\textbf{Model} & \textbf{BACC} & \textbf{macro-F1} & \textbf{macro-AUC} \\
\hline
\endhead

\hline \multicolumn{4}{r}{{Continued on next page}} \\
\endfoot
\hline
\endlastfoot

\multicolumn{4}{c}{CBIS-DDSM}\\
MedSAM (ViT-B) & 56.16\% (52.66\%, 59.95\%) & 57.02\% (53.35\%, 60.71\%) & 75.05\% (72.33\%, 77.86\%) \\
LVM-Med (ViT-B) & 65.13\% (61.62\%, 68.63\%) & 66.33\% (62.74\%, 69.77\%) & 80.41\% (77.66\%, 83.05\%) \\
LVM-Med (R50) & 60.50\% (56.72\%, 64.15\%) & 62.31\% (58.69\%, 65.99\%) & 79.33\% (76.74\%, 81.91\%) \\
Mammo-CLIP (Enb2) & 65.55\% (62.18\%, 69.05\%) & 67.03\% (63.52\%, 70.38\%) & 82.23\% (79.78\%, 84.73\%) \\
Mammo-CLIP (Enb5) & \textbf{67.65\% (64.29\%, 70.73\%)} & \textbf{68.71\% (65.26\%, 71.60\%)} & 83.36\% (80.87\%, 85.61\%) \\
MAMA (ViT-B) & 66.81\% (63.17\%, 70.31\%) & 68.18\% (64.51\%, 71.49\%) & \textbf{83.88\% (81.41\%, 86.07\%)} \\
VersaMammo (Ours) & 64.85\% (61.48\%, 68.49\%) & 66.56\% (63.13\%, 70.08\%) & 81.25\% (78.69\%, 83.76\%) \\
\end{longtable}
}
\clearpage
{\mysmall
\begin{longtable}{lccc}
\caption{External validation on pathological diagnosis prediction for different FMs. Non-parametric bootstrapping with 1,000 bootstrap replicates is used for statistical analysis. The 95\% CI is included in parentheses. The pre-training datasets is used to train the classification head. The best performing model for each metric is \textbf{bolded}. $p$-values are computed between the first-best and second-best performing models.}\\
\hline
\textbf{Model} & \textbf{BACC} & \textbf{macro-F1} & \textbf{macro-AUC} \\
\hline
\endfirsthead

\multicolumn{4}{c}{CDD-CESM}\\
MedSAM (ViT-B) & 50.85\% (49.90\%, 51.61\%) & 43.08\% (41.82\%, 44.32\%) & 44.08\% (42.40\%, 45.77\%) \\
LVM-Med (ViT-B) & 50.32\% (48.81\%, 51.82\%) & 50.03\% (48.52\%, 51.55\%) & 47.15\% (45.47\%, 48.87\%) \\
LVM-Med (R50) & 48.59\% (47.65\%, 49.54\%) & 40.06\% (38.76\%, 41.34\%) & 38.53\% (36.91\%, 40.14\%) \\
Mammo-CLIP (Enb2) & 57.99\% (56.67\%, 59.50\%) & 57.20\% (55.83\%, 58.72\%) & 55.65\% (53.96\%, 57.33\%) \\
Mammo-CLIP (Enb5) & 52.77\% (51.53\%, 54.27\%) & 51.04\% (49.75\%, 52.59\%) & 46.47\% (44.89\%, 48.17\%) \\
MAMA (ViT-B) & 59.45\% (57.95\%, 60.96\%) & 59.21\% (57.70\%, 60.72\%) & 61.15\% (59.45\%, 62.69\%) \\
VersaMammo (Ours) & \textbf{68.44\% (66.94\%, 69.95\%)} & \textbf{68.40\% (66.90\%, 69.91\%)} & \textbf{74.22\% (72.74\%, 75.66\%)} \\
\end{longtable}
}

\begin{table*}[h!]
\centering
\caption{Internal validation on mastitis differentiation on PDS5 for different FMs. Non-parametric bootstrapping with 1,000 bootstrap replicates is used for statistical analysis. The 95\% CI is included in parentheses. The best performing model for each metric is \textbf{bolded}. $p$-values are computed between the first-best and second-best performing models.}
\mysmall
\begin{tabular}{lccc}
\hline
\textbf{Model} & \textbf{BACC}& \textbf{macro-F1} & \textbf{macro-AUC} \\
\hline
MedSAM (ViT-B) & 55.75\% (52.80\%, 58.68\%) & 56.88\% (52.90\%, 60.63\%) & 68.10\% (63.74\%, 72.57\%) \\
LVM-Med (ViT-B) & 52.02\% (50.22\%, 53.97\%) & 51.34\% (48.37\%, 54.48\%) & 62.22\% (57.22\%, 66.96\%) \\
LVM-Med (R50) & 53.14\% (50.63\%, 55.84\%) & 53.45\% (50.12\%, 56.93\%) & 55.95\% (51.01\%, 60.98\%) \\
Mammo-CLIP (Enb2) & 67.63\% (63.65\%, 71.52\%) & 62.45\% (59.40\%, 65.43\%) & 75.04\% (70.87\%, 79.04\%) \\
Mammo-CLIP (Enb5) & 64.55\% (60.43\%, 68.71\%) & 62.83\% (59.27\%, 66.48\%) & 74.67\% (70.69\%, 78.95\%) \\
MAMA (ViT-B) & 73.02\% (68.90\%, 76.96\%) & \textbf{70.12\% (66.59\%, 73.67\%)} & 82.00\% (78.41\%, 85.43\%) \\
VersaMammo (Ours) & \textbf{74.80\% (70.81\%, 78.88\%)} & 68.99\% (65.92\%, 72.27\%) & \textbf{85.00\% (81.87\%, 87.69\%)} \\
$p$-value &$p<0.001$ &$p<0.001$ &$p<0.001$ \\
\hline
\end{tabular}
\label{tab:mastitis}
\end{table*}

\begin{table*}[h!]
\centering
\caption{Internal validation on invasive status determination on RSNA-Mammo for different FMs. Non-parametric bootstrapping with 1,000 bootstrap replicates is used for statistical analysis. The 95\% CI is included in parentheses. The best performing model for each metric is \textbf{bolded}. $p$-values are computed between the first-best and second-best performing models.} 
\mysmall
\begin{tabular}{lccc}
\hline
\textbf{Model} & \textbf{BACC}& \textbf{macro-F1} & \textbf{macro-AUC} \\
\hline
MedSAM (ViT-B) & 50.00\% (50.00\%, 50.00\%) & 49.56\% (49.56\%, 49.56\%) & 64.65\% (60.86\%, 68.58\%) \\
LVM-Med (ViT-B) & 50.00\% (50.00\%, 50.00\%) & 49.56\% (49.56\%, 49.56\%) & 65.90\% (62.00\%, 69.71\%) \\
LVM-Med (R50) & 50.00\% (50.00\%, 50.00\%) & 49.56\% (49.56\%, 49.56\%) & 59.83\% (55.89\%, 63.61\%) \\
Mammo-CLIP (Enb2) & 50.00\% (50.00\%, 50.00\%) & 49.56\% (49.56\%, 49.56\%) & 70.57\% (66.71\%, 74.25\%) \\
Mammo-CLIP (Enb5) & 51.49\% (50.43\%, 52.82\%) & 52.32\% (50.46\%, 54.54\%) & 66.69\% (62.50\%, 70.83\%) \\
MAMA (ViT-B) & \textbf{56.77\% (54.40\%, 59.21\%)} & \textbf{57.42\% (54.83\%, 59.90\%)} & 74.71\% (71.14\%, 78.04\%) \\
VersaMammo (Ours) & 54.18\% (52.35\%, 56.31\%) & 57.09\% (53.96\%, 60.51\%) & \textbf{80.05\% (76.91\%, 83.30\%)} \\
$p$-value &$p<0.001$ &$p<0.001$ &$p<0.001$ \\
\hline
\end{tabular}
\label{tab:invas}
\end{table*}
\begin{table*}[h!]
\centering
\caption{Internal validation on breast cancer molecular subtype prediction for different FMs. Non-parametric bootstrapping with 1,000 bootstrap replicates is used for statistical analysis. The 95\% CI is included in parentheses. The best performing model for each metric is \textbf{bolded}. $p$-values are computed between the first-best and second-best performing models.}
\mysmall
\begin{tabular}{lccc}
\hline
\textbf{Model} & \textbf{BACC}& \textbf{macro-F1} & \textbf{macro-AUC} \\
\hline

\multicolumn{4}{c}{CMMD}\\
MedSAM (ViT-B) & 44.94\% (39.24\%, 50.00\%) & 21.38\% (17.88\%, 25.16\%) & 56.85\% (52.26\%, 61.22\%) \\
LVM-Med (ViT-B) & \textbf{46.20\% (40.51\%, 51.91\%)} & 17.50\% (15.00\%, 20.38\%) & 54.64\% (50.48\%, 59.00\%) \\
LVM-Med (R50) & 43.35\% (37.97\%, 49.05\%) & 23.85\% (20.66\%, 27.00\%) & 59.38\% (55.34\%, 63.79\%) \\
Mammo-CLIP (Enb2) & 45.25\% (39.87\%, 50.63\%) & 29.55\% (24.46\%, 34.58\%) & \textbf{62.14\% (57.94\%, 66.48\%)} \\
Mammo-CLIP (Enb5) & 43.67\% (37.97\%, 49.05\%) & 23.33\% (19.12\%, 27.98\%) & 60.10\% (55.90\%, 64.07\%) \\
MAMA (ViT-B) & 43.35\% (37.97\%, 48.73\%) & 25.50\% (20.88\%, 29.84\%) & 59.03\% (55.04\%, 63.03\%) \\
VersaMammo (Ours) & 44.30\% (38.92\%, 49.69\%) & \textbf{31.57\% (26.26\%, 36.47\%)} & 61.31\% (56.82\%, 65.56\%) \\
$p$-value &$p<0.001$ &$p<0.001$ &$p<0.001$ \\
\hline
\multicolumn{4}{c}{PDS5}\\
MedSAM (ViT-B) & 24.93\% (24.81\%, 25.00\%) & 20.18\% (20.12\%, 20.22\%) & 52.84\% (49.08\%, 56.41\%) \\
LVM-Med (ViT-B) & 25.62\% (24.40\%, 27.37\%) & 21.64\% (19.94\%, 23.98\%) & 51.13\% (47.65\%, 54.83\%) \\
LVM-Med (R50) & 25.11\% (25.00\%, 25.34\%) & 20.44\% (20.20\%, 20.92\%) & 51.91\% (48.27\%, 55.25\%) \\
Mammo-CLIP (Enb2) & 26.04\% (24.89\%, 27.74\%) & 22.24\% (20.16\%, 25.15\%) & 55.41\% (51.33\%, 59.39\%) \\
Mammo-CLIP (Enb5) & 26.75\% (24.68\%, 29.30\%) & 23.53\% (20.62\%, 26.77\%) & 57.07\% (53.56\%, 60.65\%) \\
MAMA (ViT-B) & 25.00\% (23.73\%, 26.55\%) & 21.86\% (20.06\%, 24.04\%) & 57.91\% (54.61\%, 61.35\%) \\
VersaMammo (Ours) & \textbf{29.71\% (26.52\%, 32.75\%)} & \textbf{25.50\% (22.69\%, 28.18\%)} & \textbf{60.93\% (57.41\%, 64.46\%)} \\
$p$-value &$p<0.001$ &$p<0.001$ &$p<0.001$ \\
\hline
\end{tabular}
\label{tab:Subtype}
\end{table*}

\begin{table*}[h!]
\centering
\caption{Mammogram retrieval average accuracy (\%) for different FMs evaluated on merged internal and external datasets. Non-parametric bootstrapping with 1,000 bootstrap replicates is used for statistical analysis. The 95\% CI is included in parentheses. The best performing model for each metric is \textbf{bolded}. $p$-values are computed between the first-best and second-best performing models.}
\mysmall
\begin{tabular}{lccc}
\hline
\textbf{Model} & \textbf{ACC@1} & \textbf{ACC@2} & \textbf{ACC@3} \\
\hline
\multicolumn{4}{c}{3 internal (EMBED, RSNA-Mammo, VinDr-Mammo)}\\
\hline
MedSAM (ViT-B) & 50.52\% (50.22\%, 50.83\%) & 67.87\% (67.60\%, 68.15\%) & 75.65\% (75.38\%, 75.90\%) \\
LVM-Med (ViT-B) & 51.00\% (50.69\%, 51.32\%) & 68.20\% (67.92\%, 68.49\%) & 75.76\% (75.49\%, 76.01\%) \\
LVM-Med (R50) & 49.95\% (49.63\%, 50.26\%) & 67.22\% (66.95\%, 67.51\%) & 74.91\% (74.63\%, 75.15\%) \\
Mammo-CLIP (Enb2) & 52.81\% (52.51\%, 53.14\%) & 69.15\% (68.87\%, 69.45\%) & 76.38\% (76.12\%, 76.63\%) \\
Mammo-CLIP (Enb5) & 52.36\% (52.04\%, 52.65\%) & 68.88\% (68.58\%, 69.16\%) & 76.19\% (75.94\%, 76.45\%) \\
MAMA (ViT-B) & 53.82\% (53.53\%, 54.10\%) & 70.30\% (70.02\%, 70.58\%) & 77.40\% (77.14\%, 77.65\%) \\
VersaMammo (Ours) & \textbf{55.86\% (55.56\%, 56.17\%)} & \textbf{71.11\% (70.84\%, 71.38\%)} & \textbf{77.66\% (77.42\%, 77.91\%)} \\
$p$-value &$p<0.001$ &$p<0.001$ &$p<0.001$ \\
\hline
\multicolumn{4}{c}{5 external dataset (PDS1-PDS5)}\\
\hline
MedSAM (ViT-B) & 34.71\% (34.14\%, 35.31\%) & 51.19\% (50.54\%, 51.86\%) & 61.69\% (61.05\%, 62.33\%) \\
LVM-Med (ViT-B) & 34.32\% (33.71\%, 34.87\%) & 51.10\% (50.49\%, 51.70\%) & 61.81\% (61.16\%, 62.42\%) \\
LVM-Med (R50) & 32.73\% (32.16\%, 33.31\%) & 49.76\% (49.08\%, 50.40\%) & 60.70\% (60.08\%, 61.39\%) \\
Mammo-CLIP (Enb2) & 37.95\% (37.32\%, 38.54\%) & 54.52\% (53.86\%, 55.15\%) & 64.72\% (64.13\%, 65.29\%) \\
Mammo-CLIP (Enb5) & 37.62\% (37.00\%, 38.22\%) & 54.25\% (53.63\%, 54.88\%) & 64.28\% (63.69\%, 64.85\%) \\
MAMA (ViT-B) & 38.29\% (37.71\%, 38.88\%) & 55.20\% (54.57\%, 55.82\%) & 65.34\% (64.76\%, 65.96\%) \\
VersaMammo (Ours) & \textbf{39.03\% (38.43\%, 39.79\%)} & \textbf{58.70\% (58.12\%, 59.28\%)} & \textbf{68.31\% (67.71\%, 68.91\%)} \\
$p$-value &$p<0.001$ &$p<0.001$ &$p<0.001$ \\
\hline
\end{tabular}
\label{tab:retr}
\end{table*}

\begin{table*}[h!]
\centering
\caption{Comprehensive VQA Performance of different FMs for internal validation. Non-parametric bootstrapping with 1,000 bootstrap replicates is used for statistical analysis. The 95\% CI is included in parentheses. The best performing model for each metric is \textbf{bolded}. $p$-values are computed between the first-best and second-best performing models.}
\mysmall
\begin{tabular}{lccc}
\hline
\textbf{Model} & \textbf{BACC}& \textbf{macro-F1} & \textbf{macro-AUC} \\
\hline
\multicolumn{4}{c}{View}\\
MedSAM (ViT-B) & 96.40\% (96.11\%, 96.68\%) & 96.38\% (96.08\%, 96.66\%) & 98.59\% (98.41\%, 98.76\%) \\
LVM-Med (ViT-B) & 97.37\% (97.13\%, 97.60\%) & 97.36\% (97.11\%, 97.59\%) & 99.45\% (99.35\%, 99.54\%) \\
LVM-Med (R50) & 50.00\% (50.00\%, 50.00\%) & 33.66\% (33.30\%, 34.00\%) & 49.75\% (48.84\%, 50.65\%) \\
Mammo-CLIP (Enb2) & 96.70\% (96.43\%, 96.97\%) & 96.67\% (96.40\%, 96.94\%) & 99.04\% (98.89\%, 99.19\%) \\
Mammo-CLIP (Enb5) & 97.37\% (97.13\%, 97.63\%) & 97.36\% (97.12\%, 97.62\%) & 97.98\% (97.71\%, 98.26\%) \\
MAMA (ViT-B) & 98.76\% (98.60\%, 98.94\%) & 98.77\% (98.60\%, 98.94\%) & 99.86\% (99.82\%, 99.90\%) \\
\hline
\end{tabular}
\label{tab:vqa}
\end{table*}

\begin{table*}[h!]
\centering
\caption{Comprehensive VQA performance of different FMs for external validation. Non-parametric bootstrapping with 1,000 bootstrap replicates is used for statistical analysis. The 95\% CI is included in parentheses. The best performing model for each metric is \textbf{bolded}. $p$-values are computed between the first-best and second-best performing models.}
\mysmall
\renewcommand{\arraystretch}{1.1}
\begin{tabular}{lccc}
\hline
\textbf{Model} & \textbf{BACC}& \textbf{macro-F1} & \textbf{macro-AUC} \\
\hline
\multicolumn{4}{c}{Breast Composition Assessment}\\
MedSAM (ViT-B) & 28.27\% (27.39\%, 29.12\%) & 13.03\% (12.09\%, 14.01\%) & 68.73\% (67.31\%, 70.13\%) \\
LVM-Med (ViT-B) & 28.95\% (27.98\%, 29.90\%) & 14.35\% (13.38\%, 15.28\%) & 67.44\% (65.94\%, 68.72\%) \\
LVM-Med (R50) & 25.00\% (25.00\%, 25.00\%) & 5.31\% (4.75\%, 5.88\%) & 48.80\% (46.97\%, 50.85\%) \\
Mammo-CLIP (Enb2) & 28.03\% (26.69\%, 29.29\%) & 16.09\% (15.12\%, 17.09\%) & 61.11\% (59.33\%, 63.06\%) \\
Mammo-CLIP (Enb5) & 31.54\% (30.07\%, 32.79\%) & 19.65\% (18.53\%, 20.71\%) & 68.28\% (66.73\%, 69.73\%) \\
MAMA (ViT-B) & \textbf{39.83\% (37.44\%, 42.55\%)} & \textbf{32.07\% (29.66\%, 34.75\%)} & 73.55\% (72.25\%, 74.94\%) \\
\hline
\end{tabular}
\label{tab:vqa_ex}
\end{table*}

\begin{table*}[t]
\centering
\caption{Subgroup analysis on breast composition for different FMs. Non-parametric bootstrapping with 1,000 bootstrap replicates is used for statistical analysis. The 95\% CI is included in parentheses. The best performing model for each metric is \textbf{bolded}. $p$-values are computed between the first-best and second-best performing models.}
\mysmall
\begin{tabular}{lccc}
\hline
\textbf{Model} & \textbf{BACC} & \textbf{macro-F1} & \textbf{macro-AUC} \\
\hline
\multicolumn{4}{c}{\textbf{Density A}} \\
\hline
MedSAM (ViT-B) & 27.68\% (25.77\%, 29.74\%) & 65.82\% (64.68\%, 66.88\%) & 68.41\% (67.32\%, 69.49\%) \\
LVM-Med (ViT-B) & 26.17\% (25.10\%, 27.22\%) & 67.20\% (66.11\%, 68.24\%) & 70.09\% (69.06\%, 71.15\%) \\
LVM-Med (R50) & 23.92\% (21.99\%, 26.26\%) & 64.45\% (63.30\%, 65.50\%) & 64.47\% (63.36\%, 65.56\%) \\
Mammo-CLIP (Enb2) & 30.48\% (28.10\%, 33.10\%) & 68.04\% (66.95\%, 69.07\%) & 70.99\% (70.03\%, 72.02\%) \\
Mammo-CLIP (Enb5) & 31.40\% (28.73\%, 34.37\%) & 67.36\% (66.26\%, 68.37\%) & 72.10\% (71.07\%, 73.20\%) \\
MAMA (ViT-B) & 32.90\% (29.99\%, 36.01\%) & 68.22\% (67.10\%, 69.26\%) & 74.59\% (73.57\%, 75.57\%) \\
\hline
\end{tabular}
\label{tab:breast_composition}
\end{table*}

\begin{table*}[t]
\centering
\caption{Subgroup analysis on age groups for different FMs. Non-parametric bootstrapping with 1,000 bootstrap replicates is used for statistical analysis. The 95\% CI is included in parentheses. The best performing model for each metric is \textbf{bolded}. $p$-values are computed between the first-best and second-best performing models.}
\mysmall
\begin{tabular}{lccc}
\hline
\textbf{Model} & \textbf{BACC} & \textbf{macro-F1} & \textbf{macro-AUC} \\
\hline

\multicolumn{4}{c}{\textbf{<40}} \\
\hline
MedSAM (ViT-B) & 22.44\% (21.67\%, 23.24\%) & 39.03\% (37.86\%, 40.09\%) & 62.00\% (61.17\%, 62.77\%) \\
LVM-Med (ViT-B) & 21.15\% (20.51\%, 21.75\%) & 39.51\% (38.36\%, 40.60\%) & 63.12\% (62.32\%, 63.96\%) \\
LVM-Med (R50) & 19.83\% (19.26\%, 20.44\%) & 37.25\% (36.16\%, 38.36\%) & 60.75\% (59.88\%, 61.62\%) \\
Mammo-CLIP (Enb2) & 23.81\% (22.99\%, 24.71\%) & 39.07\% (37.92\%, 40.25\%) & 64.95\% (64.13\%, 65.83\%) \\
Mammo-CLIP (Enb5) & 24.46\% (23.69\%, 25.29\%) & 37.98\% (36.85\%, 39.14\%) & 65.27\% (64.49\%, 66.08\%) \\
MAMA (ViT-B) & 24.15\% (23.40\%, 24.97\%) & 42.45\% (41.34\%, 43.56\%) & 66.79\% (65.98\%, 67.62\%) \\
\hline

\end{tabular}
\label{tab:age}
\end{table*}

\begin{table*}[t]
\centering
\caption{Subgroup analysis on clinical scenarios for different FMs. Non-parametric bootstrapping with 1,000 bootstrap replicates is used for statistical analysis. The 95\% CI is included in parentheses. The best performing model for each metric is \textbf{bolded}. $p$-values are computed between the first-best and second-best performing models.}
\mysmall
\begin{tabular}{lccc}
\hline
\textbf{Model} & \textbf{BACC} & \textbf{macro-F1} & \textbf{macro-AUC} \\
\hline
\multicolumn{4}{c}{\textbf{Screening}} \\
\hline
MedSAM (ViT-B) & 27.05\% (26.38\%, 27.71\%) & 60.79\% (60.43\%, 61.17\%) & 65.76\% (65.43\%, 66.09\%) \\
LVM-Med (ViT-B) & 25.57\% (25.22\%, 25.94\%) & 62.65\% (62.28\%, 63.01\%) & 68.05\% (67.72\%, 68.39\%) \\
LVM-Med (R50) & 22.80\% (22.17\%, 23.50\%) & 59.21\% (58.85\%, 59.58\%) & 63.90\% (63.55\%, 64.23\%) \\
Mammo-CLIP (Enb2) & 29.11\% (28.21\%, 30.10\%) & 63.01\% (62.64\%, 63.36\%) & 69.55\% (69.23\%, 69.88\%) \\
Mammo-CLIP (Enb5) & 29.38\% (28.55\%, 30.36\%) & 62.86\% (62.48\%, 63.26\%) & 69.91\% (69.59\%, 70.25\%) \\
MAMA (ViT-B) & 31.04\% (29.96\%, 32.18\%) & 64.04\% (63.68\%, 64.43\%) & 71.86\% (71.54\%, 72.19\%) \\
VersaMammo (Ours) & \textbf{38.68\% (37.20\%, 40.28\%)} & \textbf{65.69\% (65.33\%, 66.07\%)} & \textbf{74.32\% (73.98\%, 74.65\%)} \\
$p$-value &$p<0.001$ &$p<0.001$ &$p<0.001$ \\
\hline
\multicolumn{4}{c}{\textbf{Diagnostic}} \\
\hline
MedSAM (ViT-B) & 25.06\% (24.87\%, 25.26\%) & 29.62\% (29.36\%, 29.88\%) & 65.49\% (65.31\%, 65.67\%) \\
LVM-Med (ViT-B) & 25.97\% (25.72\%, 26.21\%) & 34.31\% (34.04\%, 34.57\%) & 68.13\% (67.95\%, 68.32\%) \\
LVM-Med (R50) & 22.31\% (22.03\%, 22.63\%) & 25.49\% (25.24\%, 25.74\%) & 63.45\% (63.26\%, 63.64\%) \\
Mammo-CLIP (Enb2) & 29.98\% (29.55\%, 30.49\%) & 34.37\% (34.10\%, 34.62\%) & 69.21\% (69.04\%, 69.38\%) \\
Mammo-CLIP (Enb5) & 28.49\% (28.12\%, 28.86\%) & 32.89\% (32.61\%, 33.16\%) & 68.98\% (68.81\%, 69.15\%) \\
MAMA (ViT-B) & 29.43\% (29.05\%, 29.84\%) & 36.33\% (36.06\%, 36.60\%) & 71.62\% (71.45\%, 71.80\%) \\
VersaMammo (Ours) & \textbf{36.14\% (35.63\%, 36.67\%)} & \textbf{39.35\% (39.10\%, 39.61\%)} & \textbf{73.69\% (73.52\%, 73.86\%)} \\
$p$-value &$p<0.001$ &$p<0.001$ &$p<0.001$ \\
\hline
\end{tabular}
\label{tab:clinical_scenario}
\end{table*}

\begin{table*}[t]
\centering
\caption{Subgroup analysis on ethnicity for different FMs. Non-parametric bootstrapping with 1,000 bootstrap replicates is used for statistical analysis. The 95\% CI is included in parentheses. The best performing model for each metric is \textbf{bolded}. $p$-values are computed between the first-best and second-best performing models.}
\mysmall
\begin{tabular}{lccc}
\hline
\textbf{Model} & \textbf{BACC} & \textbf{macro-F1} & \textbf{macro-AUC} \\
\hline

\multicolumn{4}{c}{\textbf{African American or Black}} \\
\hline
MedSAM (ViT-B) & 26.39\% (26.14\%, 26.63\%) & 50.02\% (49.78\%, 50.23\%) & 68.39\% (68.22\%, 68.56\%) \\
LVM-Med (ViT-B) & 26.65\% (26.35\%, 26.98\%) & 52.22\% (51.99\%, 52.45\%) & 70.64\% (70.48\%, 70.81\%) \\
LVM-Med (R50) & 23.49\% (23.13\%, 23.85\%) & 47.75\% (47.53\%, 47.99\%) & 66.15\% (65.96\%, 66.33\%) \\
Mammo-CLIP (Enb2) & 30.06\% (29.56\%, 30.60\%) & 52.88\% (52.65\%, 53.09\%) & 71.76\% (71.61\%, 71.93\%) \\
Mammo-CLIP (Enb5) & 30.06\% (29.57\%, 30.52\%) & 52.28\% (52.05\%, 52.51\%) & 71.74\% (71.58\%, 71.90\%) \\
MAMA (ViT-B) & 30.68\% (30.25\%, 31.14\%) & 54.17\% (53.96\%, 54.40\%) & 74.45\% (74.29\%, 74.61\%) \\
\hline

\end{tabular}
\label{tab:ethnicity}
\end{table*}

\begin{table*}[t]
\centering
\caption{{Fairness evaluation across different foundation models under ethnicity subgroup analysis. Lower values indicate smaller performance disparities across demographic subgroups and better fairness. The best performing model for each metric is \textbf{bolded}.}}
\mysmall
\begin{tabular}{lcccc}
\hline
\textbf{Model} & \textbf{TPR Gap $\downarrow$} & \textbf{FPR Gap $\downarrow$} & \textbf{EOD $\downarrow$} & \textbf{AOD $\downarrow$} \\
\hline
MedSAM (ViT-B) & 0.2961 & 0.0231 & 0.2961 & 0.1596 \\
LVM-Med (ViT-B) & 0.1087 & 0.0142 & 0.1087 & 0.0615 \\
LVM-Med (R50) & \textbf{0.0730} & 0.0137 & \textbf{0.0730} & \textbf{0.0434} \\
Mammo-CLIP (Enb2) & 0.2845 & 0.0212 & 0.2845 & 0.1529 \\
Mammo-CLIP (Enb5) & 0.3734 & 0.0220 & 0.3734 & 0.1977 \\
MAMA (ViT-B) & 0.1475 & \textbf{0.0067} & 0.1475 & 0.0771 \\
VersaMammo (Ours) & 0.0848 & 0.0309 & 0.0848 & 0.0579 \\
\hline
\end{tabular}
\label{tab:fairness_metrics}
\end{table*}

\clearpage
{\mysmall
\begin{longtable}{llccc}
\caption{The effectiveness of different components of the VersaMammo. The 95\% CI is included in parentheses. The best performing model for each metric is \textbf{bolded}. $p$-values are computed between the first-best and second-best performing models.}\\
\hline
& \textbf{Method} & \textbf{BACC}& \textbf{macro-F1} & \textbf{macro-AUC} \\
\hline
\endfirsthead

\multicolumn{5}{c}%
{{\bfseries \tablename\ \thetable{} - continued from previous page}} \\
\hline
& \textbf{Method} & \textbf{BACC}& \textbf{macro-F1} & \textbf{macro-AUC} \\
\hline
\endhead

\hline \multicolumn{5}{r}{{Continued on next page}} \\
\endfoot

\hline
\endlastfoot

\multicolumn{5}{c}{BI-RADS  (RSNA-Mammo)}\\
 & VersaMammo (Ours) & {57.03\%	 (55.26\%, 58.86\%)} & \textbf{{59.91\%	 (57.93\%, 61.84\%)}} & \textbf{{79.92\%	 (78.74\%, 81.04\%)}} \\
 &Task-specific & 45.73\% (44.38\%, 47.09\%) & 47.01\% (45.15\%, 49.05\%) & 71.19\% (69.78\%, 72.61\%) \\ 
& w/o Mammogram& 33.29\%	 (33.23\%, 33.32\%) & 24.82\%	 (24.80\%, 24.84\%) & 50.55\%	 (49.15\%, 52.06\%) \\
 & w/o Stage 2 & 54.64\%	 (52.95\%, 56.31\%) & 57.07\%	 (55.27\%, 58.81\%) & 74.61\%	 (73.25\%, 75.89\%) \\
 & w/o Distillation & 56.44\%	 (54.81\%, 58.01\%) & 50.21\%	 (48.65\%, 51.60\%) & 78.67\%	 (77.51\%, 79.78\%) \\
 & w/o Supervision & \textbf{57.77\%	 (56.06\%, 59.42\%)} & 53.03\%	 (51.42\%, 54.59\%) & 79.16\%	 (78.03\%, 80.33\%) \\
 & w/o CNN & 54.20\% (52.45\%, 55.96\%) & 55.46\% (53.65\%, 57.22\%) & 74.39\% (72.91\%, 75.69\%) \\
\end{longtable}
}

\begin{table*}[htbp]
\centering
\caption{The effectiveness of different components of the VersaMammo pre-training on INbreast segmentation and detection tasks (both tasks and dataset were not seen in pre-training). The 95\% CI is included in parentheses. The best performing model for each metric is \textbf{bolded}. $p$-values are computed between the first-best and second-best performing models.}
\label{tab:ablation_segmentation_detection}

\mysmall
\begin{tabular}{llccc}
\hline
& \textbf{Method} & \textbf{Segmentation(Dice)} & & \textbf{Detection(IoU)} \\
\hline

& VersaMammo (Ours) & 74.53\% (59.89\%, 89.17\%) & & 71.20\% (65.44\%, 75.47\%) \\
& Task-specific & 74.54\% (58.55\%, 88.09\%) & & 62.86\% (49.69\%, 74.08\%) \\
& w/o Mammogram & 68.97\% (51.92\%, 82.57\%) & & 50.10\% (35.47\%, 63.39\%) \\
& w/o Stage 2 & 60.96\% (45.61\%, 75.72\%) & & 40.96\% (25.32\%, 57.24\%) \\
& w/o Distillation & \textbf{78.58\% (66.19\%, 89.21\%)} & & \textbf{71.63\% (65.42\%, 77.02\%)} \\
& w/o Supervision & 76.93\% (63.35\%, 88.11\%) & & 67.91\% (54.72\%, 78.47\%) \\
& w/o CNN & 64.38\% (46.73\%, 79.55\%) & & 50.95\% (36.25\%, 64.80\%) \\
& $p$-value & $p<0.001$ & & $p<0.001$ \\
\hline
\end{tabular}
\end{table*}

\begin{table*}[h!]
\centering

\caption{Analysis of the impact of pre-training sample size on model performance. Non-parametric bootstrapping with 1,000 bootstrap replicates is used for statistical analysis. The 95\% CI is included in parentheses. The best performing model for each metric is \textbf{bolded}. $p$-values are computed between the first-best and second-best performing models.}
\mysmall
\begin{tabular}{lccc}
\hline
$n_{pre-train}$& \textbf{BACC} & \textbf{macro-F1} & \textbf{macro-AUC} \\
\hline
\multicolumn{4}{c}{BI-RADS (VinDr-Mammo)} \\

1,000& 31.55\% (29.58\%, 33.63\%) & 32.76\% (30.44\%, 35.02\%) & 67.53\% (65.98\%, 69.01\%) \\
5,000& 34.42\% (32.56\%, 36.46\%) & 36.89\% (34.58\%, 39.15\%) & 71.30\% (69.89\%, 72.64\%) \\
10,000& 39.66\% (37.39\%, 41.76\%) & 43.31\% (40.81\%, 45.76\%) & 75.76\% (74.38\%, 77.08\%) \\
50,000& 47.50\% (45.14\%, 49.76\%) & 51.62\% (49.04\%, 53.98\%) & 81.36\% (80.23\%, 82.50\%) \\
100,000& 49.37\% (46.90\%, 51.71\%) & 53.41\% (50.79\%, 55.83\%) & 82.83\% (81.73\%, 83.97\%) \\
\text{all}(365,428)& \textbf{50.70\% (48.47\%, 52.97\%)} & \textbf{54.25\% (51.81\%, 56.62\%)} & \textbf{84.69\% (83.51\%, 85.77\%)} \\
$p$-value &$p<0.001$ &$p<0.001$ &$p<0.001$ \\ 
\hline
\multicolumn{4}{c}{Breast Composition (VinDr-Mammo)} \\

1,000& 57.05\% (52.35\%, 61.94\%) & 61.02\% (55.91\%, 65.85\%) & 91.11\% (90.27\%, 91.88\%) \\
5,000& 58.09\% (53.20\%, 63.42\%) & 60.94\% (56.04\%, 65.61\%) & 91.21\% (90.23\%, 92.13\%) \\
10,000& 57.35\% (52.99\%, 62.02\%) & 59.54\% (54.86\%, 64.23\%) & 91.89\% (91.15\%, 92.58\%) \\
50,000& 61.50\% (56.68\%, 66.69\%) & 65.09\% (59.94\%, 69.97\%) & 93.14\% (92.52\%, 93.84\%) \\
100,000& 67.86\% (62.87\%, 72.57\%) & 68.27\% (64.45\%, 72.16\%) & 94.19\% (93.57\%, 94.76\%) \\
{all}(365,428)& \textbf{70.67\% (66.16\%, 75.02\%)} & \textbf{70.18\% (66.66\%, 73.85\%)} & \textbf{94.98\% (94.46\%, 95.51\%)} \\
$p$-value &$p<0.001$ &$p<0.001$ &$p<0.001$ \\ 
\hline
\end{tabular}
\label{tab:sample_size_impact}
\end{table*}

\begin{table*}[h!]
\centering

\caption{Performance analysis of the reader study by seniority. Non-parametric bootstrapping with 1,000 bootstrap replicates is used for statistical analysis. The 95\% CI is included in parentheses. The best performing model for each metric is \textbf{bolded}.}
\label{tab:seniority}
\mysmall
\begin{tabular}{lccc}
\toprule
\textbf{Group} & \textbf{BACC} & \textbf{macro-F1} & \textbf{Time (s/image)} \\
\midrule

Junior (No AI) &
35.48\% (32.26\%, 38.97\%) &
36.47\% (33.12\%, 39.82\%) &
60.70 $\pm$ 22.95 \\
Junior (With AI) &
50.55\% (47.22\%, 53.92\%) &
52.51\% (49.27\%, 55.67\%) &
32.60 $\pm$ 5.79 \\
\addlinespace

Mid (No AI) &
45.72\% (42.67\%, 49.31\%) &
46.07\% (43.03\%, 49.43\%) &
38.20 $\pm$ 15.63 \\
Mid (With AI) &
\textbf{55.39\% (51.82\%, 58.92\%)} &
\textbf{56.43\% (53.06\%, 59.56\%)} &
32.82 $\pm$ 28.15 \\
\addlinespace

Senior (No AI) &
51.12\% (47.72\%, 54.40\%) &
51.10\% (47.64\%, 54.22\%) &
52.18 $\pm$ 40.15 \\
Senior (With AI) &
55.35\% (52.30\%, 58.39\%) &
54.97\% (52.01\%, 58.01\%) &
24.13 $\pm$ 12.16 \\
\addlinespace

Only AI &
52.47\% (48.29\%, 56.93\%) &
52.94\% (48.20\%, 57.37\%) &
\textbf{1.50} \\
\bottomrule
\end{tabular}
\end{table*}

\begin{table*}[h!]
\centering

\caption{Performance analysis of the reader study by individual. Non-parametric bootstrapping with 1,000 bootstrap replicates is used for statistical analysis. The 95\% CI is included in parentheses. The best performing model for each metric is \textbf{bolded}.}
\label{tab:individual}
\mysmall
\begin{tabular}{lccc}
\toprule
\textbf{Evaluator} & \textbf{BACC} & \textbf{macro-F1} & \textbf{Time (s/image)} \\
\midrule

A-J (No AI) & 34.58\% (30.66\%, 38.70\%) & 35.39\% (30.20\%, 40.15\%) & 46.44 $\pm$ 8.74 \\
A-J (With AI) & 53.32\% (49.06\%, 57.74\%) & 49.87\% (45.57\%, 54.60\%) & 32.60 $\pm$ 5.79 \\

B-J (No AI) & 36.46\% (32.18\%, 41.07\%) & 31.19\% (27.22\%, 34.94\%) & 60.70 $\pm$ 22.95 \\
B-J (With AI) & 56.57\% (51.94\%, 61.09\%) & 53.69\% (48.97\%, 57.62\%) & 26.25 $\pm$ 8.62 \\

A-M (No AI) & 41.87\% (37.54\%, 46.47\%) & 41.85\% (37.18\%, 46.55\%) & 47.19 $\pm$ 26.74 \\
A-M (With AI) & \textbf{58.03\% (53.22\%, 62.56\%)} & 55.75\% (51.30\%, 60.05\%) & 32.82 $\pm$ 28.15 \\

B-M (No AI) & 49.50\% (44.72\%, 53.79\%) & 48.79\% (44.12\%, 52.85\%) & 38.20 $\pm$ 15.63 \\
B-M (With AI) & 56.78\% (52.03\%, 61.26\%) & \textbf{56.68\% (52.10\%, 61.31\%)} & 28.10 $\pm$ 7.21 \\

A-S (No AI) & 51.91\% (47.45\%, 56.43\%) & 51.43\% (46.55\%, 55.70\%) & 25.11 $\pm$ 8.66 \\
A-S (With AI) & 57.81\% (53.84\%, 61.75\%) & 55.63\% (51.40\%, 59.47\%) & 24.13 $\pm$ 12.16 \\

B-S (No AI) & 50.33\% (45.32\%, 55.23\%) & 50.83\% (45.86\%, 55.38\%) & 52.18 $\pm$ 40.15 \\
B-S (With AI) & 53.09\% (48.87\%, 57.29\%) & 54.01\% (49.30\%, 58.24\%) & 26.87 $\pm$ 10.51 \\

Only AI & 52.47\% (48.29\%, 56.93\%) & 52.94\% (48.20\%, 57.37\%) & \textbf{1.50} \\
\bottomrule
\end{tabular}
\end{table*}

\begin{table*}[h!]
\centering

\caption{Consistency analysis of the reader study. Non-parametric bootstrapping with 1,000 bootstrap replicates is used for statistical analysis. The 95\% CI is included in parentheses. The best performing model for each metric is \textbf{bolded}.}
\label{tab:kappa}
\mysmall
\begin{tabular}{lc}
\toprule
\textbf{Group} & \textbf{Cohen's Kappa} \\
\midrule

Junior (No AI) & 0.25 (0.20, 0.29) \\
Junior (With AI) & 0.50 (0.44, 0.56) \\
\addlinespace

Mid (No AI) & 0.65 (0.59, 0.70) \\
Mid (With AI) & 0.71 (0.66, 0.76) \\
\addlinespace

Senior (No AI) & 0.76 (0.71, 0.81) \\
Senior (With AI) & \textbf{0.79 (0.74, 0.84)} \\
\addlinespace

All Evaluators (No AI) & 0.55 (0.53, 0.56) \\
All Evaluators (With AI) & 0.67 (0.66, 0.69) \\
\bottomrule
\end{tabular}
\end{table*}

\begin{table*}[h]
\centering
\caption{The number of patients and mammograms in private datasets (PDS).}
\label{tab:pds_stats}
\begin{tabular}{lrr}
\toprule
\textbf{Dataset} & \textbf{Patient} & \textbf{Mammogram} \\
\hline
PDS1 & 413 & 1,586 \\
PDS2 & 1,502 & 6,131 \\
PDS3 & 402 & 1,733 \\
PDS4 & 9,464 & 38,580 \\
PDS5 & 3,439 & 28,216 \\
\hline
Total& 15,220 & 76,246\\
\hline
\end{tabular}
\end{table*}

\begin{table*}[h]

\centering
\caption{Dataset splits. 
All datasets are mutually exclusive, ensuring no data leakage between training, validation, and test splits. 
All test datasets were used for internal validation, while \underline{underline} indicates those training splits were also used for pre-training the foundation model.
}

\label{tab:dataset_splits}
\begin{tabular}{l|cccccc}
\toprule
\multirow{2}{*}{\diagbox[width=14em]{Dataset}{Split}} & \multicolumn{3}{c}{Image Level} & \multicolumn{3}{c}{Patient Level} \\
\cmidrule(lr){2-4} \cmidrule(lr){5-7}
& Training & Validation & Test & Training & Validation & Test \\
\hline

{EMBED} & \underline{336,756} & 49,157 & 94,362 & \underline{16,278} & 2,324 & 4,651 \\
{RSNA} & \underline{38,344} & 5,454 & 10,908 & \underline{8,340} & 1,191 & 2,382 \\
{VinDr-Mammo} & \underline{14,000} & 2,000 & 4,000 & \underline{3,500} & 500 & 1,000 \\
LAMIS & 1,548 & 221 & 443 & 581 & 191 & 341 \\
DBT & 94 & 13 & 28 & - & - & - \\
MIAS&224	&33	&65	&-	&-	&- \\
MM & 511 & 73 & 161 & - & - & - \\
NLBS & 16,593 & 2,369 & 4,744 & - & - & - \\
BMCD & 280 & 40 & 80 & 100 & 33 & 61 \\
INbreast & 286 & 42 & 82 & - & - & - \\
KAU-BCMD & 1,662 & 236 & 476 & 504 & 191 & 327 \\
CBIS-DDSM & 2,242 & 285 & 576 & 1,370 & 266 & 522 \\
CMMD & 2,618 & 376 & 750 & - & - & - \\
CDD-CESM & 703 & 100 & 200 & 305 & 87 & 158 \\
CSAW-M & 7,014 & 1,002 & 2,004 & - & - & - \\
DMID-breast & 357 & 51 & 102 & - & - & - \\
PDS1 & 1,105 & 162 & 319 & 290 & 41 & 84 \\
PDS2 & 4,290 & 612 & 1,229 & 1,486 & 523 & 894 \\
PDS3 & 1,213 & 173 & 347 & 401 & 142 & 241 \\
PDS4 & 27,005 & 3,858 & 7,717 & 9,387 & 3,295 & 5,628 \\
PDS5 & 11,677 & 1,628 & 14,911 & 2,502 & 391 & 779 \\

\bottomrule
\end{tabular}
\end{table*}

\begin{table*}[h]

\centering
\caption{ Experimental setup for external validation on independent test datasets. ``+'' indicates merged datasets.
}

\label{tab:external_setup}
\begin{tabular} {p{4.5cm}p{3cm}p{3cm}p{2.5cm}}
\toprule

Task & Training & Validation & Test \\
\hline

Lesion detection & INbreast$_{train}$  $+$\newline VinDr-Mammo$_{train}$ & INbreast$_{val}$ $+$\newline VinDr-Mammo$_{val}$ & CBIS-DDSM$_{test}$ \\
\hline
Lesion segmentation & INbreast$_{train}$ $+$\newline CBIS-DDSM$_{train}$ & INbreast$_{val}$ $+$\newline CBIS-DDSM$_{val}$ & PDS5$_{test}$ \\
\hline
Breast composition assessment & EMBED$_{train}$ $+$\newline RSNA-Mammo$_{train}$ $+$\newline VinDr-Mammo$_{train}$ & EMBED$_{val}$ $+$\newline RSNA-Mammo$_{val}$ $+$\newline VinDr-Mammo$_{val}$ & CDD-CESM$_{test}$,\newline PDS1$_{test}$,\newline PDS2$_{test}$,\newline PDS3$_{test}$,\newline PDS4$_{test}$,\newline PDS5$_{test}$ \\
\hline
BI-RADS assessment & EMBED$_{train}$ $+$\newline RSNA-Mammo$_{train}$ $+$\newline VinDr-Mammo$_{train}$ & EMBED$_{val}$ $+$\newline RSNA-Mammo$_{val}$ $+$\newline VinDr-Mammo$_{val}$ & CDD-CESM$_{test}$,\newline PDS1$_{test}$,\newline PDS2$_{test}$,\newline PDS3$_{test}$,\newline PDS4$_{test}$,\newline PDS5$_{test}$ \\
\hline
Pathological diagnosis prediction & RSNA-Mammo$_{train}$ & RSNA-Mammo$_{val}$ & CDD-CESM$_{test}$,\newline DMID$_{test}$,\newline CMMD$_{test}$,\newline MIAS$_{test}$,\newline PDS1$_{test}$,\newline PDS2$_{test}$,\newline PDS3$_{test}$,\newline PDS5$_{test}$ \\
\hline
Mammogram retrieval & EMBED$_{train}$ $+$\newline RSNA-Mammo$_{train}$ $+$\newline VinDr-Mammo$_{train}$ & EMBED$_{val}$ $+$\newline RSNA-Mammo$_{val}$ $+$\newline VinDr-Mammo$_{val}$ & PDS1$_{test}$ $+$\newline PDS2$_{test}$ $+$\newline PDS3$_{test}$ $+$\newline PDS4$_{test}$ $+$\newline PDS5$_{test}$ \\
\hline
VQA & MammoVQA benchmark$_{train}$ & MammoVQA benchmark$_{val}$ & MammoVQA benchmark$_{test}$ \\

\bottomrule
\end{tabular}
\end{table*}

\begin{table*}[htbp]
\centering

\caption{ Demographic data and clinical context of datasets used in this study. ``-'' indicates that the metadata is unavailable.}
\label{tab:dataset_demographic}
\begin{tabular}{@{}lcccccl@{}}
\toprule
Dataset & Age & {Ethnicity} & Clinical Context & Manufacturer & Source \\
\hline
BMCD & - & - & Diagnostic & - & Catholic Univ. Korea \\
CBIS-DDSM & - & - & Screening & Lumisys/Howtek & MGH, Wake Forest \\
CDD-CESM & 18-91 & - & Diagnostic & GE & Cairo Univ. \\
CMMD & 18-87  & Chinese & Diagnostic & GE/Siemens & SYSUCC \\
CSAW-M & 40-74 & Swedish & Screening & - & Stockholm Region \\
DBT & - & - & Screening & Hologic & Duke Univ. \\
DMID & - & Indian & Both & - & Samved Hospital \\
EMBED & 18-100+ & Multi-racial & Both & Hologic/GE & Emory Univ. \\
INbreast & - & - & Both & Siemens & CHSJ Porto \\
KAU-BCMD & - & Saudi & Both & - & KAU \\
LAMIS & - & - & Both & - & SMBA Univ. \\
MIAS & - & - & Screening & Joyce-Loebl & MIAS Society \\
NLBS & 40-50+ & Canadian & Screening & - & Newfoundland \\
RSNA & 16-89  & Diverse & Screening & Hologic & Emory/Australian \\
VinDr-Mammo & 18-91 & Vietnamese & Screening & Planmed/Siemens & Hanoi Med. Univ. \\
PDS1 & - & Chinese & Both & Hologic/GE & Shenzhen PH \\
PDS2 & 19-82 & Chinese & Both & Planmed & Guangzhou 1st PH \\
PDS3 & 19-92 & Chinese & Both & Siemens & SYSU 3rd Hosp. \\
PDS4 & 13-88 & Chinese & Both & Hologic & SYSU Mem. Hosp. \\
PDS5 & 12-94 & Chinese & Both & Hologic/GE & Yunnan CH \\

\bottomrule
\end{tabular}
\end{table*}

\begin{table*}[ht]
\centering

\caption{Available labels for downstream tasks of different datasets.}
\label{tab:dataset_annotation_info}
\begin{tabular}{p{2.2cm}p{11cm}}
\toprule
\textbf{Dataset} & \textbf{Available labels} \\
\hline
CBIS-DDSM & Detection, segmentation, breast composition, finding, BI-RADS, and pathology labels\\
INbreast & Detection, segmentation, breast composition, finding, BI-RADS, and pathology labels \\
VinDr-Mammo & Detection, breast composition, finding, and BI-RADS labels \\
EMBED & Breast composition, finding, and BI-RADS labels \\
RSNA-Mammo & Breast composition, BI-RADS pathology, and invasive labels\\
MIAS & Breast composition, finding, and pathology labels \\
DMID & Breast composition, finding, and  pathology labels \\
CDD-CESM & Breast composition, BI-RADS, and  pathology labels\\
CMMD & Finding, and pathology and molecular subtype labels \\
BMCD & Breast composition and BI-RADS labels\\
KAU-BCMD & BI-RADS labels \\
LAMIS & Breast composition, BI-RADS, and  pathology labels \\
CSAW-M & Masking risk level labels \\
NLBS & Pathology labels \\
DBT & Pathology labels \\
MM & Pathology labels \\
PDS1 & Breast composition, finding,  BI-RADS, and  pathology labels  \\
PDS2 &  Breast composition, BI-RADS, BI-RADS 4 subcategory, and pathology labels\\
PDS3 &  Breast composition, BI-RADS, pathology labels\\
PDS4 &   Breast composition, BI-RADS.BI-RADS 4 subcategory, and pathology labels \\
PDS5 & Segmentation, breast composition, finding, BI-RADS, BI-RADS 4 subcategory, pathology, mastitis, and molecular subtype labels  \\
\bottomrule
\end{tabular}
\end{table*}

\begin{table*}[h!]
\centering
\caption{Distribution of breast composition label (Background tissue categories)}
\label{tab:composition_fatty_sta}
\begin{tabular}{l|rrr}
\hline
\diagbox[width=10em]{Dataset}{Label} & Fatty-glandular & Dense-glandular & Fatty \\
\hline
DMID & 426 & 16 & 66 \\
MIAS & 104 & 112 & 106 \\
\hline
\end{tabular}
\end{table*}

\begin{table*}[h!]
\centering
\caption{Distribution of breast composition label (Breast density A-D categories)}
\label{tab:composition_density}
\begin{tabular}{l|rrrr}
\hline
\diagbox[width=10em]{Dataset}{Label} &Density A &Density B &Density C &Density D \\
\hline
BMCD & 40 & 228 & 112 & 20 \\
CBIS-DDSM & 425 & 1,207 & 952 & 517 \\
CDD-CESM & 8 & 70 & 515 & 329 \\
INbreast & 131 & 144 & 97 & 27 \\
EMBED & 51,443 & 200,274 & 197,532 & 27,458 \\
RSNA-Mammo & 3,105 & 12,651 & 12,175 & 1,539 \\
VinDr-Mammo & 103 & 2,157 & 2,757 & 16,867 \\
LAMIS & 486 & 1,343 & 263 & 120 \\
PDS1 & 40 & 77 & 1,165 & 304 \\
PDS2 & 182 & 1,544 & 3,417 & 851 \\
PDS3 & 24 & 216 & 1,304 & 185 \\
PDS4 & 40 & 1,236 & 33,241 & 1,691 \\
PDS5 & 197 & 2,214 & 21,269 & 4,426 \\
\hline
\end{tabular}
\end{table*}

\begin{table*}[h!]
\centering
\caption{Distribution of masking label of CSAW-M.}
\label{tab:masking_sta}
\begin{tabular}{l|rrrrrrrr}
\hline
\diagbox[width=10em]{Dataset}{Label} & Level 1 & Level 2 & Level 3 & Level 4 & Level 5 & Level 6 & Level 7 & Level 8 \\
\hline
CSAW-M & 720 & 1,269 & 1,340 & 1,931 & 1,838 & 1,090 & 858 & 974 \\
\hline
\end{tabular}
\end{table*}

\begin{table*}[h!]
\centering
\caption{Distribution of findings label across different mammogram datasets.}
\label{tab:findings_sta}
\begin{tabular}{l|rrrr}
\hline
\diagbox[width=18em]{Finding}{Dataset} & CBIS-DDSM & CMMD & INbreast & DMID \\
\hline
No finding & 0 & 0 & 68 & 0 \\
Mass & 1,696 & 3,220 & 108 & 255 \\
Architectural distortion & 0 & 0 & 3 & 15 \\
Nipple retraction & 0 & 0 & 0 & 0 \\
Calcification & 1,542 & 1,446 & 308 & 128 \\
Asymmetry & 0 & 0 & 14 & 5 \\
Miscellaneous & 0 & 0 & 0 & 93 \\
\hline
\diagbox[width=18em]{Finding}{Dataset} & MIAS & VinDr-Mammo & EMBED & PDS1 \\
\hline
No finding & 0 & 0 & 401,240 & 1,186 \\
Mass & 43 & 2,228 & 17,858 & 0 \\
Architectural distortion & 19 & 221 & 3,421 & 0 \\
Nipple retraction & 0 & 58 & 0 & 0 \\
Calcification & 27 & 897 & 22,031 & 400 \\
Asymmetry & 15 & 750 & 31,565 & 0 \\
Suspicious lymph node & 0 & 82 & 0 & 0 \\
Skin thickening & 0 & 92 & 0 & 0 \\
Skin retraction & 0 & 31 & 0 & 0 \\
Mass + Asymmetry & 0 & 0 & 1,986 & 0 \\
Asymmetry + Calcification & 0 & 0 & 1,061 & 0 \\
Mass + Calcification & 0 & 0 & 860 & 0 \\
Mass + Asymmetry + Calcification & 0 & 0 & 94 & 0 \\
Mass + Architectural distortion & 0 & 0 & 84 & 0 \\
Arch. distortion + Calcification & 0 & 0 & 75 & 0 \\
Miscellaneous & 15 & 0 & 0 & 0 \\
\hline
\end{tabular}
\end{table*}

\begin{table*}[h]
\centering
\caption{Distribution of BI-RADS label across different mammogram datasets. 0 to 5 represent BI-RADS  0 to BI-RADS  5 categories.}
\label{tab:birads_sta}
\begin{tabular}{l|rrrrrr}
\hline
\diagbox[width=10em]{Dataset}{Label} & 0 & 1 & 2 & 3 & 4 & 5 \\
\hline
BMCD & 0 & 88 & 112 & 0 & 192 & 8 \\
CBIS-DDSM & 238 & 3 & 356 & 423 & 1,543 & 540 \\
CDD-CESM & 0 & 371 & 167 & 136 & 195 & 130 \\
INbreast & 0 & 67 & 220 & 23 & 43 & 49 \\
KAU-BCMD & 0 & 1,863 & 0 & 387 & 100 & 20 \\
EMBED & 60,576 & 306,708 & 75,377 & 24,824 & 10,419 & 999 \\
RSNA-Mammo & 8,249 & 15,772 & 2,265 & 0 & 0 & 0 \\
VinDr-Mammo & 0 & 13,406 & 4,676 & 1,060 & 1,661 & 1,081 \\
LAMIS & 0 & 910 & 1,150 & 0 & 74 & 78 \\
PDS1 & 1 & 399 & 785 & 130 & 247 & 24 \\
PDS2 & 9 & 8 & 1,359 & 1,915 & 2,343 & 310 \\
PDS3 & 2 & 32 & 405 & 1,076 & 119 & 91 \\
PDS4 & 7 & 400 & 7,529 & 24,359 & 5,655 & 204 \\
PDS5 & 399 & 8,265 & 5,666 & 6,036 & 3,187 & 2,567 \\
\hline
\end{tabular}
\end{table*}

\begin{table*}[h]
\centering
\caption{Distribution of BI-RADS 4 subcategory label across different mammogram datasets.}
\label{tab:birads4_sta}
\begin{tabular}{l|rrr}
\hline
\diagbox[width=10em]{Dataset}{Label} & BI-RADS  4A & BI-RADS  4B &BI-RADS  4C \\
\hline
PDS2 & 1,075 & 0 & 310 \\
PDS4 & 4,492 & 475 & 467 \\
PDS5 & 903 & 888 & 1,392 \\
\hline
\end{tabular}
\end{table*}

\begin{table*}[h]
\centering
\caption{Distribution of pathology label across different mammogram datasets.}
\label{tab:pathology_sta}
\begin{tabular}{l|rr}
\hline
\diagbox[width=10em]{Dataset}{Label} & Benign & Malignant \\
\hline
CBIS-DDSM & 1,911 & 1,327 \\
CMMD & 1,112 & 2,632 \\
DMID & 284 & 153 \\
MIAS & 68 & 51 \\
CDD-CESM & 331 & 331 \\
RSNA-Mammo & 53,548 & 1,158 \\
MM & 620 & 125 \\
NLBS & 23,400 & 306 \\
LAMIS & 2,060 & 152 \\
DBT & 70 & 65 \\
PDS1 & 1,002 & 182 \\
PDS2 & 3,342 & 2,774 \\
PDS3 & 1,378 & 355 \\
PDS4 & 2,058 & 297 \\
PDS5 & 2,772 & 5,099  \\
\hline
\end{tabular}
\end{table*}

\begin{table*}[h]
\centering
\caption{Distribution of mastitis label of PDS5.}
\label{tab:mastitis_sta}
\begin{tabular}{l|rr}
\hline
\diagbox[width=10em]{Dataset}{Label} & Malignant &Mastitis \\
\hline
PDS5 & 5,099 & 674 \\
\hline
\end{tabular}
\end{table*}

\begin{table*}[h]
\centering
\caption{Distribution of invasive label of RSNA-Mammo.}
\label{tab:invasive_sta}
\begin{tabular}{l|rr}
\hline
\diagbox[width=10em]{Dataset}{Label} & Non-invasive & Invasive \\
\hline
RSNA-Mammo & 340 & 818 \\
\hline
\end{tabular}
\end{table*}

\begin{table*}[h]
\centering
\caption{Distribution of molecular subtype label across different mammogram datasets.}
\label{tab:subtype_sta}
\begin{tabular}{l|rrrr}
\hline
\diagbox[width=10em]{Dataset}{Label} & Luminal A & Luminal B & HER2-enriched & Triple-negative \\
\hline
CMMD & 304 & 752 & 270 & 172 \\
PDS5 & 1,212 & 2,995 & 266 & 257 \\
\hline
\end{tabular}
\end{table*}

\begin{table*}[htbp]
\centering

\caption{Distribution of BI-RADS label across different subgroups in the subgroup analysis on the EMBED dataset. 0 to 5 represent BI-RADS 0 to BI-RADS 5 categories.}
\label{tab:birads_distribution}

\begin{tabular}{l |c c c c c c c}
\toprule
\diagbox[width=20em]{Subgroup}{Label} & {0} & {1} & {2} & {3} & {4} & {5} \\
\hline
{\textbf{Breast composition:}} \\
     Density A & 950 & 7,460 & 1,325 & 4,03 & 126 & 27 \\
     Density B & 4,834 & 25,260 & 5,934 & 2,030 & 636 & 60 \\
     Density C & 5,579 & 23,781 & 6,729 & 2,023 & 882 & 91 \\
     Density D & 716 & 3,104 & 986 & 332 & 153 & 8 \\
\hline
{\textbf{Clinical scenario:}} \\
     Screening & 9,454 & 61,768 & 14,947 & 3,134 & 1,112 & 124 \\
     Diagnostic & 22,383 & 45,616 & 41,851 & 26,058 & 7,614 & 906 \\
\hline
{\textbf{{Ethnicity}:}} \\
     African American or Black & 31,204 & 140,781 & 46,783 & 20,344 & 7,083 & 590 \\
     American Indian or Alaskan Native & 303 & 523 & 203 & 163 & 18 & 0 \\
     Asian & 3,663 & 19,621 & 4,983 & 1,678 & 524 & 20 \\
     Caucasian or White & 22,873 & 126,344 & 44,457 & 15,314 & 3,938 & 611 \\
     Multiple & 269 & 797 & 164 & 0 & 18 & 0 \\
     Native Hawaiian or Other Pacific Islander & 558 & 2,579 & 581 & 80 & 94 & 0 \\
     Unavailable & 3,724 & 12,186 & 4,126 & 1,699 & 544 & 0 \\
\hline
{\textbf{Age:}} \\
     $<$40 & 1,069 & 4,429 & 2,388 & 878 & 504 & 16 \\
     40--49 & 14,350 & 53,192 & 16,786 & 7,900 & 2,547 & 129 \\
     50--59 & 16,965 & 79,047 & 23,396 & 8,764 & 3,362 & 362 \\
     60--69 & 15,102 & 83,447 & 25,874 & 10,636 & 2,572 & 329 \\
     70+ & 15,108 & 82,716 & 32,853 & 11,100 & 3,234 & 385 \\
\bottomrule
\end{tabular}
\end{table*}

\end{document}